\documentclass[Afour,sageh,times]{IJRR/sagej}

\usepackage{moreverb,url}

\usepackage{url,lineno,microtype}
\usepackage[dvipsnames]{xcolor}
\usepackage[%
hyperfigures=true,%
backref=page,%
pagebackref=true,%
breaklinks=true,%
colorlinks=true,%
citecolor=darkgray,%
linkcolor=black%cyan%blue
]{hyperref}
\usepackage{amsfonts}
\usepackage{amsthm}
\usepackage{amsmath}
\usepackage[mathscr]{euscript}
\usepackage{bm}
\usepackage{enumitem}
\setenumerate{itemindent=0.0cm,labelsep=0.35cm, leftmargin =1.35cm}
\usepackage{multicol}
\usepackage[ruled,vlined ]{algorithm2e}
\usepackage{booktabs}
\usepackage{soul}
\usepackage{tablefootnote}
\usepackage{array}

\newcounter{assumptionCounter}
\newcommand{\theAssumptionCounter}{\arabic{assumptionCounter}}

\newcolumntype{N}{>{\refstepcounter{assumptionCounter}\theAssumptionCounter}c}

\newcommand{\Oeps}{{\mathcal{O}(\epsilon)}} 
\newcommand{\OepsSQ}{{\mathcal{O}(\epsilon^2)}} 
\newcommand{\x}{\bm{x}}
\newcommand{\y}{\bm{y}}
\newcommand{\z}{\bm{z}}
\newcommand{\zs}{\bm{q}_s}
\newcommand{\zp}{\bm{q}_a}
\newcommand{\g}{h_\Xi}
\newcommand{\h}{h_\omega}
\newcommand{\hs}{h_s}
\newcommand{\q}{\bm{q}}
\newcommand{\atan}{\text{atan}}
\def \Re {P} % the exact 3 dof return map
\def \P{A} % the exact 1 dof pitch return
\def \Se {S}  % the exact perturbed  translational return map
\def \Sep {S_{\phi^*}}  % the exact unperturbed  translational return map
\def \Shp { \widehat{S}_{ \phi^* }} % the averaged unperturbed translational return map
\newcommand{\mynorm}[1]{ | #1 | }

\usepackage{ulem}
\newcommand{\DIFadd}[1]{#1}

\theoremstyle{definition}
\newtheorem{definition}{Definition}
\newtheorem{conjecture}{Conjecture}
\newtheorem{lemma}{Lemma}
\newtheorem{theorem}{Theorem}
\newtheorem{proposition}{Proposition}
\newtheorem{corollary}{Corollary}
\newtheorem{observation}{Observation}

\newtheorem*{remark}{Remark}
\usepackage{cancel}

\usepackage[inline]{trackchanges}

\usepackage{mathtools} % for coloneqq
\usepackage[inter-unit-product=\cdot]{siunitx} % makes units *so* much easier
\DeclareSIUnit\leglengths{leg\ lengths}
\usepackage{collcell}
\usepackage{graphicx}
\graphicspath{ {./figs/} }
\usepackage{ragged2e}
\usepackage{tabularx}
\usepackage[nameinlink]{cleveref}
\crefdefaultlabelformat{#2\text{#1}#3} % <-- Only #1 in \text
\creflabelformat{equation}{\text{\textup{(#2#1#3)}}}
\crefname{figure}{\text{Figure}}{\text{Figures}}
\Crefname{figure}{\text{Figure}}{\text{Figures}}
\crefname{equation}{}{}
\Crefname{equation}{}{}
\crefname{definition}{\text{Definition}}{\text{Definitions}}
\Crefname{definition}{\text{Definition}}{\text{Definitions}}
\crefname{lemma}{\text{Lemma}}{\text{Lemmas}}
\Crefname{lemma}{\text{Lemma}}{\text{Lemmas}}
\crefname{conjecture}{\text{Conjecture}}{\text{Conjectures}}
\Crefname{conjecture}{\text{Conjecture}}{\text{Conjectures}}
\crefname{corollary}{\text{Corollary}}{\text{Corollaries}}
\Crefname{corollary}{\text{Corollary}}{\text{Corollaries}}
\crefname{theorem}{\text{Theorem}}{\text{Theorems}}
\Crefname{theorem}{\text{Theorem}}{\text{Theorems}}
\crefname{proposition}{\text{Proposition}}{\text{Propositions}}
\Crefname{proposition}{\text{Proposition}}{\text{Propositions}}
\crefname{table}{\text{Table}}{\text{Tables}}
\Crefname{table}{\text{Table}}{\text{Tables}}
\crefname{section}{\text{Section}}{\text{Sections}}
\Crefname{section}{\text{Section}}{\text{Sections}}
\crefname{paragraph}{\text{Section}}{\text{Sections}}
\Crefname{paragraph}{\text{Section}}{\text{Sections}}
\crefname{assumptionCounter}{\text{Assumption}}{\text{Assumptions}}
\Crefname{assumptionCounter}{\text{Assumption}}{\text{Assumptions}}
\Crefname{observation}{\text{Observation}}{\text{Observations}}
\crefname{observation}{\text{Observation}}{\text{Observations}}
\Crefname{appendix}{\text{Appendix}}{\text{Appendix}}
\crefname{appendix}{\text{Appendix}}{\text{Appendix}}

\usepackage{microtype}
\usepackage{upgreek}

\usepackage{breakurl}

\newcounter{subfigure}

\newcounter{subtable}

\let\oldequation\equation
\let\oldendequation\endequation
\renewenvironment{equation}
    {\linenomathNonumbers\oldequation}
    {\oldendequation\endlinenomath}
\let\oldalign\align
\let\oldendalign\endalign
\renewenvironment{align}
    {\linenomathNonumbers\oldalign}
    {\oldendalign\endlinenomath}

\newcommand{\vctfour}[4]{
\begin{bmatrix}
#1 \\ #2 \\ #3 \\ #4 \\
\end{bmatrix}
}
\newcommand{\vctthree}[3]{
\begin{bmatrix}
#1 \\ #2 \\ #3\\
\end{bmatrix}
}
\newcommand{\vcttwo}[2]{
\begin{bmatrix}
#1 \\ #2 \\
\end{bmatrix}
}
\makeatletter
\@addtoreset{subfigure}{figure}
\makeatother

\usepackage[english]{babel}

\def\volumeyear{2023}
\begin{document}

\runninghead{Rozen-Levy et al.}

\title{Hip Energized Monopedal Hopping}

\author{Shane Rozen-Levy\affilnum{1}, Griffon McMahon\affilnum{1}, and Daniel Koditschek\affilnum{2}}

\affiliation{
\affilnum{1} GRASP Lab, Department of Mechanical Engineering and Applied Mechanics, University of Pennsylvania, Philadelphia, PA, USA\\
\affilnum{2} GRASP Lab, Department of Electrical and Systems Engineering, University of Pennsylvania, Philadelphia, PA, USA}

\corrauth{Shane Rozen-Levy, 3401 Grays Ferry Ave, Philadelphia, PA, 19146}

\email{srozen01@seas.upenn.edu}

\begin{abstract}
We present a novel stepping strategy for pitch unlocked planar monopeds where the reaction torques from stabilizing pitch with a conventional PD + feedfoward controller are recruited to counteract energetic losses from damping.
By moving the location of the mass center, our controller increases the pitch stabilization torque, thereby adding energy to the gait.
A new stepping policy adjusts the distribution of energy between the radial and angular degrees of freedom to counteract dissipative losses and achieve a user specified balance between steady state fore-aft speed and apex height.
Hybrid averaging analysis yields closed form expressions for the fixed points and eigenvalues of the resulting gait, lending insight into the interplay between the physical and control parameters' influence on performance. Simulation studies on a generic 5 link biped and a careful model of the Penn Jerboa reveal a useful correspondence to these analytical predictions. Physical experiments on the Penn Jerboa exhibit stable locomotion with speeds  ranging from $\SI{1.02}{m/s}$ to $\SI{1.77}{m/s}$ ($\SI{5.10}{\text{leg lengths}/s}$ to $\SI{8.85}{\text{leg lengths}/s}$) in a manner effectively approximated by the mathematical analysis. 
\end{abstract}


\maketitle

\section{Introduction}
Recent progress in the design and control of legged robots has led to impressive demonstrations of agility and robustness \citep{bledt_implementing_2019, bledt_extracting_2020, kumar_rma_2021,hwangbo_learning_2019,lee_learning_2020,siekmann_blind_2021,yim_proprioception_2023,nahrendra_dreamwaq_2023,chen_integrable_2023,margolis_rapid_2022}. At the same time, the design of quadrupeds has converged to robots with 3 motors per leg in a serial configuration \citep{katz_mini_2019, hutter_anymal_2016, bledt_mit_2018}, allowing robots to command arbitrary ground reaction forces with each leg subject to friction cone constraints. This paradigm enables control strategies that plan in centroidal dynamics ignoring kinematic concerns \citep{bledt_implementing_2019, bledt_mit_2018, dai_planning_2016}. The benefits here are clear, yet questions arise regarding the need for such highly actuated  legs, or in view of inevitable mass-specific power limitations,  whether it might be better to either redistribute a subset of actuators elsewhere or design controllers which better utilize the available power with the goal of all available actuators contributing useful work during locomotion. 

\begin{figure}
    \centering
    \includegraphics[width=1\linewidth]{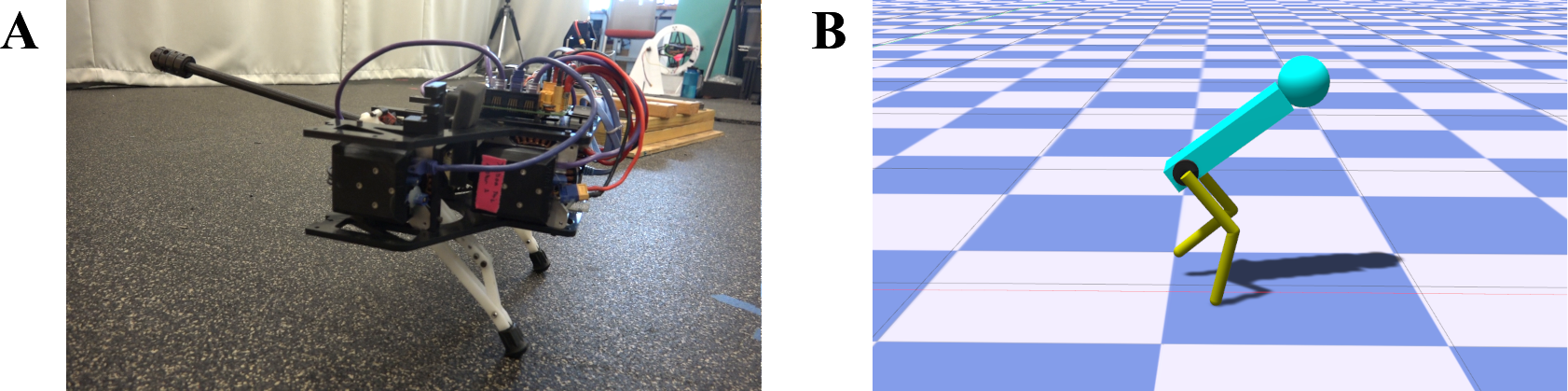}
    \caption{The two robots on which we test our control strategy, \textbf{(A)} is the Penn Jerboa \citep{shamsah_analytically-guided_2018}, a tailed biped with springy legs and only four actuators: one driving the leg angle at each hip and two controlling the 2 DoF tail. \textbf{(B)} is a generic planar 5-link biped on which we tested our controller. Most of the mass is in the torso, and the legs act like virtual springs.}
    \label{fig_robot_models}
    \begin{minipage}[b]{0.2\linewidth}
    \refstepcounter{subfigure}\label{fig_jerboa}
    \end{minipage}
    \begin{minipage}[b]{0.2\linewidth}
    \refstepcounter{subfigure}\label{fig_planr_biped}
    \end{minipage}
\end{figure}

The Penn Jerboa \citep{de_penn_2015} in \cref{fig_jerboa} attempts to address some of these questions by using a high-powered 2 DoF tail at the expense of affording only one actuator at each hip of its passive spring-loaded legs. Moreover, unlike many bipeds, it features small point toes rather than flat feet \citep{apgar_fast_2018}, limiting its angular momentum affordance in stance in exchange for fewer actuators in its legs. Jerboa's more proximal and lower number of actuators forces it to operate in a comparatively more energetic regime where all of the motors are doing useful work, suggesting that it should be able to turn much of its high power density ($\SI{43.2}{W/kg}$ \citep{kenneally_design_2016}) into speed and height.
Previous work on Jerboa has focused on using the tail to drive energy into the shank spring \citep{shamsah_analytically-guided_2018, de_penn_2015}, acting as a spring-loaded inverted pendulum (SLIP) \citep{Saranli_Schwind_Koditschek_1998}. Shank-actuated SLIP has its own benefits, but this work pursues the previous questions concerning actuator distribution by
relegating the tail actuators' role to (at most) that of low-power ``shape shifters,''  thereby exploring the possibility that a hip-actuated monoped can achieve full dynamic sagittal plane mobility using only one motor \citep{saranli_rhex_2001}.  The empirical success and analytical insight resulting from this study in extreme underactuated  mobility advances the project of  developing SLIP template control schemes for more generic anchoring bodies \citep{full_templates_1999} whose  abundant high-power actuators can more effectively energize the
system to achieve more agile behaviors.

In this paper we present a hip energized control strategy for dynamical monoped locomotion taking the form of a new stepping controller \cref{eq_stepping} that recruits the pitch stabilizing reaction torques to energize the mass center.
We leverage hybrid averaging with a novel set of coordinates  \cref{eq_hybrid_vars} to analyze the new control strategy, achieving formal stability guarantees in \cref{lemma_cascade_stability} and closed form expressions for fixed points as a function of  physical and control parameters in \cref{eq_3dof_slip_fixed}.\endnote{This is the first time in the literature that hybrid averaging has been applied to 2 DoF SLIP \citep{de_modular_2017,de_averaged_2018}. Though outside the scope of the paper, work presently in progress aims to demonstrate that the new methods presented  this paper can be used to analyze a classical shank-actuated 2 DoF SLIP unlike a previous application of hybrid averaging to SLIP which was restricted to 1.5 DoF SLIP by neglecting leg angle \citep{de_modular_2017}.} 
The analysis suggests that in cases where a biped/monoped's center of mass is not stacked directly over the hip, the orientation control should also be recruited to to energize the translational subsystem. In these cases, an asymmetric stepping strategy is beneficial --- and under reasonable assumptions, as shown in \cref{theorem_asymmetry},  will actually  be a necessary feature of steady state hopping --- to realize the energetic gains. We have implemented this strategy in simulation on two different robot morphologies and on the hardware Jerboa where it achieved stable speeds ranging from $\SI{1.02}{m/s}$ to $\SI{1.77}{m/s}$ (i.e., $5.1$\,leg~lengths$/$s to $8.85$\,leg~lengths$/$s). 
Comparisons to the tail-energized hopping results of \cite{shamsah_analytically-guided_2018} who reported speeds ranging from~$0.2$ -- $\SI{1.0}{m/s}$ on this same machine suggest that a hip energized hopping strategy might offer some advantages for directing a biped's available power budget toward its running behavior\endnote{\DIFadd{This comparison extends to more general purpose spatial-bipeds such as Cassie's $\SI{100}{m}$ dash world record-breaking average of 
$\SI{4}{m/s}$ (i.e., $\sim 4$ leg lengths/s)  \citep{Life_at_OSU} and Atlas's maximum reported speed of $\SI{2.5}{m/s}$ (i.e., $\sim 2.5$ leg lengths/s) \citep{noauthor_atlas_nodate}. These comparisons remain purely suggestive because a careful comparison of speed is complicated by the fact that Atlas and Cassie are spatial robots while Jerboa is a boom-restricted sagittal plane robot, the different length scales of the robots, and in the case of Atlas,
the substantial differences between hydraulic and direct electromechanical actuation.}}.
More broadly, the analysis and empirical results presented here offer a better understanding of how hip torques can be used to energize dynamical locomotion in legged machines.

\subsection{Background Literature}
Prior work on hip energized SLIP has focused either on approximating the SLIP return map while achieving a constant angular momentum \citep{ankarali_stride--stride_2010} or on using the hip torque of a SLIP-like robot to direct the ground reaction force through a body frame virtual pivot point (VPP, also virtual point) \citep{maus_upright_2010}.
After relating these numerical simulation studies to our results, we will conclude this review of salient prior work with a brief introduction to our central analytical tool, hybrid averaging \citep{de_hybrid_2018}. Contrasting with our appeal to those foregoing numerical and analytical advances, to the best of our knowledge, no empirical demonstration of hip energized monopedal hopping has previously appeared in the legged locomotion literature. 

Our hip torque controller takes substantial inspiration from the simple, open loop strategy introduced by \cite{ankarali_stride--stride_2010} in that  it is nearly constant and open loop once the pitch subsystem achieves its fixed point.  However, their approach does not stabilize pitch as does ours. Moreover, their SLIP numerical stance map approximation --- resulting from linearizing the leg angle about vertical and iterative approximation of the mean angular momentum --- cannot afford the insights (explicit functional expressions for the fixed points and stability conditions) provided by our mathematical analysis. Finally, they do not implement their controller on a physical robot.

% Working toward the first focus, \cite{ankarali_stride--stride_2010} approximate the stance map of SLIP by linearizing the leg angle about vertical and using an open loop hip torque before iteratively approximating a mean angular momentum. This paper suggests that a simple hip torque (in this case open loop) is sufficient in a hip energized regime. While the method accurately predicts the fixed points, the iterative process makes finding fixed points and evaluating stability numerical. Moreover, they do not implement the control strategy on a physical robot, and their control strategy does not stabilize pitch. Our controller take inspiration from \cite{ankarali_stride--stride_2010} as our energizing hip torque is simple and nearly open loop once the pitch subsystem is at its fixed point.

Our controller has also been partly inspired by the  VPP phenomenon   wherein animals’ gaits produce ground reaction forces that appear to intersect at a fixed point in the body frame.  Its decade prior identification and study 
\citep{maus_upright_2010, muller_force_2017, blickhan_global_2018, drama_postural_2020, drama_virtual_2020} offers  the first (and, to the best of our knowledge, heretofore, the only) account in the legged locomotion literature addressing the problem of stabilizing pitch while also stably energizing hopping through application of  hip torques.
Specifically,  numerical simulation suggests that hip energized  SLIP with attitude can be stabilized  by a control strategy that targets a VPP \citep{sharbafi_controllers_2012}, whose refined placement and variation can  further improve the resulting gait \citep{firouzi_tip_2019, lee_force_2017, vu_control_2017, sharbafi_robust_2013}.  
Much of this numerical work combines the VPP-targeting  hip torque with a stepping strategy that fixes  the velocity vector's touchdown angle relative to vertical  \citep{peuker_leg-adjustment_2012,sharbafi_robust_2013}, with subsequent improvements achieved by an adjusted  convex combination of the angle between the touchdown velocity and gravitational force  vectors   \citep{firouzi_tip_2019,sharbafi_robust_2013}.  In contrast, our provably correct controller does not target a VPP but, nevertheless, appears to achieve it (see \cref{fig_vpp_sim} and the associated discussion in \cref{sec_vpp}). Again to the best of our knowledge, there has been only one attempted  physical implementation of a VPP-targeting  controller on a biped (reported in the MS thesis work of \cite{peekema_template-based_2015}) which suggests that further analytical work beyond the scope of this  paper might well elucidate the stability properties of that  intriguing control approach and reveal further missing key features to facilitate its apparently challenging jump from simulation to hardware.

Hybrid averaging \citep{de_hybrid_2018} is an extension of classical averaging \citep{guckenheimer_nonlinear_2013} to hybrid dynamical systems with only one hybrid mode and whose initial development was used to analyze monopedal hopping \citep{de_hybrid_2018} and quadrupedal running \citep{de_vertical_2018}. 
We are able to extend the analysis of these previous 1.5 DoF models to the full 2 DoF SLIP by slightly relaxing the algorithmic check list of sufficient conditions developed in \cite{de_modular_2017} while still retaining  the formal guarantees of \cite{de_averaged_2018} that achieve $\epsilon$-close fixed points and stability of the Poincar\'{e} return map, even absent integrable dynamics. 
A more  recent prior contribution to the hybrid averaging literature merely conjectures averageability \citep{shamsah_analytically-guided_2018}.
In addition, an interesting new contribution to this literature \citep{sun_spring-loaded_2023,sun_posture_2023} uses hybrid averaging to design an empirically useful VPP based feedback controller for a quadruped,
 but replaces assumptions governing relative parameter magnitudes and constant approximations to (some narrow range of values attained by nonlinear-in-state) expressions in the dynamics (e.g. Tables \ref{table_decomp_assumptions}, \ref{table_hybrid_averaging_assumptions}, and \ref{table_assumption_reset}) with assumptions that impose  a priori small bounds on states --- tantamount to assuming stability in the first place.

In the space of dynamic tailed robots there have been several studies including prior work on the Jerboa in \citep{shamsah_analytically-guided_2018} where the authors explore a tail energized gait on Jerboa capable of up to 20 hops in a pitch unlocked regime (compared to the unlimited hopping achieved in this).  In \citep{an_development_2020}, the authors explore a different tail-energized gait on a Jerboa-like robot. While the theoretical controller promises additional affordances over rate-of-convergence in comparison to the \citep{shamsah_analytically-guided_2018} controller, the empirical results are limited to a whose accompanying videos suggest that it is barely able hop. In simulation work of \citep{liu_feedback_2021,liu_how_2023}, the authors explore an articulated-tail using feedback linearization and numerical optimal control to demonstrate some of the advantages of tails for quadrupedal and bipedal locomotion. In an interesting but somewhat further removed work by \citep{heim_designing_2016}, the authors use an open loop CPG controller with a tail to stabilize body pitch in the Cheetah-Cub robot suggesting that tails can help stabilize pitch. The Cheetah-Cub achieved speeds of up to  $\SI{0.56}{m/s}$ and 63 strides with a bounding gait. The work reported here advances beyond these prior contributions by enabling higher speed and more reliable hoping whose empirical performance we undergird with a complete stability proof for this sagittal plane setting.  

\subsection{Contributions}
\begin{table*}[ht]
\centering
\small\sf%\centering
\caption{List of the various models, results, and assumptions}
\label{table:models}
\begin{tabular}{l
>{\raggedright\arraybackslash}p{3.5cm}
l
>{\raggedright\arraybackslash}p{2.5cm}
>{\raggedright\arraybackslash}p{3.5cm}}
\toprule
              \textbf{Case} & \textbf{Model} & \textbf{Fixed} \textbf{Points} & \textbf{Stability} & \textbf{Assumptions}   \\ 
\midrule
3 DoF SLIP    & \cref{fig_ASLIP,eq_slip_att_ass_dyn} & \cref{eq_3dof_slip_fixed}& \Cref{lemma_cascade_stability} & \Cref{table_decomp_assumptions} to decompose the system and \cref{table_hybrid_averaging_assumptions,table_assumption_reset} for SLIP hybrid averaging\\
\midrule
1 DoF Pitch   & \cref{fig_pitch,eq_pitch_dyn}      &  \cref{eq_pitch_fixed_points} & \cref{lemma_slip_pitch}&             \cref{table_decomp_assumptions}  \\
\midrule
2 DoF  SLIP   & \cref{fig_SLIP,eq_slip_dyn,eq_reset_slip} & \cref{eq_x_star} & \cref{lemma_slip_stable} & \cref{table_hybrid_averaging_assumptions,table_assumption_reset}  \\
\midrule
Planar Biped  & \cref{fig_planr_biped}      &\cref{fig_energy_architecture}&  Simulation  & N/A \\
\midrule
Planar Jerboa & \cref{fig_jerboa}      & \cref{fig_sim_height_speed,fig_hard_hybrid_params}&  Simulation and hardware &  N/A \\
\bottomrule
\end{tabular}
\end{table*}

This paper introduces an analytically tractable hip-only actuated controller for 3 DoF (pitch-unlocked) SLIP with attitude capable of producing stable steady state hopping in simulation and hardware.  A variety of physical insights motivate a series of assumptions (listed in Tables \ref{table_decomp_assumptions}, \ref{table_hybrid_averaging_assumptions}, and \ref{table_assumption_reset}, in the order they are applied) to decompose the stance dynamics as the cascade composition of the 1 DoF attitude perturbing the 2 DoF translational subsystem, thereby inspiring the control strategy and simplifying the analysis.  The results, summarized in  \cref{table:models} include:
\begin{description}
    \item[A Novel Control Strategy: ]
    As summarized in \cref{table:controllers}, the stance control is used to stabilize pitch and, in so doing, energizes the translational subsystem. The stepping controller adjusts the leg touchdown angle so as to redistribute energy between the attitude and translational subsystems, counteracting energetic losses due to damping and ensuring a balanced distribution of energy between the fore-aft and vertical degrees of freedom.
    \item[3 DoF Fixed Point and Stability Analysis: ] \Cref{lemma_cascade_stability} exploits the assumptions and the linear time invariant structure of the pitch subsystem (affording straightforward identification of its fixed points and their stability  in \cref{lemma_slip_pitch}) to justify focusing analytical effort on the unperturbed translational (SLIP) subsystem.  The fixed points and stability of this highly nonlinear 2 DoF system are rendered tractable by an application of hybrid averaging (the first time that a physically realistic model of the full SLIP dynamics has yielded analytical results to the best of our knowledge) in \cref{lemma_slip_stable},  made possible by recourse to a novel change of coordinates \cref{eq_hybrid_vars}.
    % \item Analysis of pitch unlocked SLIP where using the assumptions in \cref{table_decomp_assumptions} we decompose the pitch unlocked SLIP model into an isolated hybrid pitching subsystem (\cref{sec_attitude}) which cascades via the hip torque into a perturbed translational SLIP subsystem (\cref{sec_slip}) as depicted in \cref{fig_robot_diagrams}. In \cref{lemma_cascade_stability} we show stability of the full pitch unlocked SLIP model using linear analysis of the pitching subsystem (\cref{lemma_slip_pitch}) and an extension of hybrid averaging \cref{sec_hybrid_averaging} applied to the SLIP subsystem (\cref{lemma_slip_stable}) using a novel set of coordinates \cref{eq_hybrid_vars}.
    \item [The Necessity of Stepping Asymmetry: ] Our choice of coordinates reveals that there is a fundamental connection between an asymmetric stepping strategy and a hip energized regime as shown in \cref{theorem_asymmetry}. The asymmetric stepping strategy is necessary to introduce a strong hybrid coupling between the weakly coupled radial and angular subsystems of SLIP in stance and prevent the energy of the angular subsystem from increasing in an unbounded fashion while energy in the radial subsystem decreases to zero.
    \item[Numerical and Hardware Implementation: ]  Simulation of the controller on a planar biped and tailed biped  where the range of hopping steady states in \cref{fig_energy_architecture} is predicted with roughly 6--14\% accuracy by the analytical model as summarized in \cref{tab_sim_accuracy}. Simulation analysis of the control strategy reveals a fourfold affordance over energy in both the planar biped and tailed biped \cref{fig_energy_architecture}. Moreover, hardware experiments on a tailed biped with a range of hopping steady states plotted in \cref{fig_hard_hybrid_params} are predicted with roughly 12--16\% accuracy by the analytical model as summarized in \cref{tab_hard_accuracy}.
    These hardware results document the high speed sagittal plane (boom constrained) hopping ($5.1$\,leg~lengths$/$s to $8.85$\,leg~lengths$/$s) mentioned above. 
        % \item An implementation of our control strategy in simulation on a planar biped and tailed biped and in hardware on a tailed biped. The simulation (\cref{sec_sim_fixed_points}) and hardware (\cref{sec_hard_accuracy}) validate the accuracy of our model. The hardware implementation on the tailed biped was able to hop endlessly with speeds ranging from  $\SI{1.02}{m/s}$ to $\SI{1.77}{m/s}$ ($5.1$\,leg~lengths$/$s to $8.85$\,leg~lengths$/$s).

\end{description}% The core contribution of this paper is the analysis of a hip energized controller for the pitch unlocked SLIP model providing insight into how and when a hip motor can energize steady state hopping and running as summarized in \cref{table:models}.

\subsection{Paper Overview}
The paper begins with a mathematical analysis of hip energized pitch unlocked SLIP in \cref{sec_math}.
\cref{sec_detail} details the small adjustments to our controller that we made to improve the basin of attraction in simulation and hardware.
\Cref{sec_sim_results} then presents the results of implementing the controller in simulation for Jerboa and the 5-link biped, and \cref{sec_hard_results} presents the hardware results for Jerboa.
The paper concludes with a brief summary of results and immediate next steps in 
\cref{sec_discussion,sec_discussion}. 

\begin{table*}[ht]
\centering
\small\sf%\centering
\caption{Symbol definitions\label{T2} }
\begin{tabular}{l >{\RaggedRight}p{0.49\linewidth} >{\centering\arraybackslash} m{2.45cm}}
\toprule
\textbf{Symbol}&\textbf{Brief Description} & \textbf{First Appearance}\\ 
\hline
\midrule
$\bm p = [p_x,p_z]$ & Fore-aft and vertical location of the COM in the world frame& \cref{fig_robot_diagrams}\\
\hline

$\phi$& Body pitch &  \cref{eq_slip_att_ass_dyn}  \\
\hline

$r,\theta$& Leg length, leg angle & \cref{eq_slip_att_ass_dyn} \\
\hline

$m, \, I$ & Body mass, body inertia  &  \cref{eq_slip_att_ass_dyn} \\
\hline

$k, \, r_0$&  Leg spring constant, rest length  & \cref{eq_slip_att_ass_dyn} \\
\hline

$\tau$& Hip torque  applied between leg and body& \cref{eq_slip_att_ass_dyn}  \\
\hline

$b$&  Leg extension damping coefficient  & \cref{eq_slip_att_ass_dyn} \\
\hline

$d$ & Distance between center of mass and hip &  \cref{eq_slip_att_ass_dyn} \\
\hline
$g$ & Acceleration due to gravity & \cref{eq_slip_att_ass_dyn}\\
\hline
$\omega_r$ & Natural frequency of the leg spring and body mass system & \cref{eq_slip_att_ass_dyn}\\
\hline
$d_x$ & $x$ projection of distance between center of mass and hip &  \cref{eq_ddphi} \\
\hline
$F_s = - k (r-r_0) - b \dot r$ & Radial force from the (unactuated) damped springy &  \cref{ass_pitch} \\
\hline
$k_p,k_d$ & Proportional, derivative gain for continuous time pitch controller & \cref{eq_torque} \\
\hline
$\bar\tau$ & Feedforward hip torque term & \cref{eq_torque}\\
\hline
$\square^d$ & Desired value for $\square$ &  \cref{eq_torque}\\
\hline 
$k_\tau$ & Integral gain for discrete time pitch controller & \cref{eq_discrete_torque} \\
\hline
\\[-9pt]
$\zs = [r, \theta, \dot r, \dot \theta]^T \in \mathcal{S} \coloneqq T(\mathbb{R}_+ \times \mathbb{S}^1)$ & Translational 2 DoF SLIP coordinates & \cref{eq_slip_state}\\ 
\hline
\\[-9pt]
$\zp = [\phi, \dot \phi, \bar\tau]^T \in \mathcal{A} \coloneqq T\mathbb{S}^1 \times \mathbb{R}$ & Orientation (1 DoF pitch attitude including hybrid discrete hip torque) coordinates & \cref{eq_pitch_state}\\
\hline
$\Re: \mathcal{S }\times \mathcal{A} \rightarrow \mathcal{S} \times \mathcal{A}$ & Exact 3 DoF SLIP with attitude liftoff return map & \cref{eq_3dof_return} \\
\hline 
$\P: \mathcal{A} \rightarrow \mathcal{A}$ & Isolated pitch attitude return map & \cref{eq_pitch_return} \\
\hline
$\Se:\mathcal{S} \times \mathcal{A} \rightarrow \mathcal{S}$ & Perturbed translational SLIP return map & \cref{eq_slip_perturbed_map} \\
\hline
$\square^*$ & Fixed point for $\square$ & \cref{eq_slip_unpertubed_map} \\
\hline
$\Sep \coloneqq S(\cdot, \zp^*) : \mathcal{S} \rightarrow \mathcal{S} $ & Unperturbed translational SLIP return map & \cref{eq_slip_unpertubed_map}\\ 
\hline
\\[-9pt]
$\Shp\ : \mathcal{S} \rightarrow \mathcal{S}$ & Averaged translational SLIP return map & \cref{eq_slip_avg_return} \\
\hline
\\[-9pt]
$\hat\square$ & Averaged quantity for $\square$ &  \cref{prop_slip_hybrid}\\
\hline
$\psi_r$ & Ratio of signed radial kinetic to spring-potential energy & \cref{eq_hybrid_vars} \\
\hline

$\psi_e$ & Ratio of  signed angular to radial kinetic and potential energy & \cref{eq_hybrid_vars} \\
\hline

$a_e$ &Twice the mass-specific square root of energy for 2 DoF SLIP energy (units of $\si{m/s}$) & \cref{eq_hybrid_vars} \\
\hline
$\z = [\psi_r, \psi_e, \theta, a_e]$ & State vector in physical (non-averageable) coordinates & \cref{eq_hybrid_vars} \\
\hline
$\hs: \zs \mapsto \z$ & Map from standard SLIP coordinates to physical coordinates& \cref{eq_hybrid_vars}\\
\hline
\\[-9pt]
$\y$  = $[\sigma, \tilde \psi_e, \psi_\theta, a_e]^T$& State vector in offset coordinates & \cref{eq_g_def} \\
\hline
$\g: \z \mapsto \y$ & Map from physical coordinates to offset coordinates &\cref{eq_g_def}\\
\hline
\\[-9pt]
$\x$ = $[\bm \delta^T \bm a^T]^T = [\delta_e, \delta_\theta,  a_e]^T$& State vector in phase difference coordinates & \cref{eq_state_phasediff} \\
\hline
$\h: \y \mapsto \x$ & Map from offset coordinates to phase difference coordinates & \cref{eq_state_phasediff} \\
\hline
$R_z: \z_\text{lo} \mapsto \z_\text{td}$ & SLIP flight map for $\z$ & \cref{eq_reset_slip} \\
\hline

$\Pi_i$ & Projection onto the ith component of a vector & \cref{eq_asymmetry} \\
\hline

$\alpha$ & Low pass filter gain for stepping controller & \cref{eq_stepping}\\
\hline

$k_e$ & Proportional energy ratio gain for stepping controller & \cref{eq_stepping_pre_filt} \\
\hline

$\omega_e(\bm a)$ & The first element of $\omega(\bm a)$ & \cref{eq_stepping_pre_filt}\\
\hline
$\omega_\theta(\bm a)$ & The second element of $\omega(\bm a)$ & \cref{theta_offset}\\
\hline

\bottomrule
\end{tabular}
\end{table*}
\section{Model and Formal Analysis}\label{sec_math}
\label{sec_model}
\begingroup
\begin{figure*}[htp]
	\centering
	\includegraphics[width=0.7\textwidth]{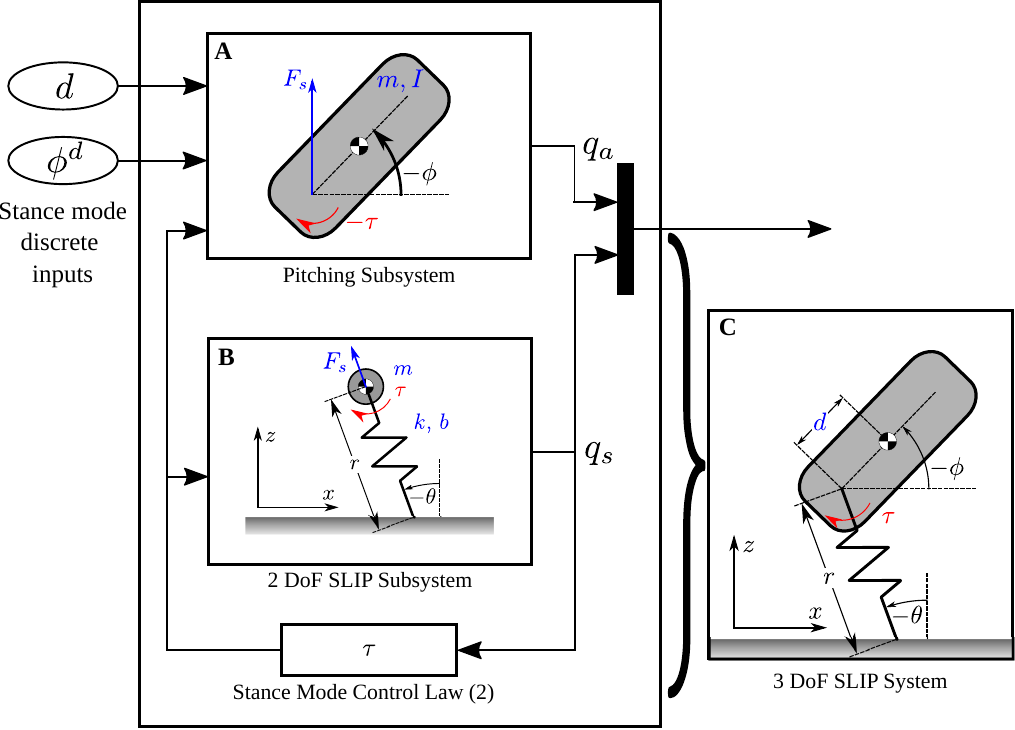}
% 	\valign{#\cr
% 		\hsize=0.3\linewidth
% 		\begin{subfigure}{\hsize}
% 			\centering
% 			\includegraphics[width=0.9\linewidth]{robots/SLIP}
% 			\caption{}
% 			\label{fig_SLIP}
% 		\end{subfigure}\vfill\vspace*{2em}{\Huge$\bm{\times}$}\vspace*{2em}
% 		\begin{subfigure}{\hsize}
% 			\centering
% 			\includegraphics[width=0.6\linewidth]{robots/pitch_subsystem}
% 			\caption{}
% 			\label{fig_pitch}
% 		\end{subfigure}\cr
% 		\hsize=0.15\linewidth
% 		\vfill\hspace*{1.5em}{\Huge$\bm{=}$}\hspace*{1.5em}\vfill\cr
% 		\hsize=0.4\linewidth
% 		\vfill
% 		\begin{subfigure}[][][c]{\hsize}
% 			\centering
% 			\includegraphics[width=\linewidth]{robots/SLIP_attitude}
% 			\caption{}
% 			\label{fig_ASLIP}
% 		\end{subfigure}\vfill\cr
% }
    \caption{The stance mode dynamics of the pitch unlocked 3 DoF SLIP model \textbf{(C)} viewed as the cascade composition of an isolated closed loop pitch subsystem \textbf{(A)} studied in \cref{sec_attitude} disturbing the translational dynamics of a 2 DoF SLIP subsystem \textbf{(B)} studied in \cref{sec_slip}. The blue symbols denote model parameters, the red symbol denotes the sole available continuous time stance control input  ($\tau$),  , and the black symbols denote state variables or functions of state variables. \Cref{fig_hybrid_block_diagram} contains a block diagram for the hybrid control strategy.}
    \label{fig_robot_diagrams}
    \setcounter{subfigure}{0}
    \refstepcounter{subfigure}\label{fig_pitch}
    \refstepcounter{subfigure}\label{fig_SLIP}
    \refstepcounter{subfigure}\label{fig_ASLIP}
%     \begin{minipage}{0.3\textwidth}\refstepcounter{subfigure}\label{fig_SLIP}\end{minipage}
% 	\begin{minipage}{\textwidth}\refstepcounter{subfigure}\label{fig_pitch}\end{minipage}
% 	\begin{minipage}{\textwidth}\refstepcounter{subfigure}\label{fig_ASLIP}\end{minipage}
\end{figure*}
\endgroup

In this section, we analyze the dynamics of pitch unlocked SLIP subject to our novel controller (\cref{table:controllers_math}). 
In order to facilitate analysis of the salient interactions between the various degrees of freedom, the assumptions in \cref{table_decomp_assumptions} decompose pitch unlocked SLIP (\cref{sec_decomp_slip}) into an independent attitude subsystem (\cref{sec_attitude}) whose output cascades to perturb the translational 2 DoF  SLIP subsystem (\cref{sec_slip}). \Cref{lemma_cascade_stability} summarizes the overarching consequences of these subordinate  results (as described more intuitively in \cref{sec_pitch_unlocked_slip}), formalizing the manner in which the analysis of the constituent  subsystems (\cref{lemma_slip_pitch,lemma_slip_stable}) yields  explicit fixed points \cref{eq_3dof_slip_fixed} and stability guarantees for the complete 3 DoF  system \endnote{\DIFadd{We remind the reader that this theory guarantees, for hyperbolic attractors of the kind achieved in \cref{lemma_cascade_stability}, that the local behavior of the actual nonlinear system will be smoothly conjugate to (i.e. enjoy the same transient properties) as the linearized dynamics (in a sufficiently small neighborhood).}}.

We model the robot's stance mode dynamics as SLIP \citep{Saranli_Schwind_Koditschek_1998} with attitude (\cref{fig_ASLIP}). In addition to the shank spring, we include radial damping to account for friction and other dissipative forces as well as a hip motor that adds energy to the system to counteract the energetic losses due to damping. The inputs to the system are the center of mass offset $d$, the target pitch $\phi^d$, and a target energy ratio $\psi_e^d$ \cref{eq_hybrid_vars}. Since not every robot has an internal degree of freedom to control $d$, we define a new input $d_x$ combining $\phi^d$ and $d$ in
\begin{equation*}
    d_x \coloneqq d \cos \phi^d \, 
\end{equation*}
as the steady state $x$ projection of the center of mass offset.

By varying the disturbance on pitch from the COM offset through adjusting $d_x$ \cref{eq_3dof_slip_fixed}, the pitch controller \cref{eq_torque,eq_discrete_torque} affects how much energy the hip torque is injecting into the system. 
This can be understood most clearly through \cref{eq_discrete_torque,eq_ddphi}: if the pitch deviates from its desired value due to $d_x$, then the energizing hip torque $\bar\tau$ will change to account for that disturbance.
As a result, the hip torque both energizes the translational SLIP subsystem and stabilizes the pitching subsystem system.
It is convenient to treat $d_x$ as a user-selected
input to the system since it affords direct  control authority over steady state energy. Independent control of steady state pitch $\phi^*$ is impossible without the further affordance of an additional  internally adjusted  DoF which is present in  Jerboa (\cref{fig_jerboa}) but not the planar biped (\cref{fig_planr_biped}).  
The stepping controller \cref{eq_stepping}, stabilizes the energy ratio $\psi_e$ \cref{eq_hybrid_vars}, which measures the distribution of the energy between the $\theta$ and $r$ subsystems of the 2 DoF SLIP subsystem. At touchdown, the stepping controller \cref{eq_stepping} moves energy from the hip torque energized angular subsystem to the decaying $r$ subsystem in order to prevent $\dot \theta$ from increasing in an unbounded fashion.

% By breaking the hip torque into a discrete time integrator in \cref{eq_discrete_torque} and a continuous time PD controller in \cref{eq_torque}, the controller uses the discrete time torque to predict what the increase in $\psi_e$ will be over the next step and reset $\psi_e$ using the stepping controller in \cref{eq_stepping}. Thus, although the controller cannot control $\theta_\text{lo}$, it has control over energy and $\psi_e$, the latter of which allows a trade off between apex height and forward velocity during locomotion.

\subsection{Decomposition of Pitch Unlocked SLIP}\label{sec_decomp_slip}
In this section we decompose the 3 DoF (pitch unlocked)  SLIP with attitude into an independent 1 DoF pitching subsystem whose dynamics perturbs the 2
DoF translational SLIP subsystem via the actuated hip torque that couples them. The hip torque stabilizes the pitching subsystem while energizing the SLIP subsystem.

\subsubsection{Decomposition Assumptions of Table 3}
\label{sec_slip_attitude_assumptions}

Starting with the Lagrangian model of the 3 DoF SLIP (\cref{fig_ASLIP}), we make the assumptions listed in \cref{table_decomp_assumptions} to decompose the system. 
\Cref{ass_1,ass_2,ass_3} serve to limit the effect of the pitching dynamics on the SLIP subsystem to influences only arising from the hip torque.  \Cref{ass_time} prevents the SLIP subsystem from coupling into the pitch subsystem via the state dependent guards of the translational SLIP subsystem.  \Cref{ass_pitch} prevents the SLIP state from coupling into the open loop pitching dynamics in a manner detailed in  \cref{sec_spring_force_just}.   Intuitively, the parameter $\chi$ stands in for influence on angular momentum about the toe contributed by the hip-displaced mass center. Specifically,  compensating for the elimination of certain coupling torques near steady state conditions,   $\chi$ abstracts the physical observation that on a steady state trajectory, this contribution is typically decreasing. Finally \cref{ass_taur,ass_xi_theta_replacement} prevent the SLIP state from affecting the pitching dynamics via the torque control law. 
Specifically, \Cref{ass_xi_theta_replacement} is a small angle approximation \citep{geyer_spring-mass_2005} about $\theta = \Xi_\theta$ which allows us to incorporate some of gravity's effect due to the slight asymmetry about $\theta = 0$ in our system near its fixed point.\endnote{See \cref{fig_sim_time_theta} for an example plot of the leg angle in simulation where $\theta$ goes from $\sim -0.2$ to $\sim 0.4$ rad in stance and \cref{fig_hard_time_theta} for the equivalent hardware plot where $\theta$ goes from $\sim0.45$ to $\sim0.5$ rad in stance. 
Because it is generated by the exact Lagrangian dynamics \cref{eq_slip_lagragian_dyn},  the simulation data is more relevant than the empirical data for assessing the efficacy of our analytical model (here, specifically, the validity of \cref{ass_xi_theta_replacement}) in capturing the behavior of the first principles physical phenomena (here, specifically, gravity's influence consequent upon the asymmetric steady state). For a more complete discussion of the relationships between  the analytical model, the numerical simulation, and the physical behavior of the hardware, please see \cref{sec_hard_sim_diff}. Historically small angle approximations in various forms have proven quite useful and general in the study and deployment of SLIP models in various robots \citep{geyer_spring-mass_2005,ankarali_stride--stride_2010,de_vertical_2018}.}
We use $\Xi_{\theta,g}$ as defined in \cref{eq_xi_theta_g} to avoid a transcendental implicit function in the fixed point computation of \cref{eq_av_dyn_slip} when $\bar\tau$ is a function of $a^*$ \cref{eq_pitch_fixed_points} and to preserve a cascade composition.

\begin{table*}[ht]
\centering  
\small\sf%\centering
\caption{Decomposition assumptions applied to continuous time stance dynamics. A $^\dagger$ denotes an assumption that is only applied to the pitching dynamics \cref{eq_ddphi_complicated}} \label{table_decomp_assumptions}
\begin{tabular}{N
>{\raggedright\arraybackslash}p{4.5cm}
l
>{\raggedright\arraybackslash}p{4.5cm}
>{\raggedright\arraybackslash}p{2.5cm}}
\toprule
\multicolumn{1}{c}{\textbf{Assumption \#}}  & \textbf{Formula} & \textbf{First use}  & \textbf{Justification} & \textbf{Consequence}
                 \\ 
\midrule
 \label{ass_1}  & $d^2 \approx 0$ & \cref{eq_slip_lagragian_dyn}& $d \approx \SI{0.06}{m}  \ll r \approx \SI{0.2}{m}$ & \cref{eq_slip_att_ass_dyn}\\
\midrule
 \label{ass_2}  & $\tau d \approx 0$      &  \cref{eq_slip_lagragian_dyn} & Small torque \cref{eq_torque} \endnote{The torque is small due to the PD terms being small, the effect of gravity being small, and $\bar\tau r$ being somewhat small on the trajectories of interest \cref{fig_sim_time}} and small $d$& \cref{eq_slip_att_ass_dyn}  \\
\midrule
 \label{ass_3}  & $\dot \phi^2 \approx 0$ & \cref{eq_slip_lagragian_dyn} & $\dot \phi$ starts small, has minimal perturbations, and vanishes along the steady state trajectory & \cref{eq_slip_att_ass_dyn} \\
\midrule
 \label{ass_time}& Time of flight, $T_f$, and time of stance, $T_s$, are fixed     & \cref{eq_pitch_dyn} & Longstanding empirical observation from \cite{raibert_legged_1986,de_vertical_2018} &  \cref{eq_pitch_dyn}\\
 \midrule
  \label{ass_pitch}$^\dagger$ & $d \cos(\theta - \phi) F_s \approx d_x \chi m g$ for $d_x \coloneqq d \cos \phi^d$ and constant $\chi$& \cref{eq_ddphi_complicated}  & See \cref{sec_spring_force_just} &  \cref{eq_ddphi}\\
%  \midrule
%  \label{ass_dx}&  $d \cos(\theta - \phi) \approx d_x \coloneqq d \cos \phi^d$ & \cref{eq_slip_lagragian_dyn} &  $\phi$ is close to $\phi^d$ and $\theta$ is close to $0$ \endnote{The error in the effect of the center of mass offset, $d$, on the steady state hip torque introduced by assuming $\theta \approx 0$ while the leg angle trajectories are asymmetric is offset by the free parameter $\chi$ which always multiplies $d_x$ and is chosen empirically.}. &  \cref{eq_ddphi} \\
% \midrule
% \label{ass_spring_replacement} $^\dagger$& $F_s \approx \chi m g$, where $F_s$ is the force in the leg \cref{eq:fs} and $\chi$ is a constant& \cref{eq_ddphi_complicated}& See \cref{sec_spring_force_just}  &  \cref{eq_ddphi} \\
\midrule
  \label{ass_taur}$^\dagger$&  $\bar\tau r \approx \bar\tau r_0$     & \cref{eq_torque}&  Small deflection ($\approx$ 0.1 leg lengths) due to spring stiffness and limited range of motion  \endnote{This assumption,  the one of two (the other being \cref{ass_small_odd}) in the paper  imposing a direct restriction on the magnitude of a state component, reflects the  physical design of the Jerboa platform \citep{shamsah_analytically-guided_2018}  whose range of motion is mechanically limited to roughly 20\% of the robot's leg length. } & \cref{eq_torque_simple} \\
\midrule
 \label{ass_xi_theta_replacement}$^\dagger$& $g \sin\theta \approx g \sin \Xi_{\theta,g}$, where $\Xi_{\theta,g}$ is the value of $\Xi_\theta(\bm a^*)$ assuming gravity acts radially    & \cref{eq_torque} & Approximation of gravity's influence arising from  asymmetric steady state stance (e.g., \cref{fig_sim_time_theta}) &   \cref{eq_torque_simple} \\
\bottomrule
\end{tabular}
\end{table*}

\subsubsection{Simplified Stance Dynamics} \label{sec_slip_attitude_simplified_dyn}
Applying the assumptions from \cref{table_decomp_assumptions} to the full dynamics in \cref{eq_slip_lagragian_dyn}, we have a 2 DoF translation SLIP subsystem \cref{eq_ddr,eq_ddtheta} and a 1 DoF pitching subsystem \cref{eq_ddphi} coupled via the control input $\tau$. The stance dynamics are
\begin{subequations}\label{eq_slip_att_ass_dyn}
\begin{align}
    \Ddot{r} &= \omega_r^2 (r_0 - r) - \frac{b \dot{r}}{m} + r \dot\theta^2  - g \cos \theta \label{eq_ddr}\\
    \Ddot{\theta} &= \frac{\tau}{m r^2} + \frac{g \sin \theta}{r} - \frac{2 \dot r \dot \theta}{r} \label{eq_ddtheta} \\
    \ddot{\phi} &= \frac{d_x \chi m g}{I} - \frac{\tau}{I}\label{eq_ddphi}
\end{align}
\end{subequations}
where $\omega_r \coloneqq \sqrt{k/m}$ is the natural frequency of the spring mass system. The simplified pitching dynamics \cref{eq_ddphi} expose  the hip torque's large control authority over the
orientation, revealing that it must counteract a disturbance from
the center of mass offset.

\subsubsection{A Pitch Stabilizing and CoM Energizing Hip Torque Law}
The goal of the torque is to stabilize the pitching system while also energizing the 2 DoF SLIP subsystem. To this end, the stance phase continuous time hip torque control policy takes the form
\begin{equation}
    \tau\coloneqq \bar\tau r - m g r \sin \theta - \epsilon k_p (\phi^d - \phi) + \sqrt \epsilon k_d \dot\phi \label{eq_torque}\,.
\end{equation}
This strategy incorporates a feedforward bias term  $\bar\tau$ whose value is held constant in stance but updated each succeeding aerial mode as a function of the pitch error at the liftoff event.
This hybrid discrete event feedback control is intended to eventually cancel $d_x \chi m g$, the disturbance on the pitching system from the radial dynamics of \cref{eq_ddr} (whose continuous variation during stance we choose to neglect in favor of the constant moment $d_x \chi mg$ for  reasons introduced in the commentary on \cref{ass_pitch} in \cref{sec_slip_attitude_assumptions} and detailed in \cref{sec_spring_force_just}).
 The second term is a continuously varying within-stance feedforward expression intended to cancel the effect of gravity on the translational angular dynamics \cref{eq_ddtheta}. The final two terms are a conventional continuously varying within-stance proportional derivative controller intended to regulate pitch $\phi$ to a target $\phi^d$ and the angular velocity $\dot\phi$ to $0$.
By including a gravity compensation term in \cref{eq_torque}, we transform the effect of gravity on \cref{eq_ddtheta} into a disturbance on \cref{eq_ddphi}. Leveraging \cref{ass_taur,ass_xi_theta_replacement} the continuous time hip torque control policy for the pitching subsystem is
\begin{align}
        \tau = \bar\tau r_0 - m g r_0 \sin \Xi_{\theta,g} - \epsilon k_p (\phi^d - \phi) + \sqrt \epsilon k_d \dot\phi \,, \label{eq_torque_simple}
\end{align}
thereby decoupling the pitch subsystem from the influence of the translational dynamics in the continuous time stance mode  (for the purposes of analysis of the pitch subsystem, albeit not in implementation). For the implementation we use \cref{eq_torque}.
 
The discrete stride event torque bias update law is
\begin{align}
    \bar\tau_{k+1} = \bar\tau_{k} + \epsilon^2 k_\tau (\phi_{\text{lo},k} - \phi^d) \label{eq_discrete_torque}\,,
\end{align}
which is calculated at liftoff, and the new value of $\bar\tau$ (adjusted by the discrepancy between the attitude at the current liftoff $\phi_{\text{lo},k}$ and the desired attitude $\phi^d$) is applied during the next stance. In stance, the proportional derivative (PD) terms in \cref{eq_torque} quickly bring the pitching subsystem to its equilibrium point, which is offset from $\phi^d$ due to the constant disturbance from $d_x \chi m g$ and $\bar\tau$ in \cref{eq_ddphi}.
The discrete time integrator \cref{eq_discrete_torque} is measuring the steady state error of the continuous time PD loop in \cref{eq_torque}. As a result, the integrator drives the stance dynamics to a trajectory where, on average, $\bar\tau$ cancels the pitch disturbance due to the force in the leg resulting in zero steady state error. In turn $\bar \tau$ will energize the 2 DoF translational SLIP subsystem.

\subsubsection{Cascade Structure}

Under the control law in \cref{eq_torque,eq_discrete_torque}, the hybrid dynamics of the pitching subsystem do not depend on the 2 DoF SLIP subsystem while the 2 DoF SLIP subsystem is perturbed by the pitching subsystem leading to a cascade composition \citep{Sontag_2008}. We denote the translational and orientation components of the system, respectively, as
\begin{subequations}
\begin{align}
    \zs & \coloneqq [r,\theta, \dot{r}, \dot{\theta}]^T  \in \mathcal{S} \coloneqq T ( \mathbb{R}_+ \times \mathbb{S}^1) \label{eq_slip_state}\\
    \zp& \coloneqq [\phi, \dot{\phi}, \bar\tau]^T   \in \mathcal{A} \coloneqq T \mathbb{S}^1 \times \mathbb{R} \label{eq_pitch_state}
\end{align}
\end{subequations}
where $\zs$ 
is the state of the 2 DoF SLIP subsystem in stance,\endnote{
We work in polar coordinates for the translational system, whereby $r \in \mathbb{R}_+$, the set of positive real numbers, 
$\theta  \in \mathbb{S}^1$, the angles on the unit circle, and $T ( \cdot )$ denotes the tangent space. 
} and $\zp$ is the state of the attitude (pitching) subsystem, augmented by the hybrid discrete event torque offset value $\bar\tau$, which will be varied from stride to stride according to \cref{eq_discrete_torque}.  In this manner, by a slight abuse of notation,  we conflate the continuous time states of the Lagrangian stance mode dynamics $\zs$ and  $\zp$ with their discrete time values sampled at the liftoff event --- introducing a subscripted time stamp, $k \in \mathbb{N}$ when the distinction is not immediately clear from the context.

We denote by 
\begin{equation}
    \Re : \mathcal{S} \times \mathcal{A} \rightarrow \mathcal{S} \times \mathcal{A} \, , \label{eq_3dof_return}
\end{equation}
the exact closed loop return map expressed in liftoff coordinates of the
complete 3 DoF SLIP with attitude depicted in \cref{fig_ASLIP}. The formal specification of this hybrid  dynamical system is given by stance dynamics \cref{eq_slip_att_ass_dyn}, subject to the closed loop feedback \cref{eq_torque} in stance, guarded by the liftoff condition ($r=r_0$ and $\dot r > 0$), followed by a reset map determined in part by hybrid hip torque adjustment \cref{eq_discrete_torque} and in part by ballistic flight interrupted by the stepping strategy \cref{eq_stepping}. 

The cascade structure of this map arising from  the assumptions of  \cref{table_decomp_assumptions} takes the form $\Re( \zs, \zp) = [ \Se(\zs, \zp), \P(\zp)]^T$, where 
\begin{equation}
    \P : \mathcal{A} \rightarrow \mathcal{A}\label{eq_pitch_return}
\end{equation} 
denotes  the isolated pitch return map governing the body attitude \cref{eq_pitch_return_full}  (depicted in \cref{fig_pitch}) to be derived in \cref{sec_attitude}, and 
\begin{equation}
    \Se :  \mathcal{S} \times \mathcal{A} \rightarrow \mathcal{S}\, \label{eq_slip_perturbed_map}
\end{equation} 
denotes the translational component of the exact return map as perturbed by the pitching states.
For purposes of analysis, we will be concerned with the unperturbed form of this map 
\begin{equation}
    \Sep :  \mathcal{S}  \rightarrow \mathcal{S} : \zs \mapsto \Se(\zs, \zp^*) \, ,\label{eq_slip_unpertubed_map}
\end{equation}
 resulting from the convergence of the (isolated, asymptotically stable) pitching subsystem to its steady state $\zp^* = [ \phi^*, 0, \bar\tau^*]^T \in \mathcal{A}$  \cref{eq_pitch_fixed_points}.
Finally, after applying a further set of assumptions  governing the translational dynamics listed in \cref{table_hybrid_averaging_assumptions}, we will derive and study carefully in \cref{sec_slip} the  map
\begin{equation}
    \Shp :  \mathcal{S}  \rightarrow \mathcal{S} \, ,\label{eq_slip_avg_return}
\end{equation}  
an approximant of the actual unperturbed translational return map $\Sep$ resulting from the hybrid averaging analysis of \cref{sec_hybrid_averaging}, whose local stability properties determine those of $\Sep$.  

Initially when the pitching subsystem is away from its fixed point, it perturbs the 2 DoF SLIP subsystem via the discrete time changes of $\bar\tau$ \cref{eq_discrete_torque} and through some small continuous time influences from the PD terms in the hip torque control law \cref{eq_torque}. Once the pitching subsystem converges to its fixed point \cref{eq_pitch_fixed_points}, the perturbations die out leaving just a constant $\bar\tau$ which energizes the 2 DoF SLIP subsystem (\cref{sec_slip_dyn}).

\subsubsection{Fixed Points and Stability}
\cref{lemma_cascade_stability} presents the overall logic of our formal result, summarizing our analysis of the full 3 DoF SLIP system by first finding fixed points and stability for the pitching subsystem before studying the unperturbed 2 DoF SLIP subsystem and, finally, reasoning about the cascade interconnection dynamics of the former considered as a transient perturbation into the latter.

\begin{theorem}[Stability of 3 DoF Pitch-Unlocked SLIP] \label{lemma_cascade_stability}
Under assumptions in \cref{table_decomp_assumptions} imposed upon the 3-DoF SLIP with attitude system depicted in \cref{fig_ASLIP} and the assumptions in \cref{table_hybrid_averaging_assumptions,table_assumption_reset} imposed upon the translational subsystem depicted in \cref{fig_SLIP}, the closed loop liftoff return map of the 3 DoF SLIP with attitude $\Re$, \cref{eq_3dof_return} has fixed points $\q^* = [\zp^*, \zs^*]$ \cref{eq_3dof_slip_fixed}. 
%for the range of set points $ (\dot{x}_{apex}^d, \y_{apex}^d, \phi^d)  \in \mathcal{A}$ delimited by \ref{eq:setpointlimits} in \cref{sec_invert_fixed_points}.
Moreover, further assuming \cref{conj:bibostablelimitcycle} and the algebraic bounds on controller parameters posited in  \cref{lemma_slip_pitch,lemma_av_stability}, there is some $\epsilon_0 >0$ such that for all $0<\epsilon <\epsilon_0$, these fixed points are asymptotically stable.
\end{theorem}

\begin{proof}
The assumptions in \cref{table_decomp_assumptions} imply that the pitching subsystem is isolated from the translational subsystem,
hence according to \cref{lemma_slip_pitch},
$\P$ has an exponentially stable fixed point at $\zp^*$ \cref{eq_pitch_fixed_points}.
Additionally according to \cref{lemma_slip_stable}, under the assumptions from \cref{table_hybrid_averaging_assumptions,table_assumption_reset},
the unperturbed translational subsystem $\Sep$ has a fixed point $\epsilon$ close the averaged fixed point $\hat \x^*$ \cref{eq_x_star}, and is exponentially stable for all $0 < \epsilon < \epsilon_0$ for some $\epsilon_0>0$ and $\epsilon k_e, \epsilon \alpha \in (0,1)$. Leveraging the changes of coordinates defined in \cref{eq_hybrid_vars,eq_g_def,eq_state_phasediff}, $\hat \x^*$ can be lifted to $\hat \zs^*$ \cref{eq_qs_star} which is $\epsilon$-close to $\zs^*$ \cref{eq_unav_slip_fixed}.

According to \cref{conj:bibostablelimitcycle}, it also follows from the exponential stability of $\Shp$ that the unperturbed limit cycle generating the return map $\Sep$ is locally exponentially stable in continuous time, hence has  bounded-input-bounded-output (BIBO) stability with respect to continuous time disturbances which are small enough to keep the state of the translational subsystem within the basin of attraction of its linearized dynamics.  Thus for sufficiently small initial conditions, the exponentially decaying disturbing inputs from the pitching subsystem excite exponentially decaying  continuous time responses in the translational subsystem (due to its BIBO stability), and both subsystems' hybrid discrete event states exponentially approach the fixed point. \qed
\end{proof}

% \begin{figure}[thb]
%     \centering
%     \includegraphics[width=0.8\linewidth]{figs/FormalLogic.pdf}
%     \caption{Conceptual diagram showing formal dependencies yielding \cref{lemma_cascade_stability} \SRL{Dan, thoughts on cutting this figure from the paper now that our logic has been cleaned up? Alternatively this could go in an appendix since most readers will not care about it.}}
%     \label{fig_formal_logic_block_diagram}
% \end{figure}%

\subsection{Pitching Subsystem}\label{sec_attitude}
Due to the assumptions in \cref{table_decomp_assumptions}, the pitching subsystem (consisting of pitch and hip torque integrator) incurs no influence from the translational (SLIP) subsystem and takes the form of a simple time switched hybrid dynamical system with two modes  characterized, respectively, by ballistic flight and PD-controlled stance dynamics. 

\subsubsection{Dynamics}
\label{sec_pitch_dyn}
In order to account for the  integrator introduced by the discrete time hybrid stride event torque control law \cref{eq_discrete_torque}, it is convenient to write  the closed loop pitch   subsystem  dynamics affected by the hip torque command \cref{eq_torque} in the state $\zp =[ \phi, \dot \phi, \bar\tau]^T$ \cref{eq_pitch_state} as
\begin{subequations}\label{eq_pitch_dyn}
    \begin{align} 
    I \ddot{\phi} &=
    \begin{cases}\!% alignment adjustment
  \begin{aligned}[c]% adjust case condition placement here, use [t]op, [b]ottom, or [c]enter (default)
        \frac{d_x F_s}{I} &- \bar\tau r0 - m g r_0 \sin \Xi_{\theta,g}  \\ 
         & + \epsilon k_p (\phi^d - \phi) - \sqrt \epsilon k_d \dot\phi,
      \end{aligned} & \scriptstyle T_f \leq t \leq T_f + T_s \\
 0,&\scriptstyle 0 \leq t \leq T_f    
 \end{cases}\\
    %     \begin{cases}
    %  0 & 0 \leq t \leq T_f \\
    %  \frac{d_x F_s}{I} - \frac{\bar\tau r0 - m g r_0 \sin \Xi_{\theta,g}}{I} - \frac{-\epsilon k_p (\phi^d - \phi) + \sqrt \epsilon k_d \dot\phi}{I} & T_f \leq t \leq T_f + T_s 
    % \end{cases}\\
    \dot{\bar{\tau}} &= 0
\end{align}
\end{subequations}
where $T_f$ and $T_s$, the time of flight and stance constants, respectively, introduced in \cref{ass_time} determine the stride period $T=T_s + T_f$ (which we identify with $t=0$ for purposes of the continuous time stance mode dynamics), triggering the reset
\begin{align}  \label{eq_pitch_reset}
    \vctthree{\phi}{\dot \phi }{\bar \tau} \mapsto \vctthree {\phi}{\dot \phi}{\bar\tau + \epsilon^2 k_\tau (\phi_\text{lo} - \phi^d)} 
\end{align}
arising from \cref{eq_discrete_torque} and the continuity of the pitch states $\phi$ and $\dot \phi$ across the liftoff event. Since the pitch dynamics \cref{eq_pitch_dyn} are linear time invariant and the pitch reset map \cref{eq_pitch_reset} is linear, the discrete time pitch return map \cref{eq_pitch_return} is a linear time invariant dynamical system.
\subsubsection{Fixed Points}
\label{pitch_fixed_points}
Fixed points of the
time switched pitching return map $\P$ can be directly calculated by finding a fixed point of the continuous time flight dynamics and stance dynamics \cref{eq_pitch_dyn} which is invariant under the reset map \cref{eq_pitch_reset}. A necessary condition for the invariance of a state $\zp^*$  under reset \cref{eq_pitch_reset} is that $\phi^* = \phi^d$. On the other hand, a sufficient condition for $\zp^*$ to be an equilibrium state of both  continuous dynamics modes is that  both $\dot \phi^* = 0$ and $\ddot \phi \mid_{\zp^*} = 0 $. Applying the necessary reset condition to the sufficient equilibrium condition 
now yields a sufficient hybrid dynamics fixed point condition taking the form 
of an implicit function in $\bar\tau$ arising from the requirement $\ddot \phi = 0$ in stance mode of \cref{eq_pitch_dyn}.  Solving this implicit function for $\bar\tau$ yields the fixed point
\begin{align}
    \label{eq_pitch_fixed_points}
    \zp^* = \vctthree{\phi^*}{\dot\phi^*}{\bar\tau^*} = \vctthree{\phi^d}{0}{\frac{m g}{r_0} \left(\chi d_x - r_0 \sin \left(\Xi_{\theta,g}\right)\right)}
\end{align}
of the complete hybrid (time switched) pitch subsystem $\P$.

\subsubsection{Pitching Subsystem Stability} \label{sec_pitch_stable}
 Since the discrete time return map $\P$ takes the form of  an LTI system, determining sufficient conditions for asymptotic stability is straightforward.
\begin{proposition}[Pitch Subsystem Stability] \label{lemma_slip_pitch}
The hybrid dynamical pitching
subsystem \cref{eq_pitch_dyn} detailed in \cref{sec_pitch_dyn} is stable for $0<\frac{\epsilon k_p T_s T_f}{I}<4$,  $\sqrt \epsilon k_d =2 \sqrt{\epsilon k_p I}$, and $0 < \epsilon^2 k_\tau < \frac{2 \epsilon I k_\phi}{r_0 T_s \left(\sqrt{\epsilon k_p I} -T_f \kappa \right)}$  where $\kappa \coloneqq \sqrt{\frac{\epsilon k_\phi T_s}{T_f}}$ and $\epsilon k_\phi \coloneqq \frac{\epsilon k_p T_s T_f}{I}$.
\end{proposition}
\begin{proof}
See \cref{sec_pitch_proof}.
\end{proof}

The constraints on $k_p,\, k_d$, and $k_\tau$ in \cref{lemma_slip_pitch} are conservatively chosen to facilitate the analytical tractability of the stability analysis. 
As discussed in \cref{sec_detail},  practical implementation of the pitching controller
entails ensuring that the system is sufficiently damped to achieve a small angular
velocity at liftoff.  This, in turn, prevents a large error in pitch
from accumulating in flight wherein there is no direct affordance over
pitch.

\subsection{SLIP with Damping and Hip Actuation: Unperturbed Analysis}
\label{sec_slip}
In this section we analyze the dynamics and stability of the unperturbed 2-DoF SLIP subsystem (\cref{fig_SLIP}) and its hybrid return map $\Sep$.
Recall that this subsystem arises from an evaluation of  the full 3-DoF SLIP with attitude (\cref{fig_ASLIP}) at the fixed point of the
pitching subsystem (\cref{fig_pitch}).
The SLIP control policy replenishes the total translational energy of the 2 DoF CoM lost in stance from damping with the hip torque \cref{eq_torque} while regulating the ratio of its energies with a new stepping feedback law \cref{eq_stepping}.

While the pitching subsystems dynamics \cref{eq_pitch_dyn} are linear time invariant, the 2-Dof SLIP subsystem includes nonlinear hybrid dynamics which include non-integrable stance dynamics \citep{Ghigliazza_Altendorfer_Holmes_Koditschek_2003}, precluding an exact expression for even the return map  (much less its fixed points and conditions for their stability).  Instead, in \cref{prop_slip_hybrid} we use hybrid averaging (\cref{sec_hybrid_averaging}) to obtain closed form expressions that approximate its fixed points and to derive (similarly computationally effective) sufficient conditions for their stability. 

The stability proof in \cref{prop_slip_hybrid} takes advantage of a novel set of coordinates presented in \cref{sec_coordinates} which treats SLIP not as a parallel composition 
\citep{raibert_legged_1986,de_averaged_2018} but instead as a unitary 2 DoF system. 
The assumptions in \cref{table_hybrid_averaging_assumptions}, reviewed in \cref{sec_slip_ass},  enable  computation of the averaged stance dynamics and fixed points in \cref{sec_slip_dyn}. Analysis of the hybrid structure in \cref{sec_slip_reset} implies not only that a hip energized gait requires an asymmetric stepping strategy as demonstrated in \cref{theorem_asymmetry}, but also facilitates the construction of our stepping strategy \cref{eq_stepping} which regulates the ratio of the SLIP subsystems energies. 
Finally,  \cref{cor.ResetFixedPoint}  relates the  equilibrium state
of the averaged stance dynamics to the fixed point
of the averaged SLIP return map  $\Shp$  leading the way to 
the stability conditions of \cref{lemma_av_stability}.

% \begin{lemma} [Stability of Unperturbed Translational SLIP Subsystem] \label{lemma_slip_stable}
%     Under the assumption in \cref{table_hybrid_averaging_assumptions,table_assumption_reset} and the controller's presented in \cref{table:controllers}, the unperturbed translational SLIP subsystem \cref{fig_SLIP}, $\Sep$ is stable for all $0 < \epsilon < \epsilon_0$ for some $\epsilon_0>0$ and $\epsilon k_e, \epsilon \alpha \in (0,1)$ with a fixed point $\epsilon$ close to $\x^*$ \cref{eq_x_star}.
% \end{lemma}
% \begin{proof}
%     According to \cref{prop_slip_hybrid} we can apply \cite{de_hybrid_2018} Theorem 2. Thus the unperturbed 2-DoF SLIP subsystem, $\Sep$, has the same stability type as the averaged unperturbed 2-Dof SLIP subsystem,  $\Shp$, for all $0 < \epsilon < \epsilon_0$ for some $\epsilon_0>0$,. According to \cref{lemma_av_stability},  $\Shp$ is stable under the conditions of this lemma, thus $\Sep$ is stable. \qed
% \end{proof}

\subsubsection{Hybrid Averaging}
 Hybrid averaging \citep{de_hybrid_2018} is a method of studying nonlinear hybrid dynamics systems and yields $\epsilon$-close fixed points of the Poincar\'{e} return map and stability guarantees without requiring integrable dynamics \citep{de_hybrid_2018}. Hybrid averaging divides a system into fast phase states with $\mathcal{O}(1)$ dynamics, and slow ``energy'' states with $\Oeps$ dynamics. The hybrid averaging theorem  \citep{de_hybrid_2018} applies to systems with one state playing the role of a phase variable and whose remaining ``energy'' states are invariant under the flight map. Such rigid requirements preclude the na\"{i}ve application of hybrid averaging to many systems of interest including SLIP.

\cite{de_modular_2017} presents an algorithm taking the form of a checklist of conditions that are, cumulatively, sufficient for applying hybrid averaging to systems with more than one phase that obey \cref{def_avik} \endnote{\Cref{def_avik} can be found in \cref{sec_hybrid_averaging} and in \cite{de_modular_2017}.}. De's insight is that on the closed orbit, all phases are locked with different relative frequencies and a constant set of phase offsets. De uses this to calculate a change of coordinates \cref{eq_state_phasediff} from many phases to one master phase whose resulting ancillary phase offsets can be conceptually reconceived as components of the ``energy” state. In \cref{sec_hybrid_averaging} we extend De's work to a similarly sufficient computational checklist (\cref{sec_recipe}), that can be applied to systems which satisfy his \cref{def_avik} (i) and (ii), but not \cref{def_avik} (iii).
However, rather than substituting a relaxed alternative
sufficient condition in our revision to the checklist of \cref{def_avik}, we instead arrive at a system that directly satisfies the hypotheses of \cite{de_hybrid_2018} Theorem 2.

\begin{proposition}[Stability of Unperturbed Translational SLIP Subsystem]
\label{prop_slip_hybrid}
\label{lemma_slip_stable}

Under the assumptions in \cref{table_hybrid_averaging_assumptions,table_assumption_reset} and the controllers presented in \cref{table:controllers_math}, there is some $\epsilon_0 > 0$ such that the unperturbed translational SLIP subsystem's return map $\Sep$  \cref{eq_slip_unpertubed_map} (depicted in \cref{fig_SLIP}) is stable for all $0 < \epsilon < \epsilon_0$ and $\epsilon k_e, \epsilon \alpha \in (0,1)$ at a fixed point $\epsilon$-close to the averaged fixed point $\hat\x^*$ \cref{eq_x_star}.
\end{proposition}
\begin{proof}
The result will follow by showing that the stated hypotheses suffice to satisfy the conditions of \cite{de_hybrid_2018} Theorem 2. To do so, first note that according to \cite{de_modular_2017} lemma 6, under the projection $\h \circ \g$ defined in \cref{sec_recipe,eq_g_def}, the system \cref{eq_slip_ass_dyn} is of the form of \cite{de_hybrid_2018} equation 12, whose averaged vector field, $\hat{f}$ \cref{eq_av_dyn_slip} has an equilibrium state, $\hat\x^*$, displayed in \cref{eq_x_star}. 
Next, \cref{cor.ResetFixedPoint} and the values of $\Xi_\theta$ \cref{eq_xi_theta}, $\Xi_e$ \cref{eq_xi_e} ensure that $R(\hat\x^*) = \hat\x^*$ \cref{eqn:Resetx}, establishing that the averaged hybrid system satisfies conditions (ii.a) and (ii.b) of \cite{de_hybrid_2018} Theorem 2. 
Moreover, \cref{prop_reset_jacobian} demonstrates that $D_x R_x(\x) = S_0  + \epsilon S_1(\x, \epsilon)$, where $S_0$ is constant and invertible, and its unity eigenvalues have diagonal Jordan blocks, satisfying condition (i) of \cite{de_hybrid_2018} Theorem 2. 

Finally the averaged return map is hyperbolic at $\hat \x^*$ according to \cref{lemma_av_stability} satisfying condition (ii.c) of \cite{de_hybrid_2018} Theorem 2. Thus, according to \cite{de_hybrid_2018} Theorem 2, the unperturbed 2-DoF SLIP subsystem's return map $\Sep$ has the same stability type as the averaged unperturbed 2-Dof SLIP subsystem $\Shp$. But \cref{lemma_av_stability} also proves under the conditions stated in the hypothesis of this proposition that $\Shp$ is asymptotically stable and the desired conclusion now follows. 
\qed
\end{proof}

\subsubsection{Choice of Coordinates for Asymmetric SLIP Stance Mode Dynamics}
\label{sec_coordinates}
The unperturbed SLIP subsystem $\Sep$ \cref{eq_slip_unpertubed_map} arises from an evaluation of  $\Se$ \cref{eq_slip_perturbed_map}   at the fixed point of the pitching subsystem \cref{eq_pitch_fixed_points}, whereby in stance the control \cref{eq_torque}  reduces to the expression
\begin{equation}
    \tau(t) = \bar\tau^* r - m g r \sin \theta \nonumber
\end{equation}
for the pitching subsystem's steady state torque $\bar\tau^*$ specified  by \cref{eq_pitch_fixed_points}. This control law imparts a constant force perpendicular to $r$ to add energy to the system while also cancelling the effect of gravity in \cref{eq_ddtheta}.

In shank energized SLIP, it is commonly assumed that the leg's angular velocity $\dot\theta$ and the ``energy" in the radial direction $a_r$ are roughly constant along the steady state hybrid dynamics trajectory corresponding to the fixed point of the return map \citep{de_penn_2015,raibert_legged_1986}. In hip energized SLIP, these assumptions are no longer valid since the hip torque accelerates $\theta$, and there is no shank actuator to counteract the energetic losses from damping on $a_r$. This inherent variation of values along the steady state stance mode trajectories precludes a direct application of the hybrid averaging ideas developed in \cite{de_hybrid_2018} since that theory requires that the “energy” have a fixed point in the averaged stance dynamics. 

This steady state asymmetry in both $a_r$ and $\dot{\theta}$, revealed to be formally necessary in \cref{theorem_asymmetry}, motivates us to track the total energy in the system $a_e$ and to promote  the angle of velocity introduced in  \cite{peuker_leg-adjustment_2012} and \cite{sharbafi_vbla_2016} as an output variable (fedback to
determine the touchdown angle) to the status of a system state.
 This new state $\psi_e$ that we call the ``energy ratio'' tracks the velocity perpendicular to the leg shank relative to twice the mass-specific total energy (yielding units of \si{m/s}), resulting in the following set of state variables under the (intuitively ``spherical'') change of coordinates
\begin{equation}\label{eq_hybrid_vars}
\z = 
\vctfour{\psi_r}{\psi_e}{\theta}{a_e} = \hs(\zs)
\coloneqq
\vctfour
{ \arctan \frac{\dot{r}}{(r_0-r) \omega_r}  }
{ \arctan \frac{\dot{\theta} r}{a_r}  }
{ \theta}
{ \sqrt{a_r^2 + \dot\theta^2 r^2}  }\,,
\end{equation}%
where 
\begin{equation}
        a_r \coloneqq \sqrt{{\dot{r}^2 + (r_0-r)^2\omega_r^2}}\,. \nonumber
\end{equation}
The map $\hs$ is invertible with inverse \cref{eq:change_inverse}.

\begin{figure}[thb]
    \centering
    \includegraphics[width=0.7\linewidth]{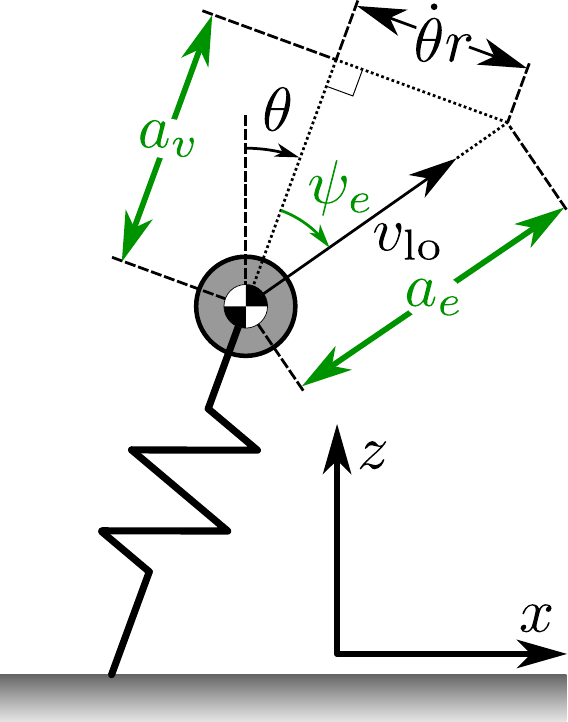}
    \caption{The variables from \eqref{eq_hybrid_vars} used in hybrid averaging displayed on a SLIP model at liftoff. The relevant variables are depicted in green. The comparison to the direction of the liftoff velocity $v_\text{lo}$ is only valid when $r=r_0$.}
    \label{fig_hybrid_variables}
\end{figure}%
A visual depiction of these variables (except $\psi_r$) at liftoff is presented in \cref{fig_hybrid_variables}. The definition of $\psi_r$, chosen shortly below to serve as the hybrid averageable phase variable following 
\cite{de_hybrid_2018}, accompanies the radial energy $a_r$ (in units of \si{m/s}) as the mechanical phase angle of the stance phase leg extension dynamics that has held a traditional place in the locomotion literature since Raibert's original insight into its key role as the ``heartbeat'' of a running limit cycle \citep{raibert_legged_1986,Koditschek_Buehler_1991}. In contrast, the energy ratio $\psi_e$  expresses the contribution to the total mechanical energy $a_e$ (in units of \si{m/s}) of the velocity perpendicular to the leg shank relative to the radial energy and can be usefully interpreted as a new, secondary phase variable. At liftoff, $a_e$ is the magnitude of the velocity of the robot, and $\psi_e + \theta$ is the angle of the velocity relative to vertical. By working in these coordinates, we view SLIP as a unitary 2 DoF system rather than the parallel composition of two 1 DoF systems \citep{de_parallel_2015}.

 A key step in applying our extension of De's hybrid averaging methods  \citep{de_hybrid_2018,de_modular_2017} to the 2 DoF model  is to shift the ancillary phase components of the ``spherical'' physical coordinates \cref{eq_hybrid_vars} via the translation
  \begin{align}
  \begin{split}
          \y \coloneqq [\sigma, \bm \psi, \bm a]^T &= [\psi_r, \tilde\psi_e, \psi_\theta, a_e]^T \\ 
          &=\g(\z) \coloneqq \z - [0, \Xi_e, \Xi_\theta, 0]^T, \label{eq_g_def}
  \end{split}
\end{align}
where $\sigma = \psi_r$ is the master phase, $\bm \psi$ is the vector of ancillary phases, $\bm a$ is the vector of energy components, and  $\Xi_e$ \cref{eq_xi_e} and $\Xi_\theta$ \cref{eq_xi_theta} are constants parameterized by the energy component of the fixed point and whose purposes are discussed in \cref{sec_phase_offset}.
Adopting the foregoing terminology and notation for the various components of the new state representation \cref{eq_g_def} aligns the present discussion with the nomenclature and analysis of \cite{de_modular_2017}. This alignment enables the use of a final set of projected coordinates introduced in \cite{de_modular_2017} that are transverse to the eventual hybrid limit cycle and whose dynamics will take the form of \cite{de_hybrid_2018} equation 12,
\begin{align}
        \x \coloneqq [\bm \delta, \bm a]^T &= \h(\bm y) = \begin{bmatrix}\sigma \bm 1 - \Delta(\upomega)\bm \psi \\
        \bm a\end{bmatrix} \label{eq_state_phasediff}
\end{align} 
 where $\bm \delta$ is a vector of phase offsets and $\omega$, defined in \cref{eq_w}, is the vector of relative frequencies between the master phases and the ancillary phases.

% Since the definition of $\omega_1$ \cref{eq_w1} ensures that $\delta = 0$ is a fixed point of averaged $\delta$ dynamics, $\Pi \delta \bar f$, then $\Xi_e$ and $\Xi_\theta$ move the fixed points of the ancillary phases relative to the master phase independent of the section's coordinates by changing the section definition. As a result we use $\Xi_e$ and $\Xi_\theta$ to place the fixed point of the ancillary phases such that $R(\bm{x}^*) = \bm{x}^*$ as shown in \cref{sec_phase_offset}. 

\subsubsection{Assumptions Underlying Hybrid Averageability of SLIP Stance Dynamics}
\label{sec_slip_ass}

\Cref{table_hybrid_averaging_assumptions} introduces a set of assumptions sufficient for restricting the dynamics to a form fit for hybrid averaging.

\begin{table*}[ht]
\centering
\small\sf%\centering
\caption{Assumptions underlying hybrid averaging of stance dynamics} \label{table_hybrid_averaging_assumptions}
\begin{tabular}{N
>{\raggedright\arraybackslash}p{4.5cm}
l
>{\raggedright\arraybackslash}p{4.5cm}
>{\raggedright\arraybackslash}p{2.5cm}}
\toprule
\multicolumn{1}{c}{\textbf{Assumption \#}}  & \textbf{Formula} & \textbf{First use}  & \textbf{Justification} & \textbf{Consequence}
                 \\ 
\midrule
   \label{ass_psi_e_replacemt}& $\cos \psi_e \approx \cos \gamma \psi_e^d$, $\sin \psi_e \approx \sin \gamma \psi_e^d$ for fixed $0 < \gamma < 1$ \endnote{
   \Cref{tab_model_parameters} contains the empirically determined value for gamma, which was obtained using the method described in \cref{sec:model_params}.} & \cref{eq_slip_dyn} & Small angle for $\psi_e$ about $\gamma \psi_e^d$ see \cref{fig_sim_time_phase}. & \cref{eq_slip_ass_dyn} \\
   \midrule
   \label{ass_theta_replacement}&  $\cos \theta \approx \cos \Xi_\theta$ for $\Xi_\theta$ defined in \cref{eq_xi_theta} & \cref{eq_slip_dyn}&Small angle approximation $\theta$ applied at $\theta = \Xi_\theta$\citep{geyer_spring-mass_2005} &\cref{eq_slip_ass_dyn}\\
      \midrule
   \label{ass_b}& $b = \Oeps$ & \cref{eq_slip_dyn}& Taken from \cite{de_hybrid_2018}\endnote{In the application setting of relevance to this paper, the Jerboa's damping (\cref{tab_model_parameters}) $b$ is not  $\Oeps$ since its value is 40 \si{Ns/m}. This is a mathematically convenient substitute for the empirically more realistic observation that during stable operation $b \dot{r} = \Oeps$ compared to the force from the spring. As a side effect of this assumption there are a few places in the dynamics where $b$ shows up without $\dot{r}$ that are now considered $\Oeps$. The most noteworthy example is in $\dot \psi_e$ \cref{eq_dpsie_ass} which in turn contributes to the error in steady state height by under estimating the amount of asymmetry in the stepping trajectories. In the future, developing  an alternate form of this assumption would likely improve the model accuracy. } & \cref{eq_slip_ass_dyn}\\
      \midrule

   \label{ass_g}& $g = \Oeps$ &  \cref{eq_slip_dyn} & Taken from  \cite{de_modular_2017}& \cref{eq_slip_ass_dyn}\\
   \midrule
   \label{ass_leg_length}& $k = \mathcal{O}(\epsilon^{-2})$ &  \cref{eq_slip_dyn} & Taken from  \cite{de_modular_2017}& \cref{eq_slip_ass_dyn}\\
      \midrule

   \label{ass_tau0_star}& $\bar\tau^* \sin \gamma \psi_e^d = \mathcal{O}(\epsilon)$ & \cref{eq_slip_dyn} & Enforced choice of $\bar\tau^*$ & \cref{eq_slip_ass_dyn} \\
      \midrule

   \label{ass_trig}& $\tan \gamma \psi_e^d \sin \gamma \psi_e^d = \mathcal{O}(\epsilon)$ & \cref{eq_slip_dyn} & Enforced choice of $\psi_e^d$ & \cref{eq_slip_ass_dyn} \\
      \midrule

  \label{ass_small_odd} & $\frac{a_e \sin \gamma \psi_e^d \sin \psi_r}{r_0} \approx 0$ & \cref{eq_slip_dyn} & $a_e$ is small steady state and $\sin \gamma \psi_e^d$ is small \endnote{While in the theory energy can grow unbounded, in the physical system torque limits and energetic losses due to friction and damping prevent the energy from being large.} & \cref{eq_slip_ass_dyn}\\
\bottomrule
\end{tabular}
\end{table*}

 \Cref{ass_psi_e_replacemt} is imposed to achieve a closed form expression for the averaging integral from \cite{de_hybrid_2018} Theorem 2.
Meanwhile \cref{ass_theta_replacement} avoids the introduction of a transcendental equation in the definition via implicit function of $a^*_e$.\endnote{Note that for ease of exposition we apply \cref{ass_theta_replacement} to all equations in \cref{eq_slip_dyn},  although our  purposes require its application only  to \cref{eq_dae}. }
\Cref{ass_b,ass_g,ass_leg_length,ass_tau0_star,ass_trig} introduce the averaging parameter $\epsilon$ and assign certain parameters to be $\Oeps$.  \Cref{ass_small_odd} is empirically observed under the conditions of a small $a_e$ and small $\sin \gamma \psi_e^d$ as plotted in \cref{fig_hard_time}. While this term is not $0$, it is odd, and if small would disappear post averaging.\endnote{Only odd terms that are $\Oeps$ disappear with averaging \cite{de_modular_2017}} Many of these assumptions rely on the energy of the system not being too high and thus break down for higher $\bar\tau^*$, foreshadowing the empirically observed loss of stability at high energy fixed points.

\subsubsection{Averaged Stance Dynamics and Fixed Points}
\label{sec_slip_dyn}
Starting from the dynamics in terms of $\zs$ from \cref{eq_ddr,eq_ddtheta}, we first apply the change of coordinates $\g \circ \hs$ from \cref{eq_g_def,eq_hybrid_vars}. Next we apply the assumptions from \cref{table_hybrid_averaging_assumptions} yielding dynamics in the form of \cref{def_avik} for $\sigma = \psi_r$, $\bm \psi = [\tilde{\psi}_e, \psi_\theta]^T$, and $\bm a = [a_e]$. For a detailed derivation and the full equations, see \cref{eq_slip_ass_dyn}.

The dynamics resulting from the change of coordinates and the simplifications of \cref{table_hybrid_averaging_assumptions} satisfy \cref{def_avik} (i) and (ii) but do not satisfy (iii). Instead, we will use the computational recipe from \cref{sec_recipe} and the projection $\h$ from \cref{eq_state_phasediff} to re-express the system in coordinates that  satisfy the averageability requirements of \cite{de_hybrid_2018}.

\label{sec_slip_fixed}
After applying the projection $\h$ \cref{eq_state_phasediff}, the system \cref{eq_phase_diff} is of the form of \cite{de_hybrid_2018} equation 12 according to \cite{de_modular_2017} Lemma 6.
Next we compute the averaged dynamics of $\x$ from \cite{de_hybrid_2018} and step \ref{comp_averaging} \cref{eq_avg_integral}:
\begin{equation} \label{eq_av_dyn_slip}
\begin{split}
    \dot {\hat {\x}} &= 
\begin{bmatrix}
\dot \delta_e \\
\dot \delta_\theta \\
\dot a_e
\end{bmatrix}  = \hat f(\hat \x) \\
&=
     \frac{2 \bar\tau^*  \sin{(\gamma \psi_e^d)} - a_e b \cos^2(\gamma \psi_e^d)}{2 m \omega_r}\begin{bmatrix}
         \delta_e/a_e \\
         -\delta_\theta/a_e \\
         1
    \end{bmatrix}\,.
\end{split}
\end{equation}
As proven by \cite{de_modular_2017} Lemma 8, our choice of $\omega_1$ \cref{eq_w1} ensures that $\bm \delta = 0$ is a fixed point for $\dot{ \bm{ \delta}}$.
Notice, as well, the averaged energy dynamics formalize the manner in which the hip torque is adding energy at a constant rate proportional to $\bar\tau^*$ \cref{eq_torque} and losing a proportional amount of energy from damping $b$. The averaged vector field $\hat{f}$ vanishes on an entire line including the (consequently degenerate) equilibrium state 
\begin{equation} \label{eq_x_star}
    \hat \x^* = \begin{bmatrix}
        0 \\
        0 \\
        \frac{2 \bar\tau^* \sec(\gamma \psi_e^d)\tan(\gamma \psi_e^d)}{b}
    \end{bmatrix}\,.
\end{equation}
 This satisfies step \ref{comp_fixed} and the fixed point requirement of \cite{de_hybrid_2018} theorem 2, and it also demonstrates that \cref{def_avik} (iii) is sufficient but not necessary for a system to have a fixed point in its averaged dynamics.
 
 While the averaged stance dynamics are non hyperbolic,\endnote{\cite{de_hybrid_2018} only require hyperbolicity of the return map.} suggesting that that the ancillary phases are marginally stable, the full averaged return map is hyperbolic as shown in \cref{lemma_av_stability}, corroborating the physical insight that the stability for the ancillary phases is driven by the hybrid reset \cref{eq_reset_slip} and the stepping strategy \cref{eq_stepping}.

\subsubsection{Hybrid Structure: Guards, Reset and Associated Assumptions}
\label{sec_slip_guards}
\label{sec_slip_reset}

Next we make the assumptions in \cref{table_assumption_reset} in order to satisfy the reset map requirements of \cite{de_hybrid_2018} Theorem 2. According to \cref{ass_liftoff}, liftoff occurs when $r = r_0$ and $\dot r >0$. By \cref{eq_hybrid_vars}, ${\psi_r}_\text{td} = -\uppi/2$ and ${\psi_r}_\text{lo} = \uppi/2$, where ``$\text{td}$'' and ``$\text{lo}$'' indicate touchdown and liftoff, respectively.

\begin{table*}[t]
\centering
\small\sf%\centering
\caption{Assumptions applied to guards and resets of 2 DoF SLIP} \label{table_assumption_reset}
\begin{tabular}{N
>{\raggedright\arraybackslash}p{4.5cm}
l
>{\raggedright\arraybackslash}p{4.5cm}
>{\raggedright\arraybackslash}p{2.5cm}}
\toprule
\multicolumn{1}{c}{\textbf{Assumption \#}}  & \textbf{Formula} & \textbf{First use}  & \textbf{Justification} & \textbf{Consequence}
                 \\ 
\midrule
   \label{ass_liftoff}& Liftoff occurs at $r = r_0$ and $\dot r  >0$ &  \cref{sec_slip_guards} & Taken from  \cite{de_modular_2017}& \cref{eq_slip_jac}\\
   \midrule
   \label{ass_flight}& Change in gravitational potential energy from touchdown to liftoff is negligible & \cref{eq_reset_slip} & Change in gravitational potential is small compared to kinetic energy \endnote{While our controller does generate trajectories where the touchdown angle, $\theta_\text{td}$  is not symmetric relative to the prior liftoff angle, $\theta_\text{lo}$ about vertical (thus gravitational potential energy is changing), the change is negligible compared to the kinetic energy in the system \cref{fig_sim_time_theta}} & \cref{eq_reset_slip}\\
   \midrule
   \label{ass_prev_td}&$\theta_{td,\text{prev}} \approx \theta_{\text{lo}} - \uppi/\omega_\theta(\bm a)$ & \cref{eq_stepping_filt} & Accurate near steady state trajectories & \cref{eq_slip_jac} \\
   % \midrule
   % \label{ass_omega_star}& $\Pi_1 \omega_0(\bm a) \approx \Pi_1 \omega_0(\bm a^*)$ & \cref{eq_s0} & Accurate near steady state trajectories & \cref{eq_s0}\\
\bottomrule
\end{tabular}
\end{table*}

Using \cref{ass_flight} to neglect the change in gravitational potential energy during flight, simplifying the second and fourth entry of the reset map, the reset map $R_z: \bm{z}_{\text{lo}} \mapsto \bm{z}_{\text{td}}$ becomes
\begin{align}
    R_z(\z) = \begin{bmatrix}
    -\uppi/2\\
    \theta_{\text{td}}(\z) + \theta_{\text{lo}} + {\psi_e}_{\text{lo}} \\
    \theta_{\text{td}}(\z) \\
    {a_e}_{\text{lo}}
    \end{bmatrix} = \begin{bmatrix}\psi_r \\ \psi_e \\ \theta \\ a_e \end{bmatrix}_\text{td}\,, \label{eq_reset_slip}
\end{align}
where $\theta_{\text{td}}(\z)$ is the stepping controller to be defined next. See Appendix \ref{app_reset} for a detailed derivation of the reset map. The reset map reveals the direct control over ${\psi_{e}}_\text{td}$  afforded by the  choice of $\theta_{\text{td}}(\z)$. 
% Then, let $R_y \coloneqq \g \circ R_z \circ \g^{-1}$ with $\g$ as defined in \cref{eq_g_def}.

Motivated by \cref{eq_reset_slip}, we now define a \textbf{symmetric stepping strategy} to be one where $\theta_\text{td} = -\theta_\text{lo}$  (and we will say that stepping is \textbf{asymmetric} when that condition fails).
The following result demonstrates the fundamental connection between an asymmetric leg angle trajectory and a change in energy ratio $\psi_e$ during stance imposed by the second component of the reset map \cref{eq_reset_slip}.
This connection suggests that in order to have period-one gait, the robot cannot use a symmetric stepping controller like the neutral angle in \cite{raibert_legged_1986} in hip-energized gaits. Instead, such a monoped must use an asymmetric stepping controller. Additionally since the following theorem is applicable to wide range of control strategies we now define a SLIP control strategy to be \textbf{hip energized} if the integral of the hip torque is non-negligible. 

\begin{theorem}[Stepping Asymmetry in Hip Energized Hopping]\label{theorem_asymmetry}
    Consider the 2 DoF, pitch locked SLIP system with stance dynamics \cref{eq_ddr,eq_ddtheta},  while neglecting the change in gravitational potential energy in flight (\cref{ass_flight}) yielding the stepping agnostic reset map \cref{eq_reset_slip}, assuming gravity acts radially in stance (\cref{ass_bad1}),\endnote{An alternative version of this assumption would be to either assume that the hip torque is directly countering the effect of gravity on the angular subsystem or that the moment from gravity is negligible compared to the hip torque.} and assuming liftoff happens at leg reset length (\cref{ass_liftoff}). A hip-torque energized period 1  hybrid closed orbit of   such a system  can only be achieved by recourse to asymmetric stepping.
\end{theorem}
\begin{proof}
    Starting with the Lagrangian, pinned toe SLIP stance dynamics in \cref{eq_ddr,eq_ddtheta} and neglecting the Coriolis term in \cref{eq_ddr} and neglecting the effect of gravity on $\dot \theta$ we apply the change of coordinates
    \begin{align}
    \bm q_e =
    \begin{bmatrix}
        \psi_r \\
        \theta \\
        a_r \\
        a_\theta \\
    \end{bmatrix} = h_\theta (\zs) = 
    \vctfour
    { \arctan \frac{\dot{r}}{(r_0-r) \omega_r}  }
    { \theta}
    { \sqrt{{\dot{r}^2 + (r_0-r)^2\omega_r^2}} }
    { r^2 \dot \theta /r_0 } \,. \nonumber
    \end{align}
    Yielding
    \begin{equation} \label{eq_d_a_theta}
        \dot a_\theta =
        \frac{\tau}{m r_0}\,.
    \end{equation}

We now make the following observations (wherein each itemized claim is justified in the ensuing text for that item):
\begin{observation} \label{claim_atheta_change}
    ${a_\theta}_\text{td} \neq {a_\theta}_\text{lo}$.
\end{observation}
\begin{proof}
    From \cref{eq_d_a_theta}, the change in $a_\theta$ over the course of stance is proportional to $\int_T \tau \; dt$  where $\int_T$ is the integral over the course of stance. Since the hip torque is energizing, $\int_T \tau \; dt \neq 0$, thus ${a_\theta}_\text{td} \neq {a_\theta}_\text{lo}$.
\end{proof}
\begin{observation}\label{claim_ae_const}
    On a period 1 closed orbit, ${a_e^*}_\text{lo} = {a_e^*}_\text{td}$.
\end{observation}
\begin{proof}
    From \cref{eq_reset_slip}, ${a_e}_\text{lo}$ is always invariant under the flight map, thus ${a_e^*}_\text{lo} = {a_e^*}_\text{td}$.
\end{proof}
\begin{observation}\label{claim_psie_const}
    On a period 1 closed orbit with a symmetric stepping strategy, ${\psi_e^*}_\text{lo} = {\psi_e^*}_\text{td}$.
\end{observation}
\begin{proof}${\psi_e^*}_\text{lo} = {\psi_e^*}_\text{td}$
    From \cref{eq_reset_slip}, if the stepping strategy is symmetric,  ${\psi_e}_\text{lo}$ is invariant under the flight map. Thus ${\psi_e^*}_\text{lo} = {\psi_e^*}_\text{td}$.
\end{proof}
\begin{observation}\label{claim_a_theta_inversion}
    At touchdown and liftoff $a_\theta = a_e \sin \psi_e$
\end{observation}
\begin{proof}
    At touchdown and liftoff $r = r_0$, thus $a_\theta = r \dot \theta$. Thus from \cref{eq:change_inverse}, at touchdown and liftoff $a_\theta = a_e \sin \psi_e$.
\end{proof}
Now suppose, contrarily that there is a period 1 closed orbit with a symmetric stepping strategy and an energizing hip torque where ${a_e^*}_\text{td}$ and ${\psi_e^*}_\text{td}$ are projections of the touchdown coordinate fixed point. 
Invoking \cref{claim_psie_const}, under this contrary assumption, in conjunction with \cref{claim_ae_const} applied to \cref{claim_a_theta_inversion} implies ${a_\theta^*}_\text{td} = {a_\theta^*}_\text{lo}$, contradicting \cref{claim_atheta_change}. Avoiding this contradiction --- i.e., negating the contrary assumption --- now yields the stated conclusion of the Theorem. \qed
\end{proof}

While \cref{theorem_asymmetry} is controller agnostic and thus does not prescribe computationally the required amount of asymmetry, we can use our averaged fixed points from \cref{eq_x_star} and averaged flows from \cref{z_flow} to calculate the necessary asymmetry for our system.
\begin{corollary}
    Consider the 2 DoF unperturbed SLIP subsystem under the assumptions in \cref{table_hybrid_averaging_assumptions,table_assumption_reset}. On a period 1 closed orbit, $\theta^*_\text{td} = -\theta^*_\text{lo} - \frac{\uppi}{\omega_e(\hat{\bm a}^*)}$.
\end{corollary}
\begin{proof}
    Since we are on a period 1 closed orbit the change in $\psi_e$ over the course of flight is equal to the opposite of the change in $\psi_e$ over the course of stance. From \cref{z_flow}, 
    \begin{equation*}
    \begin{split}
            {\psi^*_e}_\text{lo} - {\psi^*_e}_\text{td} = &\Pi_2 \big(\g^{-1} \circ \h^{-1}(\hat\x^*,\pi/2)\\
            &- \g^{-1} \circ \h^{-1}(\hat\x^*,-\pi/2)\big) = \frac{\uppi}{\omega_e(\hat{\bm a}^*)}\,,
    \end{split}
    \end{equation*}
    where $\omega_e(\bm a) \coloneqq \Pi_1 \omega(\bm a)$. Thus rearranging the second equation from \cref{eq_reset_slip}
    \begin{align}
        \theta^*_\text{td} &=- \theta^*_\text{lo} - \left( {\psi^*_e}_\text{lo} - {\psi^*_e}_\text{td} \right) \nonumber \\
        \theta^*_\text{td} &= -\theta^*_\text{lo} - \frac{\uppi}{\omega_e(\hat{\bm a}^*)}     \label{eq_asymmetry}
    \end{align}
\end{proof}

\subsubsection{Hybrid Dynamics Stepping Controller} \label{sec_stepping}
The foregoing  hybrid (guard and reset)  structure now invites the presentation of a central contribution of this paper: the discrete event triggered stepping control strategy. This controller takes the form of the following  $\alpha$-parametrized family of Raibert-style \citep{raibert_legged_1986} stepping strategies,  
\begin{subequations}\label{eq_stepping}
\begin{align}
    \theta_{\text{td}}(\z) &\coloneqq (1- \epsilon \alpha) \tilde \theta_{\text{td}}(\z) + \epsilon \alpha \theta_{\text{td},\text{prev}} \,,
     \label{eq_stepping_filt} \\
    \tilde \theta_{\text{td}}(\z) &\coloneqq - \theta_\text{lo} + \epsilon k_e (\psi_e^d - \psi_e) - \frac{ \uppi}{\omega_e(\bm a)}\,,\label{eq_stepping_pre_filt}
\end{align}
\end{subequations}
where $\alpha$ is a low pass filter gain, $\theta_{\text{td},\text{prev}}$ is the previous touchdown angle, $k_e$ is the proportional gain on the energy ratio $\psi_e$, and $\psi_e^d$ is the target energy ratio.

The stepping controller is the low pass filter\endnote{Our value of $\alpha$ is $1-\tilde\alpha$ of the $\tilde\alpha$ in the standard formulation of an exponentially weighted moving average filter \citep{everett_exponentially_2011}.}
of an asymmetric stepping controller \cref{eq_stepping_pre_filt} that uses the change in energy ratio on the closed orbit calculated in \cref{eq_asymmetry} to create a steady state for $\psi_e$ when $\psi_e= \psi_e^d$ and $\bm a = \bm a^*$.\endnote{We remind the reader that this low pass filer has no effect on the location of fixed points, but only on their stability.} As with many scissor stepping controllers \citep{de_averaged_2018}, our unfiltered stepping controller \cref{eq_stepping_pre_filt} has period 2 oscillations. Fortunately the low pass filter acts as a damper, punishing large changes in touchdown angles and collapsing the period 2 oscillations to a fixed point.

The first term of \cref{eq_stepping_pre_filt} helps to stabilize the symmetric portion of the leg angle. Combined with the filtering, it either acts like Raibert scissor stepping for $\epsilon \alpha = 0$, Raibert neutral point approximation \citep{raibert_legged_1986} for $\epsilon\alpha = 0.5$, or a fixed touchdown angle \citep{geyer_spring-mass_2005,Ghigliazza_Altendorfer_Holmes_Koditschek_2003} for $\epsilon\alpha = 1.0$. 
The second term is a proportional controller about the energy ratio $\psi_e$ at liftoff, which drives $\psi_e$ to the target energy ratio $\psi_e^d$. 
The third term is then imposed as a consequence of \cref{theorem_asymmetry} and is the expected change of $\psi_e$ over the course of the next stance. 

In order to understand what the stepping controller is doing to the energy ratio, $\psi_e$ we neglect the low pass filter (thus $\theta_{\text{td}}(\z) = \tilde \theta_{\text{td}}(\z)$ and evaluate \cref{eq_reset_slip} at \cref{eq_stepping_pre_filt} yielding
\begin{align*}
    {\psi_e}_\text{td} &= {\psi_e}_{\text{lo}} + \epsilon k_e (\psi_e^d - {\psi_e}_{\text{lo}}) - \frac{ \uppi}{\omega_e(\bm a)} \,, \\
    {\psi_e}_\text{td} + \frac{ \uppi}{\omega_e(\bm a)} &= {\psi_e}_{\text{lo}} + \epsilon k_e (\psi_e^d - {\psi_e}_{\text{lo}}) \,,\\ 
    {\psi_e}_\text{lo, next} &\approx {\psi_e}_{\text{lo}} + \epsilon k_e (\psi_e^d - {\psi_e}_{\text{lo}})\,,
\end{align*}
which demonstrates the stepping controller is creating a discrete stride-event proportional controller about the energy ratio at liftoff.

Finally, in order to reduce the dimension of the system for analytical tractability, it is helpful to approximate $\theta_{\text{td}, \text{prev}}$ from the state at liftoff by assuming that the system is already at steady state and flowing backwards in time/master phase along the averaged, phase-locked, secondary phase trajectory using the flow from \cref{z_flow}. Thus from \cref{ass_prev_td}, $\theta_{td,\text{prev}} \approx \theta_{\text{lo}} - \uppi/\omega_\theta(\bm a)$ for $\omega_\theta(a) \coloneqq \Pi_2 \omega(\bm a)$. This becomes exact on the averaged trajectory when $\x =\hat\x^*$. On the physical system we take advantage of the low cost of recording previous states and measure and remember the previous touchdown angle rather than estimating it from the state at liftoff.

\subsubsection{Secondary Phase Offsets}
\label{sec_phase_offset}
Now we consider the assignment of values to $\Xi_e$ and $\Xi_\theta$ whose purpose in \cref{eq_g_def} is to force the (projected) fixed point condition in the (projected) translated coordinates, s.t. $R(\hat\x^*) = \hat\x^*$ for
\begin{equation} \label{eqn:Resetx}
R(\bm{x}) \coloneqq \h \circ \g \circ R_z \circ \g^{-1} \circ \h^{-1}(\bm{x},\pi/2)\,.
\end{equation}
Given a putative set of ancillary phases at steady state $\bm \psi^*$, \cref{lemma_reset_map_general} provides a sufficient condition on the reset in the full, untranslated coordinates in \cref{eq_reset_requirment} as $R_z([\sigma_T,\bm\psi^*,\bm a^*]) = [ \sigma_T, \bm \psi^*, \bm a^*] - [\sigma_T-\sigma_0, \Delta(\omega^*)^{-1} \bm 1 (\sigma_T-\sigma_0),0]$ in order for $R(\hat \x^*) = \hat\x^*$ for some $\Xi_e, \Xi_\theta$. Then \cref{lemma_reset_phase_offset} provides a method for calculating the specific values for $\Xi_e, \Xi_\theta$. The following result shows that these calculations indeed achieve the reset fixed point condition listed as step \ref{comp_reset} in \cref{sec_recipe} and \cite{de_hybrid_2018} reset map fixed point requirement.
\begin{proposition}
\label{cor.ResetFixedPoint}
Given a desired energy ratio $\psi_e^d$  the choice of $\theta^*_\text{lo}$ in \cref{theta_offset}  and   $\Xi = [ \Xi_e, \Xi_\theta ] $ in \cref{eq_phase_offsets}, achieves the reset fixed point  $R(\hat\x^*) = \hat\x^*$ in \cref{eqn:Resetx}. 
\end{proposition}
\begin{proof}
 Evaluating the  $\psi_e$ component of $R_z$  \cref{eq_reset_slip} at this putative steady state value, and applying the stepping controller

  \begin{align*}
     {\psi_e}_\text{td} &=  \psi_e^d + \theta^*_\text{lo} + \theta_\text{td}^* \\
      &\stackrel{ (\ref{eq_stepping_pre_filt})}{=}
 \psi_e^d - \frac{\uppi}{\omega_e(\bm a^*)}\,,
 \end{align*}
 thereby satisfying the reset map requirement of \cref{lemma_reset_map_general} for $\psi_e$.
 Next the $\theta$ component of $R_z$ \cref{eq_reset_slip} is
 \begin{align*}
     \theta_\text{td} &= -\theta^*_\text{lo}- \frac{\uppi}{\omega_e(\bm a^*)}
 \end{align*}
by evaluation the stepping controller \cref{eq_stepping_pre_filt} at $\psi^* = [\psi_e^d, \theta^*_\text{lo}]$.
The steady state value of the leg at liftoff $\theta^*_\text{lo}$ now follows by setting the required form of the reset map  \cref{eq_reset_requirment} from \cref{lemma_reset_map_general} equal to the right hand side of the above equation, i.e.,
\begin{align}
    -\theta^*_\text{lo}- \frac{\uppi}{\omega_e(\bm a^*)} &= \theta^*_\text{lo} - \frac{\sigma_t - \sigma_0}{\omega_\theta(\bm a^*)} \nonumber \\
     &= \theta^*_\text{lo} - \frac{\uppi}{\omega_\theta(\bm a^*)} \nonumber \\
      & \Downarrow \nonumber\\
    \theta^*_\text{lo} &= \frac{\uppi}{2 \omega_\theta(\bm a^*)} -\frac{\uppi}{2\omega_e(\bm a^*)} \label{theta_offset}\,,
\end{align}
where $\omega_\theta(\bm a) \coloneqq \Pi_2 \omega(\bm a)$.

Finally applying \cref{lemma_reset_phase_offset} to calculate $\Xi_e$ and $\Xi_\theta$ yields
\begin{subequations} \label{eq_phase_offsets}
    \begin{align}
    \Xi_e &= \psi_e^d - \frac{\uppi}{2 \omega_e(\bm a^*)} \label{eq_xi_e}\\
    \Xi_\theta &= \frac{\uppi}{2 \omega_\theta(\bm a^*)} -\frac{\uppi}{2\omega_e(\bm a^*)} - \frac{\uppi}{2 \omega_\theta(\bm a^*)} \nonumber\\
               &=-\frac{\uppi}{2\omega_e(\bm a^*)} \label{eq_xi_theta}\,.
\end{align}

\end{subequations}
The system now satisfies the statement for \cref{lemma_reset_map_general}, which proves that $R(\hat\x^*) = \hat\x^*$. \qed 
\end{proof}

\subsubsection{Reset Map Check} 
\label{sec_slip_reset_check}
Given that our system satisfies  step~\ref{comp_reset} and \cite{de_hybrid_2018} Theorem 2 requirement (ii.b), $R(\hat\x^*) = \hat\x^*$, we now move checking that $DR = \bm S_0 + \epsilon \bm S_1(\x) + \OepsSQ$ for \cite{de_hybrid_2018} Theorem 2 requirement (i) and step~\ref{comp_jacob}.

\begin{lemma}
\label{prop_reset_jacobian}
    $DR = \bm S_0 + \epsilon \bm S_1(\x) + \OepsSQ$ and $S_0$ is invertible and its unity eigenvalues have diagonal Jordan blocks for $R(\x)$ in \cref{eqn:Resetx},  
\end{lemma}
\begin{proof}
    Starting from $R_z$ in \cref{eq_reset_slip} we apply the change of coordinates resulting in
    \begin{align*}
        R(\x) = \vctthree{\delta_e}{-\delta_\theta}{a_e} + \epsilon \vctthree{R_1}{R_2}{0} + \OepsSQ
    \end{align*}
    where \begin{align*}
        R_1 \coloneqq & \scriptstyle\frac{ \tan ^3(\gamma  \psi_e^d) \sec ^2(\gamma  \psi_e^d) \left(8 \alpha \delta_\theta m\bar\tau-b^2 \delta_e k_e r_0 \cos ^2(\gamma  \psi_e^d) \cot ^3(\gamma  \psi_e^d)\right)}{b^2 r_0} \\
        \begin{split}
             R_2 \coloneqq & \scriptstyle\frac{ -4 a_e^2 b^3\bar\tau \cos ^4(\gamma  \psi_e^d) \cot ^2(\gamma  \psi_e^d)+ \pi  a_e^2 b^4 r_0 \omega_r \cos ^4(\gamma  \psi_e^d) \cot ^3(\gamma  \psi_e^d)}{16 m\bar\tau^3 \omega_r}\\
             &\scriptstyle-\frac{4 b^2 \delta_e k_e r_0\bar\tau^2 \omega_r \cos ^2(\gamma  \psi_e^d) \cot ^3(\gamma  \psi_e^d)}{16 m\bar\tau^3 \omega_r}\\
             & \scriptstyle -\frac{ 4 \pi  b^2 r_0\bar\tau^2 \omega_r \cos ^2(\gamma  \psi_e^d) \cot (\gamma  \psi_e^d)+16 b\bar\tau^3 \cos ^2(\gamma  \psi_e^d)+32 \alpha \delta_\theta m\bar\tau^3 \omega_r}{16 m\bar\tau^3 \omega_r}\,.
        \end{split}
    \end{align*}
    Thus
    \begin{align*} 
    \bm S_0 = \begin{bmatrix}
    1 & 0 & 0 \\
    0 & -1 & 0 \\
    0 & 0 & 1
    \end{bmatrix},
\end{align*}
which is constant and invertible, and its unity eigenvalues have diagonal Jordan blocks. \qed
\end{proof}

\subsubsection{Stability of the Averaged Hybrid Dynamics}
\label{sec_slip_stability_check}
The utility of \cite{de_hybrid_2018} Theorem 2 is that stability of the unaveraged system depends on the stability of the averaged return map. Accordingly, \cref{lemma_av_stability} provides the last check in order to apply \cite{de_hybrid_2018} Theorem 2 in \cref{prop_slip_hybrid} -- an examination of the eigenvalues of the linearized averaged Poincar\'{e} map 
$\Shp(\hat\x^*)$,

\begin{lemma}[Stability of Unperturbed Averaged 2 DoF SLIP Return Map] \label{lemma_av_stability}
    Under the assumptions in \cref{table_hybrid_averaging_assumptions,table_assumption_reset} and for small $\epsilon$, and $\epsilon k_e, \epsilon \alpha \in (0,1)$,
    the liftoff coordinate return map of the unperturbed, averaged translational system  
    $\Shp$ \cref{eq_slip_avg_return}   is asymptotically stable at the fixed point,  $\hat\x^*$ \cref{eq_x_star}.  
\end{lemma}
\begin{proof}
    It suffices to prove that Jacobian, of the liftoff coordinate return map at the fixed point $D \Shp(\hat\x^*)$ has eigenvalues inside the unit circle of the complex plane. From the proof of \cite{de_hybrid_2018} Theorem 2, 
\begin{align} \label{eq_slip_jac}
\begin{split}
        D &\Shp(\hat\x^*) =  (\bm S_0 + \epsilon (\bm S_1 + \bm V)) + \OepsSQ \\&= \begin{bmatrix} 
    1 - \epsilon k_e & \epsilon A_1 & 0 \\
    \epsilon A_2& -1 + \epsilon 2 \alpha & \epsilon A_3\\
    0 & 0 & 1 - \frac{\epsilon b \uppi \cos^2(\gamma \psi_e^d)}{2 m \omega_r}
    \end{bmatrix} + \OepsSQ,
\end{split}
\end{align}
where $\bm V = T D \hat f$, and
\begin{align*}
    A_1 \coloneqq & \frac{8 m \alpha \bar\tau^* \sec^2 \gamma \psi_e^d \tan^3 \gamma \psi_e^d}{b^2 r_0}\,\\
    A_2 \coloneqq & \frac{-b k_e r_0 \cos^2 \gamma \psi_e^d \cot^3 \gamma \psi_e^d}{4 m \bar\tau^*}\, \\
    A_3 \coloneqq & \frac{b^2 \cos^3 \gamma \psi_e^d \cot \gamma \psi_e^d \left( -4 \bar\tau^* + b \pi r_0 \omega_r \cot \gamma \psi_e^d \right) }{4 m (\bar\tau^*)^2 \omega_r}\,.
\end{align*}
    For small $\epsilon$ the diagonal terms of \cref{eq_slip_jac} dominate the eigenvalues of the averaged unperturbed return map $\Shp$, thus the eigenvalues for small $\epsilon$ are
    \begin{align} \label{eq_eigenvalues}
        \lambda_{\psi_e} &\approx 1 - \epsilon k_e \\
        \lambda_{\theta} &\approx -1 + 2 \epsilon \alpha \\
        \lambda_{a_e} &= 1 - \frac{\epsilon b \uppi \cos^2(\gamma \psi_e^d)}{2 m \omega_r}
    \end{align}
    which are inside the unit circle for physical parameters and $\epsilon k_e, \epsilon \alpha \in (0,1)$. \qed
\end{proof}

The averaged eigenvalues in \cref{eq_eigenvalues} reveal that $k_e$, the stepping gain on $\psi_e$ controls the stability of $\psi_e$; while $\alpha$, the low pass filter stepping gain, controls the stability of $\theta$. As predicted in \cref{sec_stepping} for $\epsilon \alpha = 0$ the system is oscillatory and marginally stable which matches the empirical experience of scissor stepping.

\subsubsection{Stability of the Limit Cycle}
\Cref{prop_slip_hybrid} demonstrates how under the assumption in \cref{table_hybrid_averaging_assumptions,table_assumption_reset}, the unperturbed translational liftoff coordinate return map $\Sep$ is locally exponentially stable using \cite{de_hybrid_2018} Theorem 2. 
In contrast, whereas conditions for the exponential stability of limit cycles with hyperbolic return maps are available in the classical dynamical systems literature, the corresponding relationships are still being worked out for hybrid systems. 
As discussed in \cite{de_hybrid_2018}, averageable hybrid dynamical systems are locally topologically conjugate to classical limit cycles, but conditions under which their local exponential stability can be deduced from that of their return maps typically require more intricate analysis (e.g. see \cite{burden_event-selected_2015}).
We therefore find it expedient to simply conjecture  that this relationship holds in the present case as follows.

\begin{conjecture} \label{conj:bibostablelimitcycle}
The limit cycle associated with the averaged unperturbed SLIP dynamics is locally exponentially stable and hence bounded continuous time inputs incur only bounded excursions away from the steady state cycle. For sufficiently small $\epsilon >0$, the same property holds for the unaveraged unperturbed dynamics. 
\end{conjecture}

\subsection{Pitch Unlocked SLIP} \label{sec_pitch_unlocked_slip}
The analysis of the isolated pitching subsystem in \cref{sec_attitude} and
the unperturbed 2 DoF SLIP subsystem in \cref{sec_slip} now affords a return to
their cascade composition in the complete 3 DoF (pitch unlocked) SLIP system depicted in \cref{fig_ASLIP}.

\subsubsection{Fixed points}
As established in \cref{lemma_cascade_stability}, under the assumptions in \cref{table_decomp_assumptions}, the cascade compositional structure of the exact liftoff coordinate 3 DoF pitch unlocked SLIP return map $\Re$ depends on the fixed point of the isolated pitch return map $\P$ and the unperturbed 2 DoF translational SLIP return map $\Sep$. Thus,
\begin{subequations}\label{eq_3dof_slip_fixed}
    \begin{align}
    \q^* = \vcttwo{\zp^*}{\zs^*}\,
\end{align}
is the fixed point of 3 DoF pitch unlocked SLIP return map, $\Re$,
where from \cref{eq_pitch_fixed_points},
\begin{align*}
    \zp^* = \vctthree{\phi^*}{\dot\phi^*}{\bar\tau^*} = \vctthree{\phi^d}{0}{\frac{m g}{r_0} \left(\chi d_x - r_0 \sin \left(\Xi_{\theta,g}\right)\right)}\,,
\end{align*}
is the fixed point of the pitch return map $\P$,
\begin{align}
    \zs^* &= \hat{\zs}^* + \Oeps = \hs^{-1}(\hat \z^*_\text{lo}) + \Oeps\,, \nonumber \\
    &= \vctfour{r^*}{\theta^*}{\dot r^*}{ \dot \theta^*}_\text{lo} = \vctfour{r_0}{\frac{\uppi}{2 \omega_\theta(\bm a^*)} -\frac{\uppi}{2\omega_e(\bm a^*)}}{\frac{2 \bar\tau^* \sec(\gamma \psi_e^d)\tan(\gamma \psi_e^d) \cos\psi_e^d}{b}}{\frac{2 \bar\tau^* \sec(\gamma \psi_e^d)\tan(\gamma \psi_e^d)\sin \psi_e^d}{b r_0}}  + \Oeps \label{eq_unav_slip_fixed}
\end{align}
is the fixed point of the unperturbed translation 2 DoF SLIP return map $\Sep$, and from \cref{eq_x_star},
\begin{align} \label{eq_qs_star}
\begin{split}
        \hat\z^*_\text{lo} &= \g^{-1} \circ \h^{-1}(\hat \x^*, \pi/2)\\ &= \vctfour{\psi_r}{\psi_e^*}{\theta^*}{a_e^*}_\text{lo}=\vctfour{\pi/2}{\psi_e^d}{\frac{\uppi}{2 \omega_\theta(\bm a^*)} -\frac{\uppi}{2\omega_e(\bm a^*)}}{\frac{2 \bar\tau^* \sec(\gamma \psi_e^d)\tan(\gamma \psi_e^d)}{b}}\,
\end{split}
\end{align}

\end{subequations}
is the fixed point of the averaged SLIP return map $\Shp$ projected into $\z$ \cref{eq_hybrid_vars}. 
Given the fixed point $\q^*$ we can now calculate the value of $ \Xi_{\theta,g}$ from \cref{ass_xi_theta_replacement} which is the average leg angle when we assume gravity acts radially as found in \cref{sec_xi_theta_g} and
\begin{equation}
    \Xi_{\theta,g} \coloneqq \frac{-b \uppi \cos^2(\gamma \psi_e^d) \cot{\gamma \psi_e^d}}{2 \chi m \omega_r}\,. \label{eq_xi_theta_g}
\end{equation}

% Leveraging \cref{lemma_cascade_stability} the fixed points for the full system are the fixed points of the subsystems. Thus the liftoff coordinate fixed point is, \SRL{Dan, I might be being silly, but the stance system has 7 dimensions and since these fixed points are at the liftoff poincare section, I would expect a 6 dimensional fixed point which is what is present below.}
% \begin{subequations} \label{eq_3dof_slip_fixed}
% \begin{align}
%      a_e^* &= \frac{2 \bar\tau^* \sec(\gamma \psi_e^d)\tan(\gamma \psi_e^d)}{b} \label{eq_ae_star} \\
%     \psi_e^* &= \psi_e^d \\
%     \theta^*_\text{lo} &= \frac{\uppi}{2 \omega_\theta(\bm a^*)} -\frac{\uppi}{2\omega_e(\bm a^*)}\\
%       \phi^* &= \phi^d \\
%     \dot\phi^* &= 0 \\
%     \bar\tau^* &= \frac{m g}{r_0} \left(\chi d_x - r_0 \sin \left(\Xi_{\theta,g}\right)\right)\,, \label{eq_tau_star_2}
%     %\label{eq_tau_star}\,.
% \end{align}
% \end{subequations}
% % where
% from \cref{ass_xi_theta_replacement} is the average leg angle when we assume gravity acts radially as found in \cref{sec_xi_theta_g}.
% \Cref{eq_apex_coordinate_change} can be used to convert these liftoff coordinate fixed points to apex coordinates.

% The energy fixed point where we assume gravity acts radially is $\bm a_g^* = [ \frac{4 d_x m g \sec (\gamma \psi_e^d)\tan(\gamma \psi_e^d)}{b r_0} , \frac{2 m g d_x}{r_0}]^T$ where the first energy is $a_e$ and the second entry is $\bar\tau$. Thus
% which is used in \cref{ass_xi_theta_replacement}.

\subsubsection{Behavioral Specification: A Map from Target Apex State to Control Input Set Points}
\label{sec_invert_fixed_points}
One of the key advantages of an analytical expression for the fixed points as a function of the model and control parameters is the ease at which it enables a user to choose a set of control parameters which accomplishes a desired behavior. In this section we show that a specified apex height and speed $[\dot p_{x_\text{apex}}^s, p_{z_\text{apex}}^s]$ can be achieved using the desired control set points 
\[
\vcttwo{ \psi_e^d}{d_x} =
\Psi( \dot{\bm{p}}_\text{lo}^s ) :=
\vcttwo{  
\Psi_\zeta \left( \mynorm{\dot{\bm{p}}_\text{lo}^s}, \atan( \dot{\bm{p}}_\text{lo}^s) \right)
}
{
\Psi_d \left(
 \mynorm{\dot{\bm{p}}_\text{lo}^s}, \Psi_\zeta \left( \mynorm{\dot{\bm{p}}_\text{lo}^s}, \atan( \dot{\bm{p}}_\text{lo}^s) \right)\right)
}\,,
\]
where $\dot{\bm{p}}^s$ is the Cartesian coordinate expression for the desired velocity at liftoff,   $\atan$ is the four-quadrant arctangent function and $\Psi_d, \Psi_\zeta$ are defined, respectively, in  \cref{eq_dx_desired,eq_zeta_r_trans}, and $\square^s$ denotes a specified value of $\square$.  In order to simplify various expressions for the sake of getting a (nearly) closed form expression for $\Psi$, we find it convenient to  make the following assumptions in \cref{table_invert_reset}.
In particular, these assumptions are sufficient for the convenient monotonicity result of \cref{lemma_zeta_r_monotonic}. Computational  experience suggests
that the numerical inversion required to evaluate $\Psi_\zeta$ in \cref{eq_zeta_r_trans} works across a far wider range of parameter values on which the assumptions fail to hold.

\begin{table}[h]
\centering
\small\sf%\centering
\caption{Assumptions for inverting the fixed points. These assumptions are stronger versions of those listed in \cref{table_decomp_assumptions,table_hybrid_averaging_assumptions,table_assumption_reset} and are only imposed to achieve simple closed form expressions relating the apex set points to  fixed points. Computational experience suggests that the numerical inversion can still be achieved across a broader range of parameter values that violate these assumptions.} \label{table_invert_reset}
\begin{tabular}{N
>{\raggedright\arraybackslash}p{1.75cm}
>{\raggedright\arraybackslash}p{1.75cm}
l}
\toprule
\multicolumn{1}{c}{\footnotesize\textbf{Assumption \#}}  & \footnotesize\textbf{Formula}   & \footnotesize\textbf{Justification} & \footnotesize{\textbf{Conseq.}}
                 \\ 
\midrule
   \label{ass_bad1}& Gravity acts radially. thus $\Xi_{\theta,g} \approx 0$   & Stronger version of \cref{ass_xi_theta_replacement}& \cref{eq_a_e_star_simple}\\
   \midrule
   \label{ass_bad2}& $\epsilon \omega_1 \approx 0$   & Valid for small $\epsilon$& \cref{eq_a_e_star_simple} \\
   \midrule
    \label{ass_height}& $p_{z_\text{lo}} \approx r_0$ & Valid when $\theta_\text{lo} \approx 0$& \cref{eq_dzr}\\
\bottomrule
\end{tabular}
\end{table}

Using the equations from \cref{eq_apex_coordinate_change}  and \cref{ass_height}, 
\begin{align}
    \dot{\bm{p}}_\text{lo}^s  = \vcttwo{\dot p_{x_\text{lo}}^s}{\dot p_{z_\text{lo}}^s} = \vcttwo{\dot p_{x_\text{apex}}^s}{\sqrt{2  g(p_{z_\text{apex}}^s - r_0)} } \, . \label{eq_dzr}
\end{align}
This equation expresses symbolically the intuitively clear bijective functional mapping between a specified apex behavior and the Cartesian liftoff velocity vector that achieves it for $z^s_\text{apex} > r_0$.
In turn, expressing Cartesian liftoff velocity representation of the specified behavior $\dot{\bm{p}}^s$  in polar coordinates
\begin{equation}
    \tilde{\bm{q}}_s^s \coloneqq
 \vcttwo{
\mynorm{\dot{\bm{p}}_\text{lo}^s}
}
{ \atan( \dot{\bm{p}}_\text{lo}^s)  \label{eq_specefied}
}=\vcttwo{a_e^s}{\zeta^s}_\text{lo} \,,
\end{equation}
where as described in \cref{sec_coordinates}, $\zeta_\text{lo} \coloneqq {\psi_e}_\text{lo} + \theta_\text{lo}$, is the angle of the velocity vector and 
facilitates determining the relationship of $\dot{\bm p}^s_\text{lo}$ to the $\bm{q}$-coordinate representation of fixed points in \cref{eq_3dof_slip_fixed}.
Leveraging \cref{ass_bad1,ass_bad2}, the values of this polar coordinate representation of the liftoff velocity vector at a fixed point \cref{eq_3dof_slip_fixed} are given by
\begin{align}
    \tilde{\bm{q}}_s^* = \vcttwo{\frac{2 \chi d_x g m \sec(\gamma \psi_e^d)\tan(\gamma \psi_e^d)}{b r_0}}{\psi_e^d - \frac{b \pi \cos(\gamma \psi_e^d)^2\cot(\gamma\psi_e^d)}{4 m \omega_r} + \frac{\chi d_x g m \pi \tan(\gamma \psi_e^d)^2}{b r_0^2 \omega_r}}\label{eq_a_e_star_simple} \,.
\end{align}

Setting $\tilde{\bm{q}}_s^* =  \tilde{\bm{q}}_s^s$ from \cref{eq_specefied,eq_a_e_star_simple} and solving yields
\begin{subequations}
\begin{align}
    \begin{split}
           \Psi_\zeta^{\dagger}(a_e^s, \psi_e^d) \coloneqq \zeta^s_\text{lo} =&- \frac{b \pi \cos(\gamma \psi_e^d)^2\cot(\gamma\psi_e^d)}{4 m \omega_r}\\ &+ \psi_e^d + \frac{a_e^s \pi \sin(\gamma \psi_e^d)}{2 r_0 \omega_r}\,, \label{eq_zeta_r_trans}
    \end{split}\\
    \Psi_d(a_e^s, \psi_e^d) \coloneqq d_x =& \frac{a_e^s b r_0 \cos(\gamma \psi_e^d)\cot(\gamma \psi_e^d)}{2 \chi g m}\,, \label{eq_dx_desired}
\end{align}
\end{subequations}
where, according to \cref{lemma_zeta_r_monotonic}, for each $a_e^s$, $\Psi_\zeta^\dagger$ is a monotone function of $\psi_e^d$, over the domain $\psi_e^d \in [0, \pi/\gamma]$ and, hence,  is the (left) inverse of a well defined function of $\zeta^s$ (over the corresponding codomain) that we denote as  $\Psi_\zeta(a_e^s, \zeta^s)$. 
In practice, we compute $\Psi_\zeta$ by numerical inversion of \cref{eq_zeta_r_trans}. Numerical experience suggests that $\Psi_\zeta$ is uniquely well defined over a substantially larger codomain and domain than the conservative assumptions of \cref{table_invert_reset} allow.

\begin{lemma} \label{lemma_zeta_r_monotonic}
$\Psi_\zeta^\dagger$ \cref{eq_zeta_r_trans} is monotonic  w.r.t $\psi_e^d$ and, hence, invertible for $\gamma \psi_e^d \in [0, \pi]$.
\end{lemma}
\begin{proof}
    The derivative of \cref{eq_zeta_r_trans} is
    \begin{align*}
    \begin{split}
        \frac{\partial \Psi_\zeta^{-1}(a_e^s, \psi_e^d)}{\partial \psi_e^d} =& 1 + \frac{a_e^s \pi \gamma \cos(\gamma \psi_e^d)}{2 r_0 \omega_r} \\&+ \frac{b \pi \gamma \cos(\gamma \psi_e^d)^2}{2 m \omega_r} + \frac{b \pi \gamma \cot(\gamma \psi_e^d)^2}{4 m \omega_r}
    \end{split}
    \end{align*}
    which is always positive for $\psi_e^d$ for $\gamma \psi_e^d \in [0, \pi]$. \qed
\end{proof}

%  which is a reasonable
% 553 assumption in the region of interest leading to a trivial numeric inversion.

% which is transcendental in $\psi_e^d$, but is monotonic for $\gamma \psi_e^d \in [0, \pi]$ (\cref{lemma_zeta_r_monotonic}) which is a reasonable assumption in the region of interest leading to a trivial numeric inversion. 

\subsubsection{Analytical Fixed Points Plots}
\begin{figure*}[tb]
    \centering
    \includegraphics{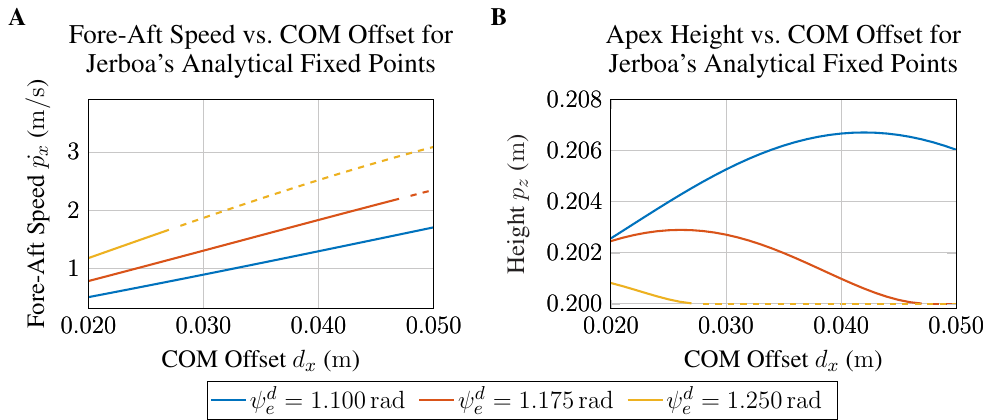}
    \caption{Analytically derived steady state conditions for Jerboa from \cref{eq_3dof_slip_fixed} as a function of the input parameters COM Offset $d_x$ and desired energy ratio $\psi_e^d$. The dashed lines represent fixed points where the apex height is less than the leg length indicating that the robot will be unable to recirculate its leg in flight. See \cref{tab_model_parameters} for model parameters. \textbf{(A)} plots the analytical fore-aft speed. The COM offset $d_x$ has a simple, monotone relationship with fore-aft speed, and increasing $\psi_e^d$ increases the speed. 
    \textbf{(B)} plots the analytical apex height. Decreasing $\psi_e^d$ increases the possible apex heights since it puts more energy in the radial direction. 
    This effect is not quite as simple as in the fore-aft speed plot. Similarly $d_x$ affects the apex height in a trigonometric fashion based on increasing the liftoff leg angle. Eventually $\theta_\text{lo} + \psi_e$ will be large enough that the apex height will be less than the leg length corresponding to the robot stubbing its toe in flight.
    }
    \label{fig_analytical_results}
    \setcounter{subfigure}{0}
    \refstepcounter{subfigure}\label{fig_analytical_speed}
	\refstepcounter{subfigure}\label{fig_analytical_height}
\end{figure*}

In order to lay bare the relationship between $d_x$, $\psi_e^d$, and the apex coordinate fixed points, we plot the analytical fixed points in fore-aft speed and apex height from \cref{eq_3dof_slip_fixed} (converted to fore-aft speed and apex height using \cref{eq_apex_coordinate_change}) in \cref{fig_analytical_results}. 

\cref{fig_analytical_speed} demonstrates that increasing $d_x$ also increases the fore-aft speed in a simple, monotone fashion which matches the intuition from \cref{eq_3dof_slip_fixed}.
Similarly, increasing $\psi_e^d$ increases the fore-aft velocity as it leads to more of the energy being in the fore-aft direction while also having a positive effect on energy $a_e^*$.

On the other hand the effect of $a_e$ and $\psi_e^d$ on height is much more complicated due to the eventual saturation caused by $\theta_\text{lo}+\psi_e^* > \uppi/2$. The violation of this constraint, denoted by a dashed line in \cref{fig_analytical_results}, corresponds to the robot failing to liftoff or not having enough height to recirculate the leg. While increasing $d_x$ will increase the speed at which the robot lifts off, it also increases $\theta_\text{lo}$ without decreasing $\psi_e^*$ which in turn decreases the vertical liftoff velocity. In contrast to the effect of $\psi_e^d$ on $a_e$,  decreasing  the value of the desired energy ratio yields higher apex heights as \cref{fig_hybrid_variables} would imply by directing more energy upward in trigonometric fashion.

% \begin{figure}
%     \centering
%     % \includegraphics[width=\linewidth]{analytical/speed_plot}
%     \input{./figs/analytical/speed_plot.tex}
%     \caption{Analytical results from \cref{eq_ae_star} as a function of input parameters for Jerboa. See \cref{tab_jerboa_parameters} for model parameters.  }
%     \label{fig_analytical_speed}
% \end{figure}

% \begin{figure}
%     \centering
%     % \includegraphics[width=\linewidth]{analytical/height_plot}
%     \input{./figs/analytical/height_plot.tex}
%     \caption{Analytical height from \cref{eq_ae_star} as a function of input parameters for Jerboa. See \cref{tab_jerboa_parameters} for model parameters. Decreasing $\psi_e^*$ increases the height since it puts more energy in the radial direction. As the COM offset increases, the liftoff angle increases while $\psi_e^*$ stays the same. Eventually the robot reaches a point where the vertical velocity is negative at liftoff, meaning the apex height is just the liftoff height.}
%     \label{fig_analytical_height}
% \end{figure}

% \input{03_02_attitude_ctrl}
\section{Implementation Details}\label{sec_detail}
As with any real-world system, there are differences between the theoretical models covered in the previous sections and reality.
The controllers presented and analyzed in \cref{sec_model,table:controllers_math} do not completely specify the software implementations in simulation and hardware. 
This section details practicalities associated with anchoring the idealized 3 DoF  Lagrangian dynamics in physical motors, joints, and links along with more substantive modifications made to the controllers to improve their reliability on a real system.

\begin{table*}[ht]
\centering
\small\sf%\centering
\caption{Controller overview listing: (a) stance mode hip torque, stride event hip torque offset, and stepping policies studied formally in \cref{sec_math}; and (b) formally unnecessary, but empirically implemented stride event terms added to increase basin of attraction as detailed in \cref{sec_basin_increase}. A block diagram of the control strategy can be found in \cref{fig_robot_diagrams,fig_hybrid_block_diagram}.}
\label{table:controllers}
\begin{tabular}{>{\raggedright\arraybackslash}p{6.5cm}
ll}
\toprule
\textbf{Nomenclature}&\textbf{Equation} & \textbf{Ref.}\\ 
\midrule
\multicolumn{3}{l}{(a) Formal Stance and Stride Controllers (See \cref{sec_math} for stability analysis)}\\
\midrule
Continuous time stance mode hip torque & $ \tau\coloneqq \bar\tau r - m g r \sin \theta - \epsilon k_p (\phi^d - \phi) + \sqrt\epsilon k_d \dot\phi$ & \cref{eq_torque}  \\
Discrete stride event hip torque bias & $\bar\tau_{k+1} = \bar\tau_{k} + \epsilon^2 k_\tau (\phi_{\text{lo},k} - \phi^d)$ & \cref{eq_discrete_torque}  \\
Discrete stride event stepping controller & $\theta_{\text{td}}(\z) \coloneqq (1- \epsilon \alpha) \tilde \theta_{\text{td}}(\z) + \epsilon \alpha \theta_{\text{td},\text{prev}}$ & \cref{eq_stepping_filt} \\
& $    \tilde \theta_{\text{td}}(\z) \coloneqq - \theta_\text{lo} +  \epsilon k_e (\psi_e^d - \psi_e) - \frac{ \uppi}{\omega_e(\bm a)}$ & \cref{eq_stepping_pre_filt}\\
\midrule
\multicolumn{3}{l}{(b) Supplemental Stride Controllers (See \cref{sec_detail} for details) }\\
\midrule
Discrete stride event energy loop & $d_{x,k+1} = d_{x,\text{ff}} + k_{p,a_e} \left(a_{e}^d - a_{e,k}\right) - k_{d,a_e} \left( a_{e_k}-a_{e,k-1} \right)$ & \cref{eq_ae_controller}  \\
Discrete stride event hip torque w/ $d_x$ varying term & $ \bar\tau_{k} = \tau_{\text{ff,k}} + f(d_x)$ & \cref{eq_hard_tau0}  \\
 & $ \tau_{\text{ff},k+1} = \tau_{\text{ff},k} + \epsilon^2 k_\tau (\phi_{\text{lo},k} - \phi^d)$ & \cref{eq_torque_imp_integrator}  \\
\bottomrule
\end{tabular}

\setcounter{subtable}{0}
\refstepcounter{subtable}\label{table:controllers_math}
\refstepcounter{subtable}\label{table:controllers_imp}
\end{table*}

\subsection{Anchoring the Center of Mass Controller}
As depicted in \cref{fig_robot_diagrams}, controlling the pitch-unlocked robot to a target steady state hopping height and fore-aft speed (characterized by fixed point energy $a_e^*$ and energy ratio $\psi_e^*$) is achieved by adjusting the center of mass offset $d_x = d \cos \phi^d$ (i.e., the $x$ projection of the vector from hip to COM at target pitch).
The strategy for changing this value, while usually simple, will likely differ between robot morphologies, as illustrated by the contrast between the 5-link and the tailed biped examples comprising the application focus of this paper.

\subsubsection{5-Link Biped}
Many robots (e.g., the biped in \cref{fig_planr_biped}) do not have the ability to internally relocate their center of mass relative to their hip. Instead, they can change their target pitch. Leaning the robot’s torso forward or backward changes $d_x$ , thus, its steady state translational energy according to the fixed point formulae \cref{eq_3dof_slip_fixed}. This reveals one of the several limitations arising from an underactuated machine whereby the present allocation of control affordance locks in a dependence between the desired CoM energy and the requisite steady state pitch.

Additionally, while Jerboa's legs are spring dampers, the 5-link biped has rigid legs and actuated knees. For simulations with the 5-link biped, the legs are controlled to act like virtual spring dampers. 
Furthermore, the 5-link biped uses a bipedal gait, alternating which leg touches the ground from step to step.

\subsubsection{Tailed Biped}
In the case where the robot has a ``shape'' degree of freedom (e.g., a spine, tail, or arm), it can move its COM relative to the hip. While this paper focuses on the Penn Jerboa, a  tailed biped, this idea is relevant to any robot with a ``shape'' degree of freedom. For Jerboa, increasing the tail angle (i.e., moving the tail forward and up) increases the steady state energy by increasing $d$ and in turn $d_x$.

\subsection{Increasing the Basin}\label{sec_basin_increase}
\begin{figure}[thb]
    \centering
    \includegraphics[width=0.95\linewidth]{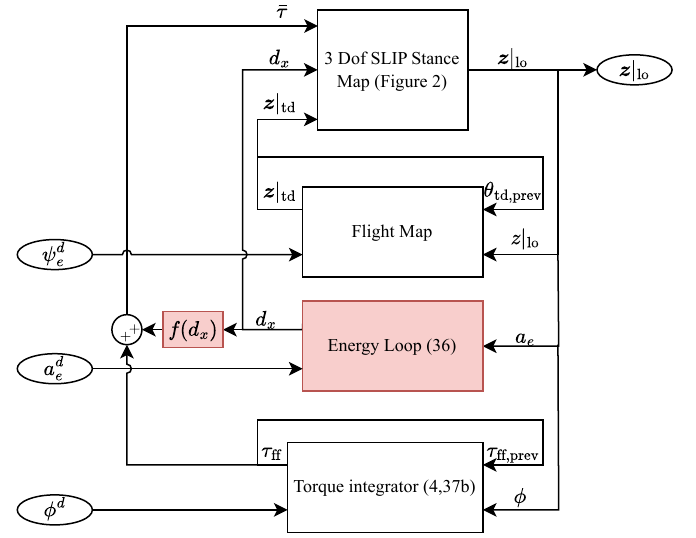}
    \caption{A block diagram describing the implemented discrete hybrid control system. The blocks in red are formally unnecessary, but introduced as an empirical measure to adjust the commanded setpoints so as to  bring the imperfectly selectable and highly variable  experimental initial conditions into the basin of attraction. They have no influence on the resulting location of the fixed points and hence are not considered in the formal analysis. \Cref{fig_robot_diagrams} contains the control block diagram for the continuous time stance controllers.}
    \label{fig_hybrid_block_diagram}
\end{figure}

One chief obstacle in testing (and, of course, to any useful implementation) is that experiments with Jerboa begin with a manual, inconsistent toss. Notwithstanding our moderate success (where given a careful throw the robot achieves steady state hopping) in applying the theoretical controller (\cref{table:controllers_math}), the variability in the toss made it challenging to explore the full range of setpoints.
Therefore, it is desirable to increase the basin of attraction for empirical testing through two modifications to the theoretical controller (\cref{table:controllers}), presented in \cref{sec_energy_loop,sec_feedfwd_explain} \endnote{\DIFadd{Formal analysis of the basin of attraction of the original and the modified system is inhibited by the lack of tools in the literature for characterizing the basin of a nonlinear hybrid dynamical system of the sort we are working with.}}.
These modifications are minor enough that trends predicted by the fixed points are not broken, as revealed in \cref{sec_sim_results,sec_hard_results}.
Although the simulated trials have consistent initial conditions, the modifications presented here are used in both simulation and hardware to permit direct comparison.
\Cref{fig_hybrid_block_diagram} shows how the added controllers change the block diagrams for the hybrid system.

%While the theoretical controller described in \cref{sec_model} in \cref{eq_stepping,eq_torque,eq_discrete_torque} (the hybrid discrete event touchdown angle control, discrete hip torque control, and the continuous within stance hip torque control, respectively) can generate analytical results (\cref{fig_analytical_results} and the analytical plots in \cref{fig_sim_height_speed,fig_hard_height_speed}), it is desirable to increase the basin of attraction for empirical testing.

\subsubsection{Hybrid Discrete Event Energy Controller}\label{sec_energy_loop}
For the controller described in \cref{sec_model} by \cref{table:controllers_math}, $d_x$ is constant at all times. During the hardware implementation the variability of initial conditions arising from the starting throw would often lead to the robot running out of energy with a fixed $d_x$. Leveraging the monotonic relationship between $d_x$ and $a_e^*$ \cref{eq_3dof_slip_fixed}, the implementation in simulation and hardware changes $d_x$ between steps, controlling to a target energy $a_e^d$ using a discrete PD and feedforward controller based on the state at liftoff: 
\begin{equation}
    d_{x,k+1} = d_{x,\text{ff}} +  k_{p,a_e} \left(a_{e}^d - a_{e,k}\right) - k_{d,a_e} \left( a_{e_k}-a_{e,k-1}\right)\,,\label{eq_ae_controller}
\end{equation}
where energies $a_e$ are values at the current and previous liftoffs $a_{e,k}$ and $a_{e,k-1}$, respectively, and $a_{e}^d$ is the target energy. 
    The constant feedforward term $d_{x,\text{ff}}$ is simply the initial value of $d_x$. We include a derivative term to limit overshoot which would occasionally push the system outside the basin of attraction.
This controller allows direct and intuitive control of the system's energy and helps to stabilize the system in cases where the energy is decaying due to an inadequately energetic initial throw.

It is important to note that since we are wrapping our initial controller with a simple PD loop, the fixed point properties of the lower dimensional system are unchanged for a given $d_x$ \cref{eq_3dof_slip_fixed}, and instead only the eigenvalues and region of attraction are changed.
Therefore, the simulation and hardware results are presented in terms of the final $ d_{x,\infty} \approx \lim_{k \rightarrow \infty} d_{x,k}$ and $\psi_e^d$ as if they were control inputs, intentionally similar to the presentation in \cref{fig_analytical_results}.

Notwithstanding its empirical convenience,  and the invariance of the resulting steady state hopping behavior, finding conditions on control gains sufficient to insure the stability of this hardware implementation may not be straightforward and lies well beyond the scope of the present paper. This is because the discrete energy loop \cref{eq_ae_controller} violates the cascade composition structure assumed by the hybrid discrete return map analyzed in \cref{sec_decomp_slip} (though not the continuous time stance dynamics) as now the input to the pitching subsystem $d_x$ depends on an output of the 2 DoF SLIP subsystem $a_e$. 

\subsubsection{Feeding Forward COM Offset into Hip Torque}\label{sec_feedfwd_explain}
\cref{eq_pitch_fixed_points} describes a monotonic relationship between $d_x$ and $\bar\tau^*$. When $d_x$ was fixed, the initialization of the $\bar\tau$ integrator was a guess of what $\bar\tau^*$ would be. Once $d_x$ was changing due to the energy control loop \cref{eq_ae_controller}, it became beneficial to add a $d_x$ varying feedforward term to the $\bar\tau$ update function in an attempt to invert \cref{eq_pitch_fixed_points}.

This new discrete hip torque---updated in flight at the same time as $d_x$ in \cref{eq_ae_controller}---takes the form of
\begin{subequations}\label{eq_torque_imp_integrator}
\begin{align}
    \bar\tau_{k} &= \tau_{\text{ff,k}} + f(d_x)\,,\label{eq_hard_tau0} \\
    \tau_{\text{ff},k+1} &= \tau_{\text{ff},k} + \epsilon^2 k_\tau (\phi_{\text{lo},k} - \phi^d)
\end{align}
\end{subequations}
where $\tau_\text{ff}$ is a rebranded variable describing just the integrator on pitch as described in \cref{eq_discrete_torque}, and $f(d_x)$ is a tuned linear function of $d_x$ which attempts to invert \cref{eq_pitch_fixed_points} without relying on the model parameters. We did not need to exactly invert \cref{eq_pitch_fixed_points} as the system still had a pitch integrator driving $\bar\tau$ to $\bar\tau^*$ according to the LTI stability of the pitching subsystem \cref{sec_pitch_stable}. 

As before, the discrete hip torque $\bar\tau_{k}$ from \cref{eq_hard_tau0} is substituted normally into \cref{eq_torque} to yield the continuous time hip torque control policy. 
Due to the integration from \cref{eq_torque_imp_integrator}, the fixed point torque $\bar\tau^*$ produced by \cref{eq_torque_imp_integrator} is equivalent to that analyzed in \cref{sec_math} for a fixed $d_x$. 
This change resulted in more trials successfully stabilizing in simulation and hardware. In contrast to discrete energy controller discussed in \cref{sec_energy_loop}, this controller is analytically tractable under our present analysis. The logic for omitting it is that in our model $d_x$ is constant and for a constant $d_x$ \cref{eq_torque_imp_integrator} simply describes the initial condition of the torque integrator. 

\subsection{Energy Ratio Estimation}\label{sec_psie_est}
Although knowing the energy ratio $\psi_e$ at liftoff is important for the stepping controller in \cref{eq_stepping}---specifically in the term from \cref{eq_stepping_pre_filt}---it is difficult to measure. Velocity is notoriously challenging to accurately measure, and the issue is only compounded since $\psi_e$ depends on the ratio of two different velocity measures. 
Even worse, both terms in $\psi_e$ undergo drastic changes directly after liftoff. Thus, any error in liftoff detection causes a large error in $\psi_e$.
% We will discuss the exact effect of these challenges using results from simulation data in \cref{sec_sim_time_traces}.

For implementation we estimate the energy ratio by replacing the exact value of the leg's velocities $\dot{\theta}$ and $\dot{r}$ in \cref{eq_hybrid_vars} with their mean values  $\hat{\dot{\theta}}$ and $\hat{\dot{r}}$.  Where
\begin{align*}
   \hat{\dot{\theta}} &= \frac{\theta_\text{lo}-\theta_\text{td}}{t_\text{lo}-t_\text{td}} \,, \\
    \hat{\dot{r}} &= \frac{r_\text{lo}-r_\text{b}}{t_\text{lo}-t_\text{b}}\,,
\end{align*}
and $\square_b$ denotes a value measured at bottom.
    This results in an estimated energy ratio $\hat{\psi}_e$ calculated as
    \begin{equation}\label{eq_psie_est}
        \hat{\psi}_e \coloneqq \arctan \frac{\hat{\dot{\theta}} r}{\hat{a}_v}\,,
    \end{equation}
    where $\hat{a}_v = \sqrt{{\hat{\dot{r}}^2 + (r_0-r)^2\omega_r^2}}$. This estimate is updated only at liftoff when it is required by the stepping controller.
    
% While the simulation results in \cref{sec_sim_time_traces} show that it is possible to directly measure the energy ratio and  it to avoid disturbances at liftoff, this approach is less likely to work on hardware, where the sensors are noisy even with filtering.
    Due to the imprecise detection of liftoff and the sensor noise in hardware, the simulation and hardware stepping controller in \cref{eq_stepping} uses the energy ratio estimate $\hat{\psi}_e$ in place of a measured value for $\psi_e$ to calculate the touchdown angle.

\subsection{Finding Free Model Parameters}
\label{sec:model_params}

The fixed point formulae presented in \cref{eq_3dof_slip_fixed} include expressions that depend upon two free parameters associated with the simplifying assumptions of \cref{table_decomp_assumptions,table_hybrid_averaging_assumptions} whose values must be determined empirically: $\gamma$ from \cref{ass_psi_e_replacemt} describes how close to $\psi_e^d$ the mean value of $\psi_e$ is over the course of stance, and $\chi$ from \cref{ass_pitch} describes the mean force from the leg. We found the values of $\gamma$ and $\chi$ by numerically fitting the fixed points from \cref{eq_3dof_slip_fixed} to the simulation (\cref{fig_sim_height_speed}) and hardware (\cref{fig_hard_hybrid_params}) fixed points. The values for $\gamma$ and $\chi$ are in \cref{tab_model_parameters}. While $\gamma$ was the same in simulation and hardware, we found the hardware required a higher value of $\chi$ due to the kinematics of the hardware Jerboa's leg as discussed in \cref{sec_hard_sim_diff}.

\subsection{Value of \texorpdfstring{$\epsilon$}{epsilon}}
Throughout the theory section we discussed the parameter $\epsilon$ used formally in \cref{lemma_cascade_stability}. For the sake of implementation, we set $\epsilon = 1$, though in the controllers (\cref{table:controllers}) the value of $\epsilon$ does not matter since it is always multiplied by a gain.
\section{Simulation Results}\label{sec_sim_results}
This section presents both representative trajectories and fixed point data resulting from numerical simulation of the planar 5-link biped and Jerboa in Gazebo.  The fixed point data was obtained from recording steady state values achieved following convergence of initial transients given a fixed desired energy $a_e^d$ and energy ratio $\psi_e^d$. See \cref{tab_model_parameters} for model parameters. The simulation models match the analytical models from \cref{fig_robot_diagrams} with the exception of having a light $\sim \SI{100}{g}$ toe instead of a massless toe. The control loop is then executed at \SI{1}{kHz}. 

\begin{table}[tbh]\centering
    \caption{Model parameters for Jerboa and the 5-link biped.}
    \label{tab_model_parameters}
    \begin{tabular}{l l r r}
        \toprule
        \multicolumn{2}{l}{Parameter} & Jerboa Values & 5-link Biped Values\\
        \midrule
        $m_b$ & ($\si{kg}$) & 3.242 & 3.5\\
        $m_t$ & ($\si{kg}$) & 0.247 & --- \\
        $l_t$ & ($\si{kg}$) & 0.4 & --- \\
        $I$ & ($\si{kg.m^2}$) & 0.04 & 0.08 \\
        $d$ & ($\si{m}$) & 0.05 & 0.05 \\
        $r_0$ & ($\si{m}$) & 0.2 & 0.27 \\
        $k$ & ($\si{N/m}$) & 4000 & 3500 \\
        $b$ & ($\si{N.s/m}$) & 40 & 32 \\
        $\gamma$ & & 0.9 & -- \\
        \multicolumn{2}{l}{$\chi$ (sim)}
         & 1.7 & --\\
        \multicolumn{2}{l}{$\chi$ (hardware)} & 2.4 & --\\
        \bottomrule
    \end{tabular}
\end{table}

\subsection{Energy Comparison}
\begin{figure}[tbh]
        \centering
        \includegraphics[width=\linewidth]{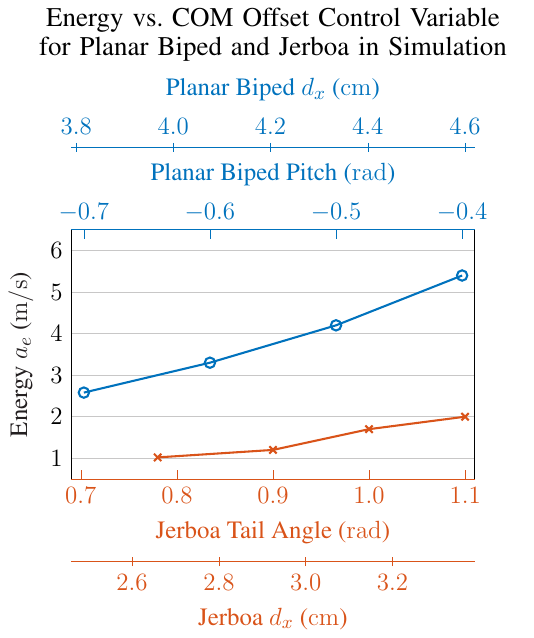}

        \caption{Energy at liftoff versus Jerboa's tail angle and the 5-link biped's pitch for $\psi_e^d = \SI{1.2}{rad}$ in simulation.
        Recalling  that the state coordinate, $a_e$ \cref{eq_hybrid_vars}, expresses COM energy in units of square-root Joules per kg, this plot illustrates that the controller's  affordance through $d_x$ is sufficient to achieve a fourfold increase in energy in joules for both morphologies. Note that values of $d_x$ are not comparable between architectures due to differing parameter values as listed in \cref{tab_model_parameters}.}
        \label{fig_energy_architecture}
    \end{figure}

One of the key predictions of our controller (\cref{table:controllers}) is that moving the COM forward relative to the hip increases the steady state energy $a_e$ \cref{eq_3dof_slip_fixed}.
\cref{fig_energy_architecture} validates this relationship by plotting plotting $a_e$ vs.\ Jerboa's tail angle and the 5-link biped's target pitch.
By increasing $d_x$ the controller is able to to vary over a fourfold range the steady state energy (in joules, twofold range for twice the mass-specific square root of energy) for both morphologies.
Therefore, the following results present the fixed points as a function of the input, $d_x$, rather than in terms of tail angle or target pitch to be independent of the anchoring controller. 

\subsection{Representative Steady State Trajectories}\label{sec_sim_time_traces}
\begin{figure*}[tbh]
    \centering
    % {\Large Simulated Jerboa Time Domain Traces for $a_e^d=\SI{2.3}{m/s}$ and $\psi_e^d=\SI{1.2}{rad}$}\par\medskip
    \includegraphics{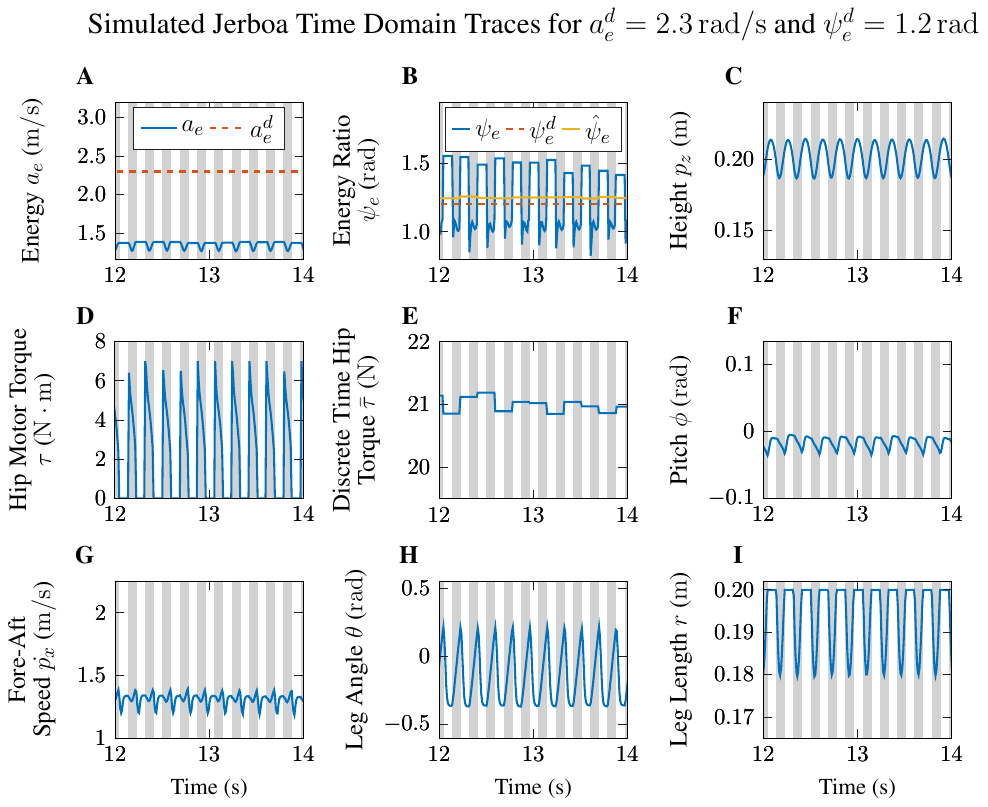}
    \caption{Example trajectories of various state values for Jerboa in simulation during steady state locomotion. Periods of stance are highlighted with a gray background. Note that \cref{fig_sim_time_phase} presents both the energy ratio $\psi_e$ calculated using \cref{eq_hybrid_vars} as well as the estimated energy ratio $\hat{\psi}_e$ from \cref{eq_psie_est}. Only the estimate is used in application of the stepping controller. For model parameters, see \cref{tab_model_parameters}.}
    \label{fig_sim_time}
    \setcounter{subfigure}{0}
    \refstepcounter{subfigure}\label{fig_sim_time_energy}
    \refstepcounter{subfigure}\label{fig_sim_time_phase}
    \refstepcounter{subfigure}\label{fig_sim_time_height}
    \refstepcounter{subfigure}\label{fig_sim_time_torque}
    \refstepcounter{subfigure}\label{fig_sim_time_tau0}
    \refstepcounter{subfigure}\label{fig_sim_time_pitch}
    \refstepcounter{subfigure}\label{fig_sim_time_dx}
    \refstepcounter{subfigure}\label{fig_sim_time_theta}
    \refstepcounter{subfigure}\label{fig_sim_time_r}
\end{figure*}
With the controller effecting the expected trends in $a_e$, \cref{fig_sim_time} presents a set of time domain data from a simulated Jerboa.
\cref{fig_sim_time_energy} displays that the energy $a_e$ from \cref{eq_hybrid_vars} converges to a roughly fixed value at liftoff, while also decreasing and increasing during stance (plotted with a gray background in the figure).  Note that the target energy $a_e^d$ is not achieved because the energy controller in \cref{eq_ae_controller} lacks an integrator and the effect of gravity in \cref{eq_pitch_fixed_points} acts as a disturbance on the linear relationship between $a_e$ and $d_x$ that the energy controller \cref{eq_ae_controller} is operating on. Even so, the controller properly controls energy to a fixed value at liftoff, validating previous mathematical analysis.

\cref{fig_sim_time_phase} renders both the actual energy ratio $\psi_e$ as well as its estimate $\hat{\psi_e}$ alongside the target $\psi_e^d$.
Before analyzing the time domain trace itself, it is important to highlight the rationale for using an
estimate (rather than the simulation ``ground truth'' value), paying special attention to the values of each line during stance (in gray) and in flight (in white).
    As \cref{fig_sim_time_phase} reveals, while the energy ratio $\psi_e$ does increase over the course of stance (while oscillating), it rapidly spikes right before liftoff.
    This spike is largely due to imprecise liftoff detection as mentioned in \cref{sec_psie_est}: Jerboa detects liftoff after it really happens.
    Therefore, during the supposed end of ``stance,'' Jerboa's legs are free to move, allowing the legs' angular velocity $\dot{\theta}$ to increase rapidly while $a_r$ quickly goes to 0 due to the robot being in the air. These two phenomena cause $\psi_e$ to spike at liftoff \cref{eq_hybrid_vars}.
    Thus, the energy ratio at liftoff is measured in the range of $1.4$--$\SI{1.6}{rad}$ despite consistently staying closer to $\SI{1}{rad}$ midway through stance. The value of $\psi_e$ right before liftoff (partway through the spike) is likely its true value at liftoff.
Unfortunately, the stepping controller \cref{eq_stepping} depends heavily on the value of $\psi_e$ precisely at liftoff. It is for this reason that the controller uses the discretely estimated energy ratio $\hat{\psi}_e$ from \cref{eq_psie_est} instead of $\psi_e$ itself.

The estimated energy ratio $\hat{\psi_e}$ lines up with a value of $\psi_e$ during its spike at liftoff. Furthermore, it experiences a much smaller level of variation than the measured $\psi_e$ at liftoff, making $\hat{\psi}_e$ more in line with the level of variation in the measured $\psi_e$ midway through stance.   Thus, we study the estimate and not the measurement.
\cref{fig_sim_time_phase} demonstrates that the controller keeps the estimated energy ratio $\hat\psi_e$ near the desired energy ratio $\psi_e^d$. This is in contrast to the theory \cref{eq_3dof_slip_fixed} which predicts that $\hat\psi_e^* = \psi_e^d$.

The plot of the height over time in \cref{fig_sim_time_height} follows a very steady trajectory which suggests that $\psi_e$ is controlled to a roughly fixed value.
As shown in \cref{fig_hybrid_variables}, changes in $\psi_e$ would result in different apex heights for a given energy $a_e$.
Because the variance in height is small, the variance in the energy ratio $\psi_e$ is also acceptable.
    
\Cref{fig_sim_time_torque,fig_sim_time_tau0} depict the hip torque from \cref{eq_torque} and discrete time hip torque from \cref{eq_discrete_torque}, respectively.
    As expected, the hip torque decreases over the course of stance because the gravity compensation term $-mgr\sin{\theta}$ decreases as the leg angle $\theta$ approaches zero. 
        The other two terms focused on controlling the pitch are rather small since the pitch is close to its target of $\SI{0}{rad}$, as shown in \cref{fig_sim_time_pitch}.
    Meanwhile, the discrete time hip torque $\bar\tau$ in \cref{fig_sim_time_tau0} varies around only $\SI{1}{\percent}$ per step, evidence that the controller is achieving a fixed $\bar\tau$.
    
    The plot of leg angle $\theta$ in \cref{fig_sim_time_theta},  illustrates the expected level of asymmetry between touchdown and liftoff angles,  validating \cref{ass_xi_theta_replacement}.
    Similarly the change in gravitational potential energy in flight from the steady state stepping asymmetry is small compared to the steady state speed (\cref{fig_sim_time_dx}) validating \cref{ass_flight}
    
Overall, the controller is very successful at managing the transient as well as steady state behavior of Jerboa in simulation in a manner that matches the analytical predictions. Energy $a_e$ is properly regulated with the hip torque control law, and the energy ratio, $\psi_e$, is properly controlled with the stepping strategy \cref{eq_stepping} specified  in \cref{table:controllers}.

\subsection{Fixed Points and Model Accuracy}\label{sec_sim_fixed_points}
    \begin{figure*}
        \centering
	    \includegraphics{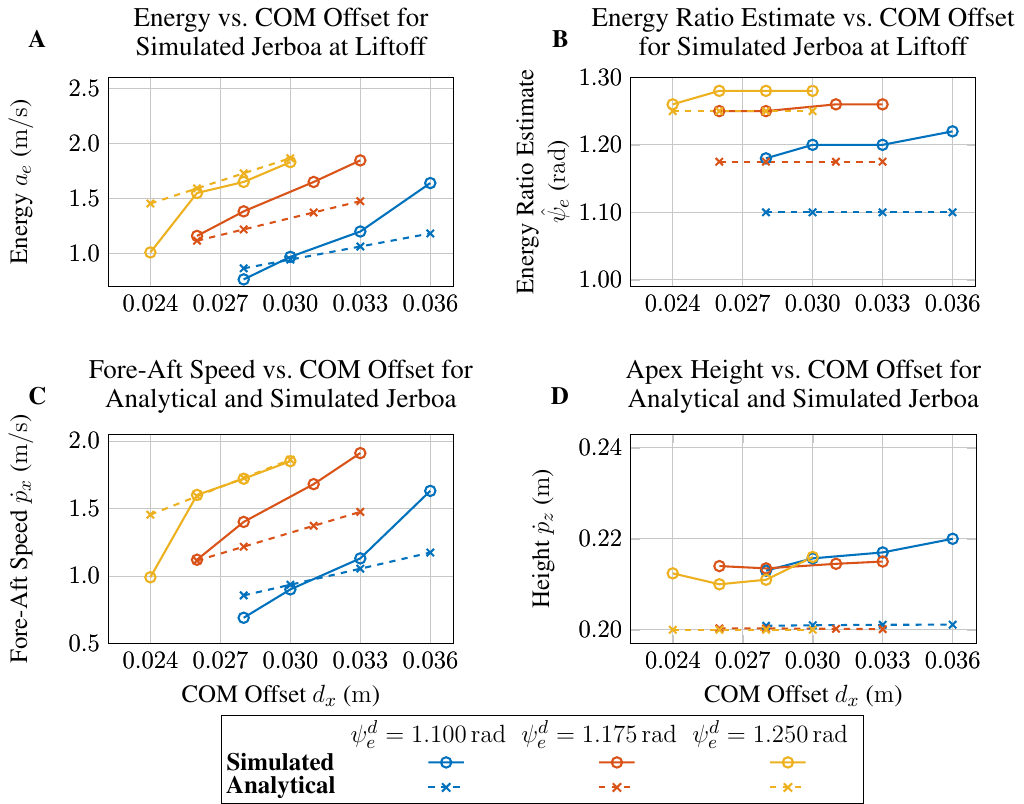}
	    \caption{Fixed points of Jerboa in simulation compared to those predicted by the analytical model in \cref{eq_3dof_slip_fixed} and plotted in \cref{fig_analytical_results}. 
	    \textbf{(A)} plots the energy and demonstrates that increasing $d_x$ tends to increase energy. \textbf{(B)} shows the energy ratio estimate $\hat{\psi}_e$. Increasing the target energy ratio increases the achieved values, though the estimate is always higher than the target. Furthermore, the energy ratio is roughly constant for a given target independent of the COM offset.
	    \textbf{(C)} displays the fore-aft speed. As predicted by the model, increasing the COM offset $d_x$ increases the speed, and increasing the target energy ratio $\psi_e^d$ also increase the speed. \textbf{(D)} shows the apex height. In line with the model, increasing the target energy ratio $\psi_e^d$ also increases the height. The model consistently underestimates the apex height. For mean and maximum error see \cref{tab_sim_accuracy}, and for model parameters see \cref{tab_model_parameters}. }
	    \label{fig_sim_height_speed}
	    \setcounter{subfigure}{0}
	    \refstepcounter{subfigure}\label{fig_sim_energy}
	    \refstepcounter{subfigure}\label{fig_sim_phase}
        \refstepcounter{subfigure}\label{fig_sim_speed}
	    \refstepcounter{subfigure}\label{fig_sim_height}
    \end{figure*}
\begin{table*}[tbh]\centering 
    \caption{Accuracy of the analytical apex coordinate fixed points from \cref{eq_3dof_slip_fixed} compared to the simulation's apex coordinate fixed points for Jerboa. $\pm$ denotes the standard deviation. The data is from the large operating regime explored in  \cref{fig_sim_height_speed} where the number of different desired and numerically recorded return map steady state values (following hundreds of stance map integrations from various initial conditions) is  $N = 12$.}
    \label{tab_sim_accuracy}
    \begin{tabular}{l l l l l}
        \toprule
        State       & {\begin{tabular}[c]{@{}c@{}} Mean  \\  Magnitude of Error \end{tabular}}   & {\begin{tabular}[c]{@{}c@{}} Mean Percent \\ Error \end{tabular}}  & Max Error & {\begin{tabular}[c]{@{}c@{}}Max Percent \\ Error \end{tabular}}  \\
            %   &             &  Percent Error &           & Error \\%\hline
        \midrule
        $\dot{p}_x$      & $\SI{0.179\pm0.188}{m/s}$ & $\SI{13.8\pm14.6}{\percent} $& $\SI{0.463}{m/s}$   & $\SI{46.8}{\percent}$   \\ %\hline
        $p_z$            & $\SI{0.014\pm0.002}{m}$   & $\SI{6.5\pm1.05}{\percent} $ & $\SI{0.018}{m}$    & $\SI{8.6}{\percent}$   \\ %\hline
        \bottomrule
    \end{tabular}
\end{table*}

        % $\dot{x}$      & $\SI{0.176}{} \pm \SI{0.171}{m/s}$ & $\SI{13.5}{} \pm \SI{12.4}{\percent}$      & $\SI{0.488}{m/s}$   & $\SI{34.8}{\percent}$   \\ %\hline
        % $z$            & $\SI{0.011}{} \pm \SI{0.002}{m}$   & $\SI{5.5}{} \pm \SI{1.1}{\percent}$       & $\SI{0.016}{m}$    & $\SI{8.0}{\percent}$   \\ %\hline

Taking an input output perspective to verifying the model presented in \cref{sec_model}, we compare the simulated apex coordinate fixed points to the analytical apex coordinate fixed points from \cref{eq_3dof_slip_fixed}. We collect the data by ``throwing'' the robot in simulation before waiting for its convergence to a near steady state behavior. Once the robot reaches steady state we collect data for a few seconds to reject any small step to step variations in the behavior.
We choose the range of fixed points to be similar to the operating regime explored by the hardware fixed points in \cref{sec_hard_accuracy}.
Through comparison of simulated data to analytical fixed points, the results reveal that the simulation behaves as the model would expect. \Cref{fig_sim_energy} demonstrates that as the model predicts, increasing $d_x$ or increasing $\psi_e^d$ increases $a_e$. On the other hand \cref{fig_sim_phase} demonstrates that while the simulated robot is controlling the steady state energy ratio $\psi_e^*$ independent of $d_x$, there is some steady state error in energy ratio not predicted by the model \cref{eq_3dof_slip_fixed}.
\Cref{fig_sim_speed} displays the fore-aft speed vs.\ the input parameters. Not only are the trends predicted by the model present, but the model also does a excellent job predicting the value of the fixed point.
 Next the apex height vs.\ input parameters is plotted in \cref{fig_sim_height}.
The model effectively predicts that an increasing energy ratio leads to greater height, yet it consistently underestimates the actual apex height. This discrepancy arises from the model's predicted liftoff angle, which tends to be closer to vertical compared to the liftoff angle observed in simulation. As a consequence, the simulated liftoff velocity angles are shallower when compared to the model's liftoff velocity angles.
    
Despite these differences, the model does usefully predict the simulation results. 
As summarized in \cref{tab_sim_accuracy}, the model predicts the the simulated robot's fore-aft speed and height with a mean percent error in speed of $\SI{13.8}{\percent}$ and a mean percent error in height of $\SI{6.5}{\percent}$.
While the mean percent error is small, the coefficient of variation in apex speed is quite high at almost $1$. This high coefficient of variation is a consequence of the choice of $\gamma$ and $\chi$ as outlined in \cref{sec:model_params}. Selecting different values for $\gamma, \chi$ that do not try to minimize the mean error; for example using $\gamma = 0.85$, $\chi = 1.9$, results in the mean percent error for speed being $\SI{19.5\pm8.77}{\percent} $ which has a coefficient of variation around $0.5$.

\subsection{Ground Reaction Forces}
\label{sec_vpp}
\begin{figure}[tbh]
    \centering
    \includegraphics{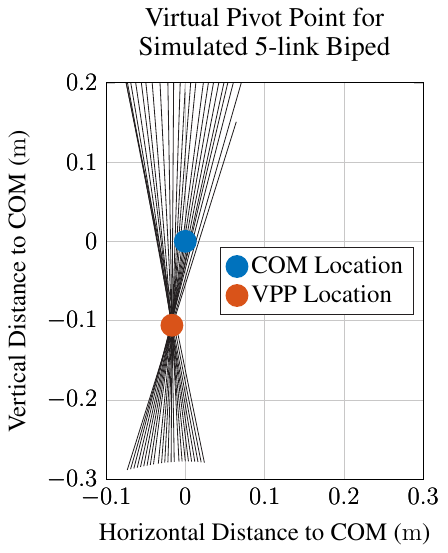}
    \caption{A VPP plot similar to the one presented in \cite{maus_upright_2010}, where the COM is always at the center of the coordinates and the forces close to the beginning and ending of stance are omitted. 
    Each line represents a time-stripped ground reaction force beginning at the point of contact relative to the COM and pointing in the direction of the applied force.
    These ground reaction forces were generated from a single stance of the 5-link biped in \cref{fig_planr_biped}. Note that the forces clearly converge and create a VPP below the COM.}
    \label{fig_vpp_sim}
\end{figure}

    The ground reaction forces of animal (including human) runners appear to converge upon a single point on the body above the COM, termed the Virtual Pivot Point (VPP) \citep{maus_upright_2010}.
    Some extant controllers explicitly control the hip torque with the aim of creating a VPP in the belief that these forces cause a system to mimic a physical pendulum, enforcing stability \citep{firouzi_tip_2019,vu_control_2017,sharbafi_robust_2013}.
    
The summary of our control policy displayed in \cref{table:controllers} reveals many similarities to VPP-based controllers, especially in regards to its  use of hip torque to stabilize pitch, but it does not explicitly attempt to create a VPP with its ground reaction forces. 
Even so, the controller appears to create a VPP across a wide range of operating conditions for both the simulated Jerboa and 5-link biped. 
\Cref{fig_vpp_sim} shows the ground reaction forces of one stance from the 5-link biped in simulation, where a VPP is clearly visible, although it is placed below the COM, as opposed to above it as in \cite{maus_upright_2010}.
One potential explanation for this pattern is that the 5-link biped's torso is not as upright as would be expected in humans, which could cause deviations in the location of the VPP \citep{muller_force_2017}.
Accounting for the intriguing appearance of a VPP (as well as the discrepancy in its locus relative to animal data) lies beyond the scope of this paper, but strikes us as an important question for future study.  The most parsimonious hypothesis arising from our present analysis would hold that  the VPP is a consequence rather than a
goal of animals'   locomotion controllers \citep{maus_upright_2010}. However,  there is no reason to reject without further analysis the intriguing possibility that some appropriate strengthening of such a  strategy might well yield asymptotically stable hip energized running in a hardware biped. 
\section{Hardware Results}\label{sec_hard_results}
% plot Griffon wants:

% energy at steady state
% phase at steady state
% pitch at steady state
% effort at steady state

This section details the results of applying the controller from \cref{table:controllers} to the physical Jerboa robot \cref{fig_jerboa} \citep{shamsah_analytically-guided_2018}. The hardware version of Jerboa \cref{fig_jerboa} was the basis for the model parameters in \cref{tab_model_parameters} so the same parameters also apply here, though there are some additional details in the hardware implementation.
    First, a $\SI{2}{m}$-long boom\endnote{Unlike many booms, the boom used in this paper does not have a counterweight, so its weight must be supported by the robot throughout each stance. Similarly the boom does not have a parallelogram linkage meaning the robot will roll slightly during stance.} constrains Jerboa to movement in the fore-aft plane while allowing for pitching motions. This replicates the situation in simulation. The boom has two rotational encoders which measure the $x$ and $z$ positions of the robot. 
    Additionally, Jerboa's legs are not pogo sticks as in the simulation. Instead they consist of four bar linkages with an extension spring \citep{shamsah_analytically-guided_2018}. This not only causes the spring to act nonlinearly, but also creates non-axial forces while compressed, largely in the forward direction. Additionally, because the hopping controller does not allow the robot to start from rest, hardware experiments begin with a gentle toss, creating some variability in the initial state, necessitating a wide basin of attraction.
    
Despite these implementation differences, this section reveals that the controller achieves similar (though not necessarily equal levels of) success on hardware as in software, following the same trends with regards to COM offset and achieving steady state.
Unsurprisingly, hardware Jerboa's fixed points in \cref{fig_hard_hybrid_params} do not follow the analytical model's fixed point from \cref{eq_3dof_slip_fixed} as well as the simulation's fixed points in \cref{fig_sim_height_speed}  due to these hardware-specific modeling inaccuracies (in particular, the four-bar legs and boom disturbances). In addition to the plots and analysis in this section, \url{https://youtu.be/HhdYK-vSufY} is a video of Jerboa hopping using the controller where the robot demonstrates steady state hopping and hopping over minor obstacles. 

\subsection{Energy and Energy Ratio Control Validation}
\begin{figure*}
    \centering
    \includegraphics{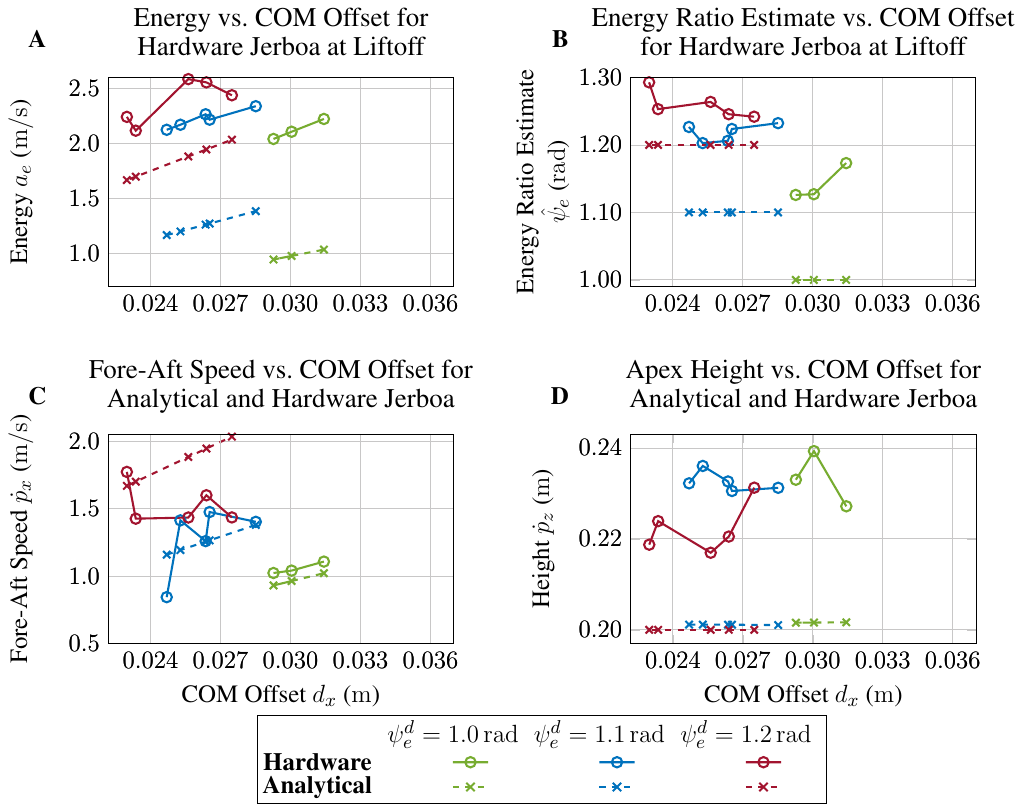}
    \caption{Hardware fixed point validation in energy and energy ratio control vs.\ COM offset on hardware Jerboa for a selection of three different target energy ratios.  Each of these four panels contrasts  analytical  predictions with empirical results in  correspondence with an  equivalently lettered analogue contrasting the analytical predictions and simulation results in \cref{fig_sim_height_speed}. \textbf{(A)} compares empirically measured with analytically computed liftoff energy,  corroborating its predicted monotonic dependence on $d_x$ for the two lower energy ratios. For a target energy ratio of $\psi_e^d=\SI{1.2}{\radian}$ and at higher energies, the motors and touchdown angle begin to saturate due to joint and torque limits, respectively, leading to a plateau and decrease in performance. \textbf{(B)} conveys the independence of estimated liftoff energy ratio values, $\hat{\psi}_e$,  to variations in CoM offsets, $d_x$, at relatively constant magnitudes that roughly increase (while beginning  to saturate)  with increases in the fixed values of their commanded target, $\psi^d_e$.   \textbf{(C)} \& \textbf{(D)} compares empirically measured with model predictions of fore-aft speed and apex height respectively. In \textbf{C}, increases to $d_x$ are correlated to increases in fore-aft speed, while in \textbf{D} increases to the desired energy ratio, $\psi_e^d$ is correlated with a decreased hopping height.}
    \label{fig_hard_hybrid_params}
    \setcounter{subfigure}{0}
    \refstepcounter{subfigure}\label{fig_energy_hardware}
    \refstepcounter{subfigure}\label{fig_phase_hardware}
    \refstepcounter{subfigure}\label{fig_speed_hardware}
    \refstepcounter{subfigure}\label{fig_apex_hardware}
\end{figure*}

Before analyzing the overall performance of the controller in the coming sections, this section corroborates the key prediction from \cref{sec_model}: increasing the horizontal COM offset $d_x$ increases the steady state ``energy" $a_e$ measured in $\si{m/s}$ (more specifically $a_e$ is twice the mass-specific square root of energy) as defined in \cref{eq_hybrid_vars}. While this is not the classical definition of energy measured in joules, there is a bijective relationship where $E = 1/2 m a_e^2$ and $E$ is the energy in \si{joules}.
\Cref{fig_energy_hardware} verifies that the control roughly maintains this trend even on hardware.
Furthermore, \cref{fig_phase_hardware} displays data that helps to validate the efficacy of feedback control of the estimated energy ratio $\hat{\psi}_e$ as calculated by \cref{eq_psie_est} from the stepping controller in \cref{eq_stepping}. 
% \begin{figure}[tbh]
%     \centering
%     % \includegraphics[width=\linewidth]{hardware/phase_sweep}
%     \input{./figs/hardware/phase_sweep.tex}
%     \caption{Estimate of the energy ratio at liftoff vs.\ COM offset on Jerboa in hardware. Increasing the target energy ratio results in a clear increase of achieved values, though $\hat{\psi}_e$ does begin to saturate at higher values, and the final value is always higher than the target, though this may also be a result of previously-mentioned over estimation. Even further, the energy ratio is roughly similar for a given target independent of the COM offset.}
%     \label{fig_phase_hardware}
% \end{figure}
The estimated energy ratio increases when its target increases and is relatively independent of COM offset with no obvious dependence on $d_x$. 
As \cref{sec_psie_est} notes, the estimate of energy ratio will have error compared to actual energy ratio at liftoff due to noise and imprecise mode detection.
Furthermore, the roll from the boom causes Jerboa's two legs to touch down and lift off from the ground at slightly different times, making mode transitions harder to detect, further injecting noise into the $\psi_e$ estimate.
Fortunately, the expected trend is present: increasing the target energy ratio $\psi_e^d$ increases the estimated value of $\hat\psi_e$.

\subsection{Representative Steady State Trajectories}\label{sec_hard_time_traces}
\begin{figure*}[tb]
    \centering
    \includegraphics{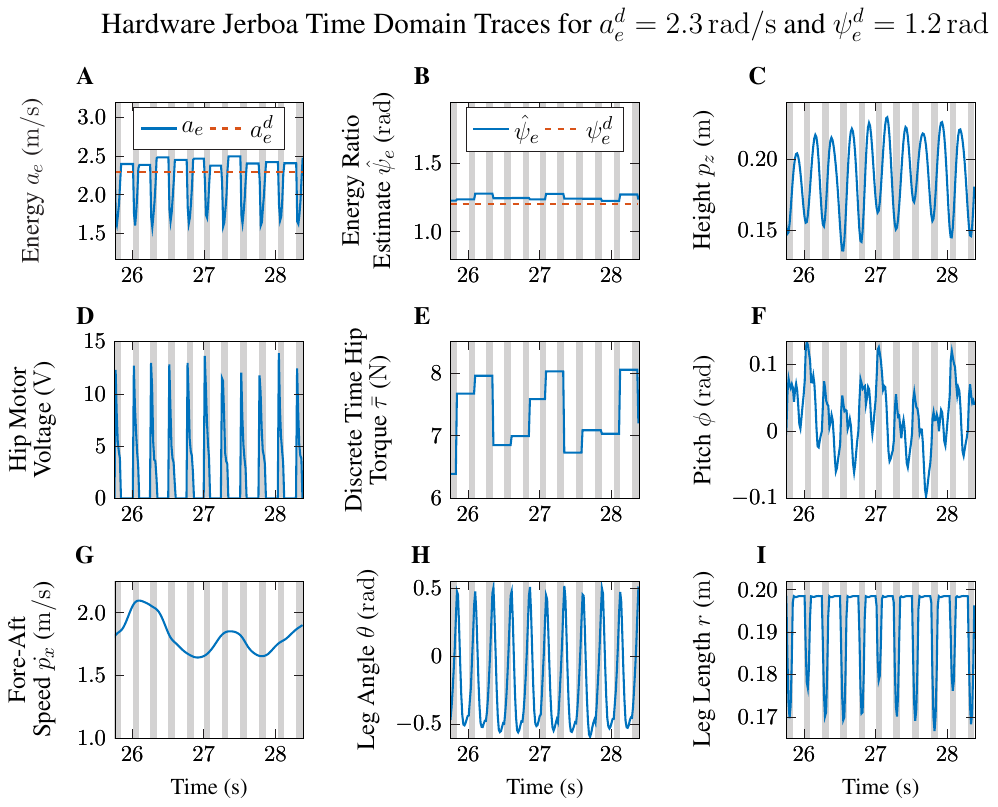}
    \caption{Example trajectories of various state values on hardware during steady state locomotion. Periods of stance are highlighted with a gray background. These plots can generally be compared directly to those in \cref{fig_sim_time} with the exception of the motor torque in \cref{fig_hard_time_torque}, which is now measured in volts as opposed to the newton meters in \cref{fig_sim_time_torque}. All values are measured or calculated using sensors on-board Jerboa with the exception of the height (\cref{fig_hard_time_height}) and fore-aft speed (\cref{fig_hard_time_dx}), which are measured by encoders on the boom. The controller uses the same target values for energy and energy ratio as in \cref{fig_sim_time}.}
    \label{fig_hard_time}
    \setcounter{subfigure}{0}
    \refstepcounter{subfigure}\label{fig_hard_time_energy}
    \refstepcounter{subfigure}\label{fig_hard_time_phase}
    \refstepcounter{subfigure}\label{fig_hard_time_height}
    \refstepcounter{subfigure}\label{fig_hard_time_torque}
    \refstepcounter{subfigure}\label{fig_hard_time_tau0}
    \refstepcounter{subfigure}\label{fig_hard_time_pitch}
    \refstepcounter{subfigure}\label{fig_hard_time_dx}
    \refstepcounter{subfigure}\label{fig_hard_time_theta}
    \refstepcounter{subfigure}\label{fig_hard_time_r}
\end{figure*}

This section discusses features of a set of steady state trajectory traces recorded from a typical physical  experiment  that reprise corresponding traces from simulation data presented in \cref{sec_sim_time_traces}.   To facilitate comparison, the hardware controller used the same target energy $a_e^d$ and energy ratio $\psi_e^d$ as the simulation controller in \cref{fig_sim_time} albeit the corresponding gains were different in consequence of the inevitable discrepancy between physics and models (the "sim-to-real" gap).
\Cref{fig_hard_time_energy,fig_hard_time_height,fig_hard_time_torque,fig_hard_time_theta} exhibit a persistent steady state pattern even as sensor noise and the inevitable unmodeled physical disturbances 
incur substantial variations in step-to-step peaks and troughs relative to the corresponding traces of \cref{fig_sim_time}. These multi-step variations are particularly apparent in \cref{fig_hard_time_phase,fig_hard_time_tau0} whose peaks and troughs roughly coincide.
Although the  oscillations in fore-aft speed plotted in  \cref{fig_hard_time_dx} also follow the same pattern of peaks and troughs,  its substantial variations between steps (again,  stemming from encoder noise in Jerboa’s boom), motivate our use of average steady state speed when discussing fixed points in the next section.
    
Notwithstanding these step-to-step fluctuations, control is still successful. Despite some variation in energy ratio, discrete time hip torque, and maximum leg length compression (\cref{fig_hard_time_phase,fig_hard_time_tau0,fig_hard_time_r}), they all stay fairly close to a nominal steady state value.
Furthermore, the energy $a_e$ in \cref{fig_hard_time_energy} exhibits the same trend as it did in simulation (\cref{fig_sim_time_energy}), decreasing and increasing repeatedly during each stance. 
The energy $a_e$ also appears to be closer to its target energy than $a_e$ in simulation; this is likely coincidental due to the lack of an integrator in the energy controller \cref{eq_ae_controller}.
    Similar to the energy $a_e$, the height (\cref{fig_hard_time_height}) and leg angle $\theta$ (\cref{fig_hard_time_theta}) both appear to converge to a neighborhood of a fixed point.
        Interestingly, the leg angle is even more symmetric between touchdown and liftoff than it is in simulation (cf.\ \cref{fig_sim_time_theta}).
At the same time, the hip motor voltage\endnote{The figure contains the voltage for hardware as opposed to presenting torque directly because Jerboa's hardware does not allow for control of current.} in \cref{fig_hard_time_torque} follows the same trend it did in simulation (\cref{fig_sim_time_torque}).
These all demonstrate successful control of Jerboa in hardware.

\subsection{Fixed Points and Model Accuracy}
\label{sec_hard_accuracy}
In this section we apply the same input output perspective from \cref{sec_sim_fixed_points} to the hardware data, comparing hardware apex coordinate fixed points for fore-aft speed and apex height in \cref{fig_speed_hardware,fig_apex_hardware}, respectively, to those from the analytical model in \cref{eq_3dof_slip_fixed} converted to apex coordinates using \cref{eq_apex_coordinate_change}. As in \cref{sec_sim_fixed_points} the data is collected by first throwing the robot before waiting for convergence (typically after $\sim$20 steps) to a near steady state behavior. Once the robot reaches steady state we collect data for $\sim\SI{30}{seconds}$ to reject the step to step variations in behavior. The range of set points is chosen to demonstrate the full stable operating regime of the robot.
The data in these two figures are less clean than they were for simulation, but nevertheless the same trends are present. For the lower target energy ratio $\psi_e^d$ there is a clear monotonic relationship between center of mass offset $d_x$ and fore-aft speed shown in \cref{fig_speed_hardware}. At $\psi_e^d = 1.2 \si{rad}$ this relationship begins to fail suggesting that the assumptions are breaking down in the face of operation at a higher energy level as discussed in \cref{sec_slip_ass} and that the robot's ability to recirculate the leg is saturating due to the increasingly short apex height. 
Next, for the apex height in \cref{fig_apex_hardware}, the equally-messy hardware data again follows the analytical model's trends: changing the COM offset has little discernible effect on the apex height, while decreasing the phase results in an increased height. 
The hardware result also appears to follow the analytical prediction that the apex height should not substantially increase from a target energy ratio of $\psi_e^d=\SI{1.1}{rad}$ to one of $\psi_e^d=\SI{1.0}{rad}$.

As a result, the accuracy of the model relative to the hardware data in \cref{tab_hard_accuracy} has a mean percent error in speed of $16.4\%$ and in apex height of $12.1\%$ which is worse than the parallel table for the simulation accuracy \cref{tab_sim_accuracy}. As with \cref{tab_sim_accuracy} the data present in \cref{tab_hard_accuracy} for fore-aft speed has a high coefficient of variation, $\approx 0.82$, from the assignment of $\gamma$ and $\chi$ as described in \cref{sec:model_params}.  Selecting different values for $\gamma, \chi$ that don't try to minimize the mean error, for example using $\gamma = 0.88$, $\chi = 2.2$, results in the mean percent error for speed being $\SI{18.4\pm10.1}{\percent} $ which has a coefficient of variation around $0.55$.

\begin{table*}[tb]
    \centering 
    \caption{Accuracy of the apex coordinate fixed points for Jerboa as predicted by the analytical form in \cref{eq_3dof_slip_fixed} compared to the hardware results. The data is from the large operating regime plotted in \cref{fig_hard_hybrid_params} where the number of different desired and measured  robot liftoff steady state values  (each following dozens of hops from various initial conditions) is  $N = 13$.}
    \label{tab_hard_accuracy}
    \begin{tabular}{l l c l c}
        \toprule
        State       & Mean Error   & \begin{tabular}[c]{@{}c@{}} Mean Percent \\ Error \end{tabular}  & Max Error & \begin{tabular}[c]{@{}c@{}}Max Percent \\ Error \end{tabular}  \\
        %   &             &  Percent Error &           & Error \\%\hline
        \midrule
        $\dot{p}_x$      & $0.215\pm\SI{0.177}{m/s}$ & $16.4\pm\SI{13.4}{\percent}$ & $\SI{0.599}{m/s}$ & $\SI{41.8}{\percent}$\\ %\hline
        $p_z$            & $0.028\pm\SI{0.006}{m}$   & $12.1\pm\SI{2.43}{\percent}$  & $\SI{0.037}{m}$ & $\SI{15.7}{\percent}$\\ %\hline
        \bottomrule
    \end{tabular}
\end{table*}
%     In hardware, the mean percent errors in fore-aft speed and height are $\SI{16.4}{\percent}$ and $\SI{12.1}{\percent}$, respectively.
% % Despite this inaccuracy, the hardware's data mirror the trends present in the analytical model's prediction.
% %     For the fore-aft speed in \cref{fig_speed_hardware}, a positive correlation between the speed and the COM offset is still present as expected from the analytical model. Increasing the target energy ratio results in a higher energy, though this increase appears to saturate at higher targets. The analytical model does not predict this saturation perhaps due to energy ratio estimation inaccuracies in implementation.
% Thus, even despite the differences between the simulation and hardware, the model is able to reasonably predict behaviors for both cases.

% hardware sweep:%%%%%%%%%%%%%%%%%%%%%%%%%%%%%%%%%%%%%%%%%%
% compare to analytical:

% energy versus shape
% phase versus shape
% speed versus shape
% apex height versus shape

% want to look at symmetry of gait to justify assumptions
\section{Discussion}\label{sec_discussion}

\subsection{Model, Simulation, and Hardware Differences}\label{sec_hard_sim_diff}
As with any system, the hardware, simulation, and analytical model differ. Fortunately the key parametric trends predicted by the model are present in both experimental data sets with moderate accuracy.

The main difference between the model and experimental results is that the model underestimates the apex height in both simulation and hardware in \cref{fig_sim_height,fig_apex_hardware}. This is likely caused both by the steady state error in the energy ratio from \cref{fig_sim_phase,fig_phase_hardware} as well as the model underestimating the stepping asymmetry resulting in the model predicting a shallower liftoff velocity angle in comparison to the experimental liftoff velocity angle. One explanation for this is that our model is not accurately estimating $\omega_e$, the relative frequency of $\psi_e$ relative to the master phase, which is used in both controlling the energy ratio in \cref{eq_stepping} and in estimating the stepping asymmetry in \cref{eq_xi_theta} due to assuming $b = \Oeps$ in \cref{ass_b}. \Cref{ass_b} is a mathematically
convenient substitute for the empirically more realistic observation during stable operation that $b \dot r = \Oeps$ compared to the force from the spring. As a side effect of this assumption the role of damping in $\dot \psi_e$ \cref{eq_slip_ass_dyn} is considered $\Oeps$ even though $b$ does not appear with $\dot r$ impacting the computation of $\omega_e$.

% First looking at the control of the energy ratio $\psi_e$. 

%  First looking at the energy ratio $\psi_e$'s effect on height, there is a clear relationship between $\psi_e$ and apex height predicted by the model (\cref{fig_hybrid_variables,fig_analytical_height}).
%  While both \cref{fig_sim_height,fig_apex_hardware} indicate inflated apex heights, they both follow the expected relationship between $\psi_e$ and apex height even if there is some offset in the actual value for apex height. One explanation for this discrepancy is that measuring and controlling the energy ratio $\psi_e$ is hard, thus the measured fixed points for energy ratio are greater than the desired energy ratio (\cref{fig_sim_phase,fig_phase_hardware}).

On the other hand the main difference between the simulation and experimental results is that the physical Jerboa's legs are not pogo sticks; they are four bar linkages with extension springs as stated in the beginning of \cref{sec_hard_results}.
    The four bar linkage design results in large non-axial forces especially during a small compression (i.e., $0$--$\SI{2}{cm}$, which is the general range of compression in \cref{fig_sim_time_r,fig_hard_time_r}) \citep{shamsah_analytically-guided_2018}.
    Therefore, instead of applying force entirely in the radial direction, forces are also directed forward.
    Neither the model nor the simulation account for these forward forces and therefore do not account for the legs' obvious effect: they will accelerate the robot forward.
    As such, we expect that Jerboa would hop faster in hardware than the model or simulation predict.
This expectation is verified with the data and in the resulting values for $\chi$ in \cref{tab_model_parameters} which compensates for the increased energization from Jerboa's legs.
    % Even though the model from \cref{eq_3dof_slip_fixed} predicts the fore-aft speed in simulation (\cref{fig_sim_speed}) well, it consistently underestimates the speed for hardware in \cref{fig_speed_hardware}.
    % This difference is a clear and expected result from the design of Jerboa's legs in hardware.
    Despite these large perturbations, the controller still functions in hardware albeit with some periodic oscillations and increased energy.

\subsection{Failure Modes}
While in \cref{sec_hard_results} we presented cases where the robot performed well, the control strategy can fail either due to the robot starting outside of the basin of attraction or due to the onset of instability at the boundaries of the operating regime as delimited by our formal assumptions. Below we categorize the main failure modes observed in the hardware experiments.

\subsubsection{Insufficient COM Offset}
The stability of the pitching subsystem relies on the COM offset $d_x$ being sufficiently large  to counteract the hip torque. While in \cref{ass_pitch} we assume that the lever arm does not depend on $\phi$, in reality it does. 
Specifically, excessively vertical pitching transients during stance create a vanishingly small effective CoM offset,  resulting in correspondingly small hip torques insufficient to replenish the robot's dissipating energy. 
The energy controller in \cref{sec_energy_loop} helps offset this error, though it does not completely mitigate it.
In particular, for Jerboa, increasing $d_x$  entails pitching the tail up in flight, where due to the
conservation of angular momentum causing the body to also pitch, the robot may fail to increase $d \cos \phi$ at the next touchdown.
The persistence of these conditions at the next liftoff
event will decrease $\bar{\tau}$, in a manner that can exacerbate rather than reverse the collapse of offset $d_x$ and twice the mass-specific square root of energy $a_e$.

\subsubsection{Missed Touchdown Angle}
Occasionally the robot is unable to hit the target touchdown angle due to kinematic limits or limited motor speeds. This results in $\psi_e$ and $\dot\theta$ increasing until eventually the robot may stub its toe and
topple as the apex height decreases.

\subsection{Connection Between Posture and Energy}
\label{sec:posture_energy}
While the idea that increasing the average hip torque increases the energy should come at no surprise because of \cref{eq_3dof_slip_fixed}, there is also a connection between posture and the amount of hip torque needed to stabilize pitch in \cref{eq_pitch_fixed_points}. This suggests that in cases where the center of mass is not directly above the hip at steady state, the hip torque that stabilizes pitch will also energize the system. While in some bipedal systems this net  hip torque might pose a problem by violating the assumption that speed is roughly constant in stance (e.g., Raibert stepping in \cite{raibert_legged_1986,de_penn_2015}), a stepping controller can enable the system to realize the energy benefits without ruining stability by compensating for the hip torque with an asymmetry term which balances energy between the radial and angular subsystems as explained in \cref{theorem_asymmetry}.

\subsection{Energy Ratio and Its Connection to Asymmetric Gaits}
\label{sec:energy_ratio_asym}
The second equation in \cref{eq_reset_slip} along with \cref{theorem_asymmetry} reveals a fundamental connection between the symmetry of a gait (or lack thereof) and the change in energy ratio over the course of stance. Thus for gaits where $\psi_e$ changes on the fixed orbit, such as one resulting from a controller with a net hip torque, the stepping policy will be asymmetric about vertical on the fixed orbit as in \cref{fig_sim_time_theta}. This insight suggests that in these cases,
arising from the lack of substantial energetic losses in the angular subsystem,
it is key to treat SLIP as a unitary 2 DoF system rather than as a parallel composition of 1 DoF systems.

\subsection{Stepping Controller Comparison}
Our stepping controller \cref{eq_stepping} is very similar to a conventional Raibert scheme \citep{raibert_legged_1986}, comprising of a neutral point approximator consisting of the first and last term combined with the filter, and a proportional term. The main difference comes from the last term, which accounts for the expected asymmetry at the fixed point.
Additionally, departing from
Raibert’s formulation of the symmetric neutral point, we
view its erstwhile symmetric component as a convex combination of
scissor stepping and a fixed touchdown angle (achieving
his neutral point only at that combination's midpoint).
Not only does this allow us to analyze Raibert stepping with hybrid averaging by treating the contribution of the previous touchdown angle as $\Oeps$, but we are also able to design controllers which have the large basin of scissor stepping while avoiding its period doubling effect.

Comparatively, stepping controllers like VBLA and Peuker's controller \citep{sharbafi_vbla_2016,peekema_template-based_2015} can be viewed as choosing the energy ratio $\psi_e$ at touchdown by setting the touchdown angle relative to one where $\psi_e$ would equal $\SI{0}{rad}$. In contrast, our stepping controller \cref{eq_stepping} attempts to control $\psi_e$ at touchdown to some fixed value using a proportional controller and properties of the reset map.

\subsection{Uses of Core Actuation (Spines and Tails)}
In this paper we compare the application of a hip-energized gait to a planar biped, and a tailed planar biped. While for the planar biped the controller stabilizes the pitch to some fixed point, it can not select the pitch independent of energy. In contrast, when the robot has a tail, the controller stabilizes to a desired pitch and desired energy. This example illustrates how an extra DoF such as a spine or tail enables a secondary objective that can be imposed independently from that of the primary controller.

\subsection{Comparison to Tail-Energized Hopping}
Previous work on Jerboa in \citep{shamsah_analytically-guided_2018} featured a tail-energized controller where the tail pumped energy into the shank, leg springs to energize hopping. The tail-energized controller enabled hopping at speeds of ~$0.2$ -- $\SI{1.0}{m/s}$ with a maximum of 20 hops due to the control strategy's reliance on accurate pitch estimation. Our control strategy represents a strict upgrade in these metrics. It enables Jerboa to hop faster and hop endlessly without falling over at the cost of not being able to hop in place. The success of the hip energized controller likely stems from the improved formal analysis. The math in \citep{shamsah_analytically-guided_2018} was limited to the vertical hopping analysis leaving the hybrid averageability of the overall planar dynamics to the realm of conjecture. In contrast \cref{lemma_cascade_stability} is applied to the full pitch-unlocked Jerboa and instead uses physically motivated assumptions to apply hybrid averaging. The resulting insights led to a more robust controller. Future work on Jerboa should seek to combine tail-energized hopping and hip-energized hopping to result in still more agile and higher performance behavior.

\subsection{Application of Assumptions to Other Legged Machines}
The core theory of this paper presented in \cref{lemma_cascade_stability}, relies on a set of carefully motivated assumptions presented in \cref{table_decomp_assumptions,table_hybrid_averaging_assumptions,table_assumption_reset}. In \cref{sec_math} we discuss these assumptions in the context of Jerboa.  There remains the apt question of how these assumptions hold up when applied to more general bipedal platforms such as Atlas \citep{noauthor_atlas_nodate} or Cassie \citep{Life_at_OSU}? 
For example, at first glance, \cref{ass_1} is immediately violated by Atlas,  whose physical  hip is considerably offset from its mass center.  However, it is important to keep in mind that the the anchored SLIP is a virtual machine, whose effective morphology can be abstractly selected by the anchoring strategy on such a highly actuated physical machine --- its virtual "hip"  need not be co-located with the robot’s  physical hip.
\Cref{ass_pitch,ass_psi_e_replacemt} more likely to prove problematic over the wide range of operating conditions that users would expect their robot to be operable.
In such contexts, the applicability of these results might well require the recourse to gain scheduling over an empirically developed  family of $\chi$ and $\gamma$  values. The rest of the assumptions are, similarly, likely to be applicable within carefully managed SLIP model parameters and subject to the choice of anchoring formulation. 

\subsection{Applications}
Beyond the immediate utility of our novel controller for a unitary 2 DoF SLIP template 
(\cref{table:controllers}) as distinct from the past literature's focus on parallel composition of independent 1 DoF angular and radial subsystems \citep{raibert_legged_1986,de_parallel_2015} to both planar and spatial legged machines \endnote{\DIFadd{While our analysis and experimental results are limited to the sagittal plane, the field has a long history of successfully applying sagittal plane control strategies on spatial robots \citep{de_vertical_2018,greco_anchoring_2023,raibert_legged_1986,peuker_leg-adjustment_2012}.}}, our formal development of  the connections between posture and energy
(\cref{sec:posture_energy}) and between hip energization and gait asymmetry (\cref{sec:energy_ratio_asym}) has two
more general  applications.
First, as the field moves towards bipedal robots
that can manipulate objects of significant mass, these  legged systems may be forced to operate in regimes of high net hip torque that require appropriately asymmetric gaits capable of redirecting their energy in an appropriate and timely enough  manner to avoid pitching over. For example, in today's context of MPC generated strategies,  insights from  \cref{theorem_asymmetry} could easily be applied by reworking the cost function to regularize around an asymmetric stepping pattern rather than a symmetric one. 
Second, still more broadly, starting from the days
of the field’s first untethered running machines that 
omitted shank actuators \citep{Buehler_Battaglia_Cocosco_Hawker_Sarkis_Yamazaki_1998}  and conceived their hip motors, foremost, as indirect sources of  energy to drive passive shank  springs \cite{saranli_rhex_2001},  the effective recruitment of high powered  hip torques has
remained something of an art.
We hope that the combined insight (summarized in \cref{table:controllers}) to ``lean in'' during stance with stepping punctuated to redistribute the resulting momentum may help roboticists reap higher energy legged behaviors by more aggressive use of  hip actuation.

% Outside of the direct application of the controller presented in this paper (\cref{table:controllers}), the ideas present in this paper, particularly the connection between posture and energy (\cref{sec:posture_energy}) and the connection between using the hip torque to energize and an asymmetric gait (\cref{sec:energy_ratio_asym}) have two main applications.
% First, when a robot requires a higher energy behavior and the hip motor is underutilized \SRL{I don't have any good examples of this since its going to be control strategy, behavior, and robot design dependant. I know of situations from my time at BD and from kodlab fokelore, but those are hard to put in the paper.}, adjusting the robot's center of mass forward (through pitching or internal degrees of freedom) and using an asymmetric gait can harness the hip torque to energize the system.
% Second, as the field moves towards bipedal robots that can manipulate objects of significant mass, robots might start to have a net hip torque and a resulting asymmetric gait that energizes the system in a manner that must be carefully redirected in a timely manner to the robot falling over. In the context of MPC, the necessity for an asymmetric gait from \cref{theorem_asymmetry} could easily be applied by tweaking the cost function to regularize around an asymmetric stepping pattern rather than a symmetric one.

\section{Conclusion}\label{sec_conclusion}
In this paper we presented a new hip-energized hopping controller for planar bipeds that allowed the Penn Jerboa to hop at speeds approaching 9 leg lengths/s. 
The controller uses the hip torque to both stabilize pitch and to energize the translational SLIP subsystem.
Meanwhile, the stepping controller moves energy from the angular subsystem to the radial subsystem to counteract energetic losses due to radial damping. We use a novel set of coordinates combined with hybrid averaging to get analytical fixed points and eigenvalues for the resulting closed loop dynamics, providing insight into how the physical and control parameters affect the robot's behavior within an appropriately delimited regime of operation specified by physically motivated assumptions that facilitate the analysis.

In the future we hope to combine this hopping controller with Jerboa's roll controller \citep{wenger_frontal_2016} to allow Jerboa to hop off the boom. Additionally we would like to apply our advances in hybrid averaging to other systems such as a tail energized Jerboa \citep{shamsah_analytically-guided_2018} or a spine quadruped bounding \citep{duperret_towards_2014}.

\begin{acks}
We thank J.\ Diego Caporale for his support with the experimental setup and  Charity Payne, Diedra Krieger, and the rest of the GRASP Lab staff for keeping our lab safe and open during these tumultuous times. 
\end{acks}

\begin{funding}
This work was supported  by ONR grant \# N00014-16-1-2817, a Vannevar Bush Fellowship held by the last author, sponsored by the Basic Research Oﬃce of the Assistant Secretary of Defense for Research and Engineering and partly by the Army Research Office under the
SLICE Multidisciplinary University Research Initiatives Program award under
Grant \#W911NF-18-1-0327
\end{funding}

\theendnotes

\bibliographystyle{IJRR/SageH}
\bibliography{Jerboa}

\appendix
\section{SLIP w/ Attitude Dynamics Derivation}
Given SLIP w/ Attitude (\cref{fig_ASLIP}) consisting of a spring leg with damping, a hip torque, and an offset center of mass, the Lagrangian equations of motion are
\begin{subequations} \label{eq_slip_lagragian_dyn}
    \begin{align}
        \begin{split}
            \ddot{r} =& -\frac{b \dot{r} \left(d^2 m \cos (2 (\theta-\phi))+d^2 m+2 I\right)}{2 I m } \\
            & -\frac{-k r_0 \left(d^2 m \cos (2 (\theta-\phi))+d^2 m+2 I\right) }{2 I m } \\
            & - \frac{-2 d I m \dot{\phi}^2 \sin (\theta-\phi)+2 d m \tau  \cos (\theta-\phi)}{2 I m } \\
            & - \frac{2 g I m \cos (\theta)-2 r I m \dot\theta^2}{2 I m } \\
            & -\frac{r k \left(d^2 m \cos (2 (\theta-\phi))+d^2 m+2 I\right)}{2 I m}\\
            & -\frac{d^2 m \tau  \sin (2 (\theta-\phi))}{2 I m r}\\
        \end{split} \\
        \begin{split}
            \ddot{\theta}=&\frac{ \dot{r} \left(b d^2 \sin (2 (\theta-\phi))-4 I \theta '(t)\right) }{2 I r} \\
            &+\frac{ d^2 k r \sin (2 (\theta-\phi)) + 2 d I \dot{\phi}^2 \cos (\theta-\phi)}{2 I r} \\
            &+\frac{d \sin (\theta-\phi) (\tau -d k r_0 \cos (\theta-\phi))+g I \sin (\theta)}{ I r} \\
            &+ \frac{\tau  \left(-d^2 m \cos (2 (\theta-\phi))+d^2 m+2 I\right)}{2 I m r^2} \\
                    \end{split}\\
        \begin{split}
            \ddot{\phi} = &-\frac{\cos (\theta-\phi) \left(r \left(b d \dot{r}-d k r_0+\tau  \sec (\theta-\phi)\right)\right)}{I r} \\
            &-\frac{\cos (\theta-\phi) \left(d k r^2+d \tau  \tan (\theta-\phi)\right)}{I r} 
            \label{eq_ddphi_complicated}
        \end{split}
    \end{align}
\end{subequations}

In order to obtain \cref{eq_slip_att_ass_dyn} from \cref{eq_slip_lagragian_dyn} we apply \cref{ass_1,ass_2,ass_3}.
\begin{subequations}
\begin{align}
    \Ddot{r} &= \omega_r^2 (r0 - r) - \frac{b \dot{r}}{m} + r \dot\theta^2  - g \cos \theta \\
    \Ddot{\theta} &= \frac{\tau}{m r^2} + \frac{g \sin \theta}{r} - \frac{2 \dot r \dot \theta}{r} \\
    \ddot{\phi} &= \frac{\left(d \cos (\theta - \phi) \right) F_s}{I} - \frac{\tau}{I}\label{eq_ddphi_unsimple}
\end{align}
\end{subequations}
Where 
\begin{equation*}
\label{eq:fs}
F_s \coloneqq - k (r-r_0) - b \dot r    
\end{equation*}
Finally we apply \cref{ass_pitch} to transform \cref{eq_ddphi_unsimple} into \cref{eq_ddphi}.

% \subsection{Averaging Force in the Leg}
% \label{sec_force_leg}
%  In \cref{ass_pitch} we replace the force in the leg with a constant value based on a rough averaging of the force in leg. In \cref{eq:fs_bar} we use the change of coordinates from \cref{eq_hybrid_vars} and explicitly calculate the average value.

% \begin{subequations}
%  \begin{align}
%  \label{eq:fs_bar}
%     F_s &= \frac{a_e \cos \psi_e k \cos \psi_r}{\omega_r} - a_e \cos \psi_e b \sin \psi_r \\
%     \bar F_s &= \int_{-\pi/2}^{\pi/2} \frac{a_e \cos \psi_e k \cos \psi_r}{\omega_r} - a_e \cos \psi_e b \sin \psi_r d \psi_r \\
%     &= \frac{2 k a_e \cos \psi_e}{\omega_r}
%  \end{align}
% \end{subequations}

\subsection{Justification for spring force replacement}
\label{sec_spring_force_just}
\Cref{ass_pitch} implies that the moment about the center of mass from the radial force in the leg is constant. This modeling choice, a key step underlying the isolation of the  pitch subsystem from the translational dynamics as depicted in \cref{fig_robot_diagrams},  does not arise from direct physical observation, but is instead inspired by the role of gravity in the rate of change of the angular momentum about the toe, $L_\text{toe}$. First note that independent of any control law,
\begin{equation}
        \dot L_\text{toe} = (d  \cos \phi + r \sin \theta) m g \label{eq_dltoe1} \,.
\end{equation}
 In order to relate the angular momentum about the toe to the pitching subsystem we first explore what happens on the pitch steady trajectories (i.e. $\phi = \phi^d, \dot \phi = \Ddot{\phi} = 0$). The angular momentum about the toe is 
\begin{align*}
        L_\text{toe} &=  m r^2  \dot \theta + \Upsilon \,,
\end{align*}
where $\Upsilon$ is the angular momentum about the toe arising from the mass center's displacement from the hip and is equal to
\begin{align*}
    \Upsilon & \coloneqq -d m \dot r \cos (\phi^d - \theta)  - d m r \dot \theta \sin (\phi^d - \theta)\,.
\end{align*}
From \cref{eq_ddtheta},
\begin{align*}
    \frac{d}{dt} (m r^2 \dot \theta) &= \tau + m g r \sin \theta\,.
\end{align*}
Thus
\begin{align}
    \dot L_\text{toe} &= \tau + m g r \sin \theta + \dot \Upsilon \label{eq_dltoe2}\,,
\end{align}
and from \cref{eq_ddphi_unsimple} on a pitch steady trajectory,
\begin{align}
    \tau &= d \cos (\theta - \phi)F_s \label{eq_tau} \,.
\end{align}
Setting \cref{eq_dltoe1} and \cref{eq_dltoe2} equal and using \cref{eq_tau} for $\tau$ yields
\begin{align}
    (d  \cos \phi^d + r \sin \theta) m g  &=  d \cos (\theta - \phi)F_s + m g r \sin \theta + \dot \Upsilon \nonumber \\
    d \cos \phi^d mg &= d \cos (\theta - \phi)F_s + \dot \Upsilon \nonumber \\ 
    d_x mg &= d \cos (\theta - \phi)F_s + \dot \Upsilon \label{eq_almost}
\end{align}
Since $\Upsilon$ is small (both terms are multiplied by $d \ll r$),  and is empirically monotone decreasing along steady state motion, we approximate the main effect of  $\dot \Upsilon$ in \cref{eq_almost} by recourse to its abstraction via the fixed parameter, 
$\chi \geq 1$, that replaces the sum on the right hand side with the product. Thus,
\begin{equation}
        d_x \chi mg  \approx  d \cos (\theta - \phi)F_s\,. 
\end{equation}
$\chi$ is chosen numerically according to the methods in \cref{sec:model_params}. 
Intuitively, $\chi$ compensates for the fact that on a steady state trajectory the first term of $\Upsilon$ is usually decreasing moving some momentum into the other terms of $L_\text{toe}$. The approximation  is crucial in maintaining the cascade composition that characterizes the post assumption dynamics. Initial analysis with Jerboa in \cref{sec_hard_accuracy} suggests that for the range of set points a single value of $\chi$ is sufficient to accurately model this phenomenon.  A potentially larger set of operating conditions may require multiple values of $\chi$ depending on energy level to accurately model this phenomenon. 

\section{Proof of Pitching Stability} \label{sec_pitch_proof}
The following is a proof for \cref{lemma_slip_pitch}.
\begin{proof}

We write our stance pitching dynamics in the form of 
\begin{equation*}
    \ddot{\phi} + 2 \zeta \omega_\phi \dot \phi + \omega_\phi^2  \phi + F =  D
\end{equation*}
where 
    \begin{subequations}
    \begin{align*}
    \omega_\phi &= \sqrt{\epsilon k_p/I} \\
    \zeta & = \frac{\sqrt\epsilon k_d}{2 \sqrt{\epsilon k_p I}} \\
    F &= \bar\tau r_0/I \\
    D &= \frac{d_x \chi m g + m g r_0 \sin(\Xi_{\theta,g}) + \epsilon k_p \phi^d}{I} \,.
    \end{align*}
\end{subequations}
From the constraint on $k_d$ in the statement of the lemma, $\zeta = 1$, making the system critically damped. Solving for the stance flow and evaluating it at $T_s$, the stance map is
\begin{subequations}
\begin{align*}
\begin{split}
        \phi_\text{lo}  =& \frac{e^{-T_s \omega_\phi } \left(D \left(-T_s \omega_\phi +e^{T_s \omega_\phi }-1\right)\right)}{\omega_\phi ^2} \\
        & +e^{-T_s \omega_\phi } \left(D (\dot\phi_\text{td} T_s+T_s \phi_\text{td} \omega_\phi +\phi_\text{td}+F T_s) \right) \\
        &+\frac{e^{-T_s \omega_\phi } \left(F \left(-e^{T_s \omega_\phi }\right)+F\right)}{\omega_\phi ^2} 
\end{split}\\
    \dot \phi_\text{lo}  =& e^{-T_s \omega_\phi } \left(T_s \left(D-F-\phi_\text{td} \omega_\phi ^2\right)-\dot\phi_\text{td} T_s \omega_\phi +\dot\phi_\text{td}\right) \, .
    \end{align*}
\end{subequations}
The flight dynamics are $\ddot \phi = 0$, and the flight map is
\begin{subequations}
\begin{align*}
    \phi_\text{td} &= \phi_\text{lo} + \dot \phi_\text{lo} T_f \\
    \dot\phi_\text{td} &= \dot \phi_\text{lo}\\
    F_{k+1} &= F_k +  \frac{\epsilon^2 k_\tau r_0}{I} (\phi_\text{lo} - \phi^d)\,,
    \end{align*}
\end{subequations}
which is also fixed time and includes the discrete time integrator. Evaluating the flight map into the stance map yields the liftoff coordinate return map $\P$ \cref{eq_pitch_return} which takes the form
\begin{equation}
    \P(z_\phi) = S_\P^{-1} A_\P S_\P z_\phi + S_\P^{-1}D_\P\, , \label{eq_pitch_return_full}
\end{equation}
where $S_\P: z_\phi \mapsto [\phi, \dot \phi, F]^T \coloneqq \Delta([1, 1, r_0/I])$ is a linear change of coordinates, $A_\P$ is the state matrix for the pitching subsystem, and $D_\P$ is the constant disturbance.
\begin{align*}
    A_\P & = \left[\!\begin{array}{*{24}{c@{\hspace{2pt}}}}
    \scriptstyle e^{-\kappa} \left(\kappa+1\right) & \scriptstyle e^{-\kappa} \Omega_\phi & \scriptstyle\frac{T_s T_f e^{-\kappa} \left(\kappa-e^{\kappa}+1\right)}{\epsilon k_\phi} \\
\scriptstyle -\frac{\epsilon k_\phi e^{-\kappa}}{T_f} &\scriptstyle -e^{-\kappa} \left(\kappa+\epsilon k_\phi-1\right) & \scriptstyle T_s \left(-e^{-\kappa}\right) \\
\scriptstyle\frac{\epsilon^2 k_\tau r_0   e^{-\kappa} \left(\kappa+1\right)}{I} & \scriptstyle\frac{\epsilon^2 k_\tau r_0   e^{-\kappa} \Omega_\phi}{I} & \lambda_\tau %
  \end{array}\!\right]\\
    D_\P &= \begin{bmatrix}
        \frac{D e^{-T_s \omega_\phi}\left(e^{T_s \omega_\phi -1 - T_s \omega_\phi} \right)}{\omega_\phi^2}\\
        D e^{-T_s \omega_\phi}T_s \\
        \frac{\epsilon^2 k_\tau r_0\left(D - \phi^d \omega_\phi^2 - D e^{-T_s \omega_\phi}\left(1 + T_s\omega_\phi \right)\right)}{I \omega_\phi^2}
    \end{bmatrix}
\end{align*}
where $\Omega_\phi \coloneqq \sqrt{\epsilon k_\phi T_s T_f}+T_s+T_f$, $\lambda_\tau \coloneqq 1 + \frac{e^{-\kappa}k_\tau r_0 T_s \epsilon \left( T_f - e^\kappa T_f + \sqrt{T_s T_f \epsilon k_\phi}\right) }{k_\phi I}$ $\kappa \coloneqq \sqrt{\frac{\epsilon k_\phi T_s}{T_f}}$ and $\epsilon k_\phi \coloneqq \frac{\epsilon k_p T_s T_f}{I} = \omega_\phi^2 T_f T_s$, and---dropping terms that are $\OepsSQ$---the last row of $A_\P$ becomes
\begin{equation*}
    e_3 A_\P = \begin{bmatrix}
    0 & 0 & 1- \frac{\epsilon k_\tau r_0 T_s T_f \kappa }{I k_\phi} + \frac{\epsilon k_\tau r_0 T_s \sqrt{\epsilon k_p I}}{I k_\phi}
    \end{bmatrix}\,.
\end{equation*}

The magnitudes of the eigenvalues of $A_\P$ are
\begin{align*}
    \mynorm{\lambda_1} & = \frac{1}{2} \mynorm{\sqrt{(\epsilon k_\phi-4) \epsilon k_\phi} + (\epsilon k_\phi-2)}   e^{-\kappa}\,,\\
    \mynorm{\lambda_2} & = \frac{1}{2} \mynorm{\sqrt{(\epsilon k_\phi-4) k_\phi} - (\epsilon k_\phi-2)}  e^{-\kappa}\,,\\
    \mynorm{\lambda_3} & = 1- \frac{\epsilon k_\tau r_0 T_s T_f \kappa }{I k_\phi} + \frac{\epsilon k_\tau r_0 T_s \sqrt{\epsilon k_p I}}{I k_\phi}\,.
\end{align*}
From the inequality constraint on $k_p$ in the statement of the lemma, $0<\epsilon k_\phi<4$. Thus, $\mynorm{\sqrt{(\epsilon k_\phi-4) \epsilon k_\phi} \pm (\epsilon k_\phi-2)} = 2$, and the magnitudes of the eigenvalues are $e^{-\kappa}, e^{-\kappa}, 1- \frac{\epsilon k_\tau r_0 T_s T_f \kappa }{I k_\phi} + \frac{\epsilon k_\tau r_0 T_s \sqrt{\epsilon k_p I}}{I k_\phi}$, which are all inside the unit circle since $\kappa > 0$ and because $-1<1- \frac{\epsilon k_\tau r_0 T_s T_f \kappa }{I k_\phi} + \frac{\epsilon k_\tau r_0 T_s \sqrt{\epsilon k_p I}}{I k_\phi}<1$ by inequality constraint on $k_\tau$ in the statement of the lemma. \qed
\end{proof}

\section{Hybrid Averaging}\label{sec_hybrid_averaging}
Hybrid averaging \citep{de_hybrid_2018} is an extension of classical averaging \citep{guckenheimer_nonlinear_2013} to hybrid dynamical systems with only one hybrid mode and has previously been used to analyze monopedal hopping \citep{de_hybrid_2018} and quadrupedal pronking and bounding \citep{de_vertical_2018}. Hybrid averaging gives $\epsilon$-close fixed points of the Poincar\'{e} return map and stability guarantees without requiring integrable dynamics \citep{de_hybrid_2018}.

Hybrid averaging as formulated in \cite{de_hybrid_2018, de_modular_2017} has two key conditions: 
\begin{enumerate}
    \item The continuous time dynamics must be written in the form $\dot \x = \epsilon f(x, \sigma, \epsilon)$,  $\dot\sigma = 1$, where $\sigma \in \mathbb{S}^1$ is the master phase, and $x \in \mathbb{R}^n$ describes the energy coordinates.
    \item The averaged dynamics $\dot y = \epsilon \bar f (y)$, $\dot \sigma = 1$ have fixed points $y^*$, and the reset map $R: y_- \mapsto y_+$ has the property that $R(y^*) = y^*$.
\end{enumerate}

Many systems of interest such as SLIP have a discontinuity in multiple states at the reset event. Requirement 2 means there must be projection from the full state space --- i.e., the $y$-coordinates  governed by the original dynamics  --- onto some transverse section --- i.e. the $x$-state component governed by the autonomous  (master-phase-decoupled) vector field, $f_x$ --- along the constant velocity master phase orbit of the averaged dynamics  wherein the (master-phase-coupled) reset map projects to the identity map. .

%For most systems of interest SLIP, where even at the fixed orbit we expect to have a discontinuous change of value in both the radial velocity and leg angle at the reset event, requirement 2 means we cannot apply hybrid averaging using the naive coordinates. Therefore, we must find a valid change of coordinates.\newnote[GGMc]{What are you trying to say here? It sounds like you want to introduce the change of coordinates, but I'm not sure what tone you want. Mentioning SLIP just sort of comes out of nowhere and I can't tell what your point is.}

\subsection{Averageable Symmetric Hybrid Systems}

In \cite{de_modular_2017} Theorem 4, De takes a very large step in resolving this issue by proving that hybrid averaging can be applied if a system satisfies \cref{def_avik} and the reset map has certain properties.

\begin{definition}[Weakly-coupled almost-reversible continuous dynamics, taken directly from \cite{de_modular_2017}, Definition 2] \label{def_avik}
Given a dynamical system on space $\mathscr{Y}\coloneqq \mathbb{T}^{m+1} \times \mathbb{R}^n$ partitioned in ``phase'' and ``energy'' compartments as 
\begin{align*}
    \bm{y} = [\sigma, \bm\psi, \bm a]^T, 
\end{align*}
where $\sigma \in S^1$ is the master phase, $\psi \in \mathbb{T}^m$ is the coordinate representation of the secondary phases, and $a \in \mathbb{R}^n$ represents the energy of various subsystems, the system has \emph{weakly-coupled almost-reversible continuous dynamics} if 
\begin{enumerate}[label={\roman*)}]
    \item the vector field is of the form\label{def_avik_i}
    \begin{equation}
        \dot{\bm{ y}} = \begin{bmatrix} F_0(\bm a) \\ 0
        \end{bmatrix} + \epsilon F_1(\bm y, \epsilon)\,;\label{eq_gen_form}
    \end{equation} 
    \item the $\sigma$ component of $F_0 > 0$ and $F_0$ is odd in the sense of \cite{de_modular_2017} where odd or even in this case depends on how the ancillary phases relate to the master phase; and\label{def_avik_ii}  % hack to make endnotes work
    \item there is an $a^*$ such that for $\mathcal{R}\coloneqq \{ \bm y \in \mathscr{Y}: a = a^*\}$, $F_1^e(\mathcal{R}) = 0$, where $F_1^e$ is the even component of $F_1$.\label{def_avik_iii}
\end{enumerate}
\end{definition}

De's insight is that, for a system at steady state with multiple phases, the phases will be locked with related frequencies and a constant set of phase offsets. In \cite{de_modular_2017} Lemma 6 and Lemma 8, the author explicitly shows how to calculate the relative frequencies and how to compute a projection from multiple phases to a coordinate system for a section based on a master phase and the relative phase differences. In the following subsection we extend De's result by relaxing \cref{def_avik} (iii), instead checking for the fixed point in the energy subset of the averaged dynamics. This extension will allow the application of \cite{de_hybrid_2018} Theorem 2 to systems that do not satisfy all conditions of \cref{def_avik} (iii).

\subsection{Hybrid Averaging Stability Analysis Computational Recipe for Time Switched Systems}
\label{sec_recipe}
The following computational procedure is a modified version of \cite{de_modular_2017}, Figure 40. We relax the requirement of satisfying \cref{def_avik} (iii) and instead add a check to ensure the averaged system has the necessary fixed point in step \ref{comp_fixed}.
\begin{enumerate}[label={\Roman*}]
    \item Given a system that satisfies \cref{def_avik} (i) and (ii), write it in the form of \cref{eq_gen_form}.
    \item Compute the projection from multiple phases to a single master phase and phase differences.   \label{comp_change}
    Let 
    \begin{align*}
        \bm y &\coloneqq [\sigma, \bm \psi, \bm a]^T \in \mathbb{T}^{m+1}\times \mathbb{R}^n\\
        \bm x &\coloneqq [\bm \delta, \bm a]^T \in \mathbb{R}^{m+n},
    \end{align*}
    \begin{align*}
        \bm x &= \h(\bm y) &&= \begin{bmatrix}\sigma \bm 1 - \Delta(\upomega(\bm a))\bm \psi \\
        \bm a\end{bmatrix} \\
        \bm y &= \h^{-1}(\bm x, \sigma) &&= \begin{bmatrix} \sigma \\ \Delta(\upomega(\bm a))^{-1}(\sigma \bm 1 - \bm \delta) \\ \bm a \end{bmatrix},
    \end{align*}
    where $\Delta(v)$ is a diagonal matrix with diagonal entries $v$ and $\upomega$ is defined as the ratio of frequencies between the master phase and the secondary phase. This ratio $\upomega$ will either be $\upomega = \omega_0(a)$ or $\upomega = \omega(a) \coloneqq \omega_0(a) + \epsilon \omega_1(a)$, depending on if the term $\upomega$ appears in is $\Oeps$ or $\mathcal{O}(1)$ due to dropping $\OepsSQ$ terms.
    \begin{subequations}\label{eq_w}
    \begin{align}
        \omega_0(\bm a) &\coloneqq \Pi_\sigma F_0 \Delta(\Pi_\psi F_0)^{-1} \bm 1         \label{eq_w0}\\
         \omega_1(\bm a) &\coloneqq  \scriptstyle \frac{1}{T} \Delta(\Pi_\psi F_0)^{-1} \Pi_\delta \int_\mathcal{T} H_1 F_1 \circ h_{\omega_0}^{-1}([0, a], \sigma)\, \mathrm{d}\sigma. \label{eq_w1}
    \end{align}
    \end{subequations}
    
    The state dynamics in phase difference coordinates are
    \begin{equation}\label{eq_phase_diff}
        \begin{split}
        \frac{\mathrm{d} \bm x}{\mathrm{d} \sigma} &= \epsilon f(\bm x, \sigma, \epsilon)\\
        &= \epsilon\frac{H_0 F_0 + H_1 F_1}{\Pi_\sigma F_0} \circ h_{\omega_0}^{-1}(\bm x, \sigma) + \OepsSQ\,,\quad
        \end{split}
    \end{equation}
    where \begin{align*}
        H_0(\bm a) &\coloneqq \begin{bmatrix}
                       0 & - \Delta(\omega_1(\bm a)) & 0 \\
                       0 & 0 & 0
                \end{bmatrix}\\
        H_1(\bm a,\bm \psi) &\coloneqq 
        \left[\!\begin{array}{*{24}{c@{\hspace{2pt}}}}
0 & - \Delta(\omega_0(\bm a)) & -\Delta(\bm\psi)D\omega_0(\bm a) \\
                       0 & 0 & I_n  \end{array}\!\right]
            \end{align*}
    according to \cite{de_modular_2017} Lemma 6.
    \item Compute the averaged dynamics with
    \begin{equation}
        \hat{f}(\bm x) \coloneqq \frac{1}{T} \int_\mathcal{T} f(\bm x, \sigma, 0)\, \mathrm{d} \sigma. \label{eq_avg_integral}
    \end{equation} \label{comp_averaging}
    Our choice of $\omega_1$ ensures that $\Pi_\delta \hat{f}([0,\bm a]) = 0$.
    
    \item Check to see if there exists an $a^*$ s.t.\ for $\hat \x^*\coloneqq [0,\bm a^*]$, the averaged dynamics $\bar{f}(\hat\x^*) = 0$. If the system satisfies all of \cref{def_avik}, including (iii), then \cite{de_modular_2017} Lemma 8 guarantees the existence of an $a^*$, though \cref{def_avik} (iii) is sufficient but not necessary. \label{comp_fixed}
    \item Compute $R(\bm x, \sigma_T) = \h \circ R_y\circ \h^{-1}(\bm x,\sigma_T)$ for $\omega = \omega_0(a) + \epsilon \omega_1(a)$ and check that $R(\hat \x^*) = \hat\x^*$. \label{comp_reset}
    \item Compute $D R = \bm S_0 + \epsilon \bm S_1(\hat\x^*) + \OepsSQ$ and check that $S_0$ is constant and invertible and that its unit eigenvalues have a diagonal Jordan block. \label{comp_jacob}
    \item Check that the averaged return map, $D \hat{\bm  P} = \bm S_0 + \epsilon (\bm S_1 + \bm V)$ is hyperbolic for $\bm{V} = T D \hat{f}(\x)$ \label{comp_full_jacob}. 
    % \item By \cite{de_hybrid_2018} Theorem 2, the unaveraged return map has the same stability type as $D \bm P$, and its fixed points are $\epsilon$-close to $\hat\x^*$.\label{comp_eps_fixed}\endnote{In examples presented in \cite{de_modular_2017}, De includes an extra step in checking for hyperbolicity and stability where he drops $\OepsSQ$ terms that arise from computing the eigenvalues even though $D \bm P$ has no $\OepsSQ$ terms. We conjecture here that this step is not necessary. The advantage of this conjecture is that we can apply hybrid averaging to systems with a non sparse Jacobian and more than three dimensions}
\end{enumerate}
% Following this computational recipe yields a system which satisfies the requirements for \cite{de_hybrid_2018} Theorem 2.
\begin{corollary}
    Following the computation recipe in \cref{sec_recipe} yields a system which under \cite{de_hybrid_2018} Theorem 2 there exists an $\epsilon_0 > 0$ s.t. for all $0 < \epsilon < \epsilon_0$, the unaveraged system has a fixed point $\epsilon$ close to $\hat\x^*$ with the same stability type as $D \hat{\bm  P}$.
\end{corollary}
\begin{proof}
    According to \cite{de_modular_2017} Lemma 6, the change of coordinates from step~\ref{comp_change} results in a system of the form of \cite{de_hybrid_2018} equation 12. Next the check in step~\ref{comp_jacob} satisfies condition (i) of \cite{de_hybrid_2018} Theorem 2. The check in step~\ref{comp_fixed} satisfies condition (ii.a) of \cite{de_hybrid_2018} theorem 2, while the check in step~\ref{comp_reset} satisfies the condition (ii.b) of \cite{de_hybrid_2018} Theorem 2. Finally the check in step~\ref{comp_full_jacob} satisfies condition (ii.c). Thus by \cite{de_hybrid_2018} Theorem 2, there exists an $\epsilon_0 > 0$ s.t. for all $0 < \epsilon < \epsilon_0$, the unaveraged system has a fixed point $\epsilon$ close to $\hat\x^*$ with the same stability type as $D \bm P$. \qed
\end{proof}

The advantage of our computational procedure over \cite{de_modular_2017} Fig.~40 is that it allows us to relax \cref{def_avik} and apply \cite{de_hybrid_2018} Theorem 2 to systems that do not satisfy \cref{def_avik} (iii). This comes from checking for the fixed point in the energy subset of the averaged dynamics post change of coordinates (i.e., \cref{eq_av_dyn_slip}), rather than in the even part of the $\Oeps$ unaveraged dynamics for all values of the phase variables (i.e., the even part of \cref{eq_slip_ass_dyn}).

\section{2 DoF Translational SLIP Dynamics Derivation}
The SLIP dynamics in polar coordinates are \cref{eq_ddr,eq_ddtheta}. From here we convert the SLIP dynamics to our desired coordinates using the change of coordinates  defined in \cref{eq_hybrid_vars} where 
\begin{equation}
\label{eq:change_inverse}
    \begin{bmatrix}
        r \\
        \theta \\
        \dot r \\
        \dot \theta 
    \end{bmatrix} = \hs^{-1}(\z) = \begin{bmatrix}
        r_0 - \frac{a_e \cos \psi_e \cos \psi_r}{\omega_r}\\ 
        \theta \\
        a_e \cos \psi_e \sin \psi_r \\
        \frac{a_e \sin \psi_e}{r}
    \end{bmatrix}
\end{equation}
 yielding
\begin{subequations}\label{eq_slip_dyn}
    \begin{align}
        \begin{split}
        \dot{\psi}_v ={}& \omega_r + \frac{a_e \cos \psi_r \sin \psi_e \tan \psi_e}{r} - \frac{b \cos \psi_r \sin \psi_r}{m} \\
        &- \frac{g \cos \theta \cos \psi_r \sec \psi_e}{a_e} 
        \end{split}\\
        \begin{split}
        \dot{\psi}_e ={}& \frac{\bar\tau \cos \psi_e}{m a_e} + \frac{b \cos \psi_e \sin \psi_e \sin^2 \psi_r}{m} \\ 
        &- \frac{a_e \sin \psi_e \sin \psi_r}{r} + g \frac{ \sin \psi_e \sin \psi_r \cos \theta}{a_e }
        \end{split}\\
        \dot{\theta}  ={}&\frac{a_e \sin \psi_e}{r}\\
        \begin{split}
        \dot{a}_e  ={}&\frac{\bar\tau \sin \psi_e}{m}-\frac{b a_e \cos^2 \psi_e \sin^2 \psi_r}{m}\\
        &- g \cos \theta \cos \psi_e \sin \psi_r\,.  \label{eq_dae}
        \end{split}
    \end{align}
\end{subequations}

Next we apply our assumptions (\cref{table_hybrid_averaging_assumptions}). First since $r_0 -r \coloneqq a_e \sin \psi_e \cos \psi_r / \omega_r$, $r_0 -r = \Oeps$\ (\cref{ass_leg_length}). Next we take a first order Taylor series of $r$ about $r_0$. Then we replace $\psi_e$ with $\gamma \psi_e^d$ and $\sin \theta$ with $\sin \Xi_\theta$. Then we group the terms by $\Oeps$ and drop the $\OepsSQ$ terms. This results in the following dynamics
\begin{subequations}\label{eq_slip_ass_dyn}
    \begin{align}
    \begin{split}
        \dot \psi_r  ={}&  \omega_r 
                     + \epsilon \bigg(\frac{-g \cos (\psi_r) \sec (\gamma  \psi_e^d) \cos (\Xi_\theta)}{a_e} \\ 
                     &+  \frac{a_e \cos (\psi_r) \sin (\gamma  \psi_e^d) \tan (\gamma  \psi_e^d)}{r_0} \\
                     &-\frac{b \sin (\psi_r) \cos (\psi_r)}{m} \bigg) 
    \end{split}\\
    \begin{split}
        \dot \psi_e ={}& \dot{ \tilde {\psi}}_e =  \frac{\bar\tau \cos(\gamma \psi_e^d)}{m a_e} \\
        & + \epsilon \bigg(\frac{g \sin (\psi_r) \sin (\gamma  \psi_e^d) \cos (\Xi_\theta)}{a_e} \\
        &+\frac{b \sin ^2(\psi_r) \sin (\gamma  \psi_e^d) \cos (\gamma  \psi_e^d)}{m} \bigg) \label{eq_dpsie_ass}
    \end{split} \\
    \begin{split}
        \dot \theta ={}&  \frac{a_e \sin (\gamma \psi_e^d)}{r_0} \\
        &+ \epsilon \bigg(\frac{a_e^2 \cos (\psi_r) \sin (\gamma  \psi_e^d) \cos (\gamma  \psi_e^d)}{r_0^2 \tilde \omega_r}  \bigg)
    \end{split}\\
    \begin{split}
        \dot a_e ={}& \epsilon \bigg(\frac{\bar\tau \sin (\gamma  \psi_e^d)}{m} -\frac{a_e b \sin ^2(\psi_r) \cos ^2(\gamma  \psi_e^d)}{m} \\
        &-g \cos (\Xi_\theta ) \sin (\psi_r) \cos (\gamma  \psi_e^d) \bigg)
    \end{split}
\end{align}
\end{subequations}

To compute the averaged vector field we first use the projection $\h \circ \g$ as specified in \cref{eq_state_phasediff} and \cref{eq_g_def} where \cref{eq_w0} yields
\begin{align}
    \omega_0(\bm a)  = \begin{bmatrix}
        \frac{a_e m \omega_r \sec (\gamma \psi_e^d)}{\bar\tau} \\
        \frac{r_0 \omega_r \csc (\gamma \psi_e^d)}{a_e}
    \end{bmatrix}\,,
\end{align}
and $\omega_1$ from \cref{eq_w1} is

% \begin{align}
%     \omega_1(\bm a) = \begin{bmatrix}
%         \frac{m \tan (\gamma  \psi_e^d) \left(4 \bar\tau \tan (\gamma  \psi_e^d) \left(a_e^2-g r_0 \cos \left(\frac{\pi  b \cos ^2(\gamma  \psi_e^d) \cot (\gamma  \psi_e^d)}{4 m \omega_r}\right)\right)-\pi  a_e^2 b r_0 \omega_r\right)}{2 \pi  r_0 \bar\tau^2}\\
%         \frac{2 r_0 \cot (\gamma  \psi_e^d) \left(\frac{a_e^2 \left(\tilde \omega_r \tan ^2(\gamma  \psi_e^d)-\omega_r\right)}{r_0 \tilde \omega_r}-g \left(\sec ^2(\gamma  \psi_e^d)+1\right) \cos \left(\frac{\pi  b \cos ^2(\gamma  \psi_e^d) \cot (\gamma  \psi_e^d)}{4 m \omega_r}\right)\right)}{\pi  a_e^2}
%     \end{bmatrix}\,.
% \end{align}

\begin{subequations}
    \begin{align}
    \begin{split}
            \scriptstyle \Pi_1 \omega_1(\bm a) =& \scriptstyle\frac{m \tan (\gamma  \psi_e^d)}{2 \pi  r_0 \bar\tau^2} \bigg(-\pi  a_e^2 b r_0 \omega_r \\&\scriptstyle+ 4 \bar\tau \tan (\gamma  \psi_e^d) \left(a_e^2 -g r_0 \cos \left(\frac{\pi  b \cos ^2(\gamma  \psi_e^d) \cot (\gamma  \psi_e^d)}{4 m \omega_r}\right)\right)\bigg)
    \end{split}\\
    \begin{split}
        \scriptstyle \Pi_2 \omega_1(\bm a) =&\scriptstyle \frac{2 r_0 \cot (\gamma  \psi_e^d)}{\pi  a_e^2} \bigg(\frac{a_e^2 \left(\tilde \omega_r \tan ^2(\gamma  \psi_e^d)-\omega_r\right)}{r_0 \tilde \omega_r} \\ &\scriptstyle-g \left(\sec ^2(\gamma  \psi_e^d)+1\right) \cos \left(\frac{\pi  b \cos ^2(\gamma  \psi_e^d) \cot (\gamma  \psi_e^d)}{4 m \omega_r}\right)\bigg)
    \end{split}
\end{align}

\end{subequations}
\section{SLIP Reset Map Derivation}
\label{app_reset}
Under assumption \cref{ass_flight} we neglect the change in gravitational potential energy in flight, thus the flight map inverts the vertical velocity, $\dot p_z$. Projecting this into the polar coordinates, $[\dot \theta r, \dot r]^T$ the flight map first converts the polar liftoff velocity into the Cartesian liftoff velocity using the liftoff leg angle, then reflects the vertical component, before finally rotating the touchdown Cartesian velocity into the polar touchdown velocity using the touchdown leg angle. Thus,
\begin{align*}
    \begin{bmatrix}
    \dot \theta r \\
    \dot r
    \end{bmatrix}_{\text{td}} 
    &= R_v(\theta_\text{td},\theta_\text{lo})\begin{bmatrix}
    \dot \theta r \\    \dot r
    \end{bmatrix}_{\text{lo}} \, ,
    \end{align*}
where $R_v(\theta_\text{td},\theta_\text{lo}) \coloneqq \bm R(\theta_\text{td})\Delta(1, -1 ) \bm R(-\theta_\text{lo})$ is the matrix which maps liftoff polar velocity to the touchdown polar velocity, $\bm R(\theta)$ is a rotation matrix which rotates a vector by $\theta$, and $\Delta(v)$ is a diagonal matrix whose diagonal elements are $v$. Since at touchdown $a_v = - \dot r$, and at liftoff $[\dot \theta r, \dot r]^T = [a_e \sin \psi_e, a_e \cos \psi_e]^T$,
\begin{align*}
    \vcttwo{\dot \theta r}{a_v} _\text{td} &= \Delta(1,-1) R_v(\theta_\text{td},\theta_\text{lo})\vcttwo{a_e \sin \psi_e}{ a_e \cos \psi_e} _\text{lo}\,,\\
    &= \vcttwo{a_e \sin \left( \theta_\text{td} + \theta_\text{lo} + \psi_e \right)}{a_e \cos \left( \theta_\text{td} + \theta_\text{lo} + \psi_e \right)}\,.
\end{align*}
Thus from \cref{eq_hybrid_vars}, 
\begin{align*}
    \vcttwo{\psi_e}{a_e}_\text{td} = \vcttwo{\theta_{\text{td}}(\z) + \theta_{\text{lo}} + {\psi_e}_{\text{lo}} }{{a_e}_\text{lo}} \,.
\end{align*}

% \begin{align}
%     \begin{bmatrix}
%     \dot \theta r \\
%     a_r
%     \end{bmatrix}_{\text{td}} = \Delta(1,-1)R(\theta_\text{td})\Delta(1, -1 ) R(-\theta_\text{lo})
%     \begin{bmatrix}
%     a_e \sin \psi_e\\ a_e \cos \psi_e
%     \end{bmatrix}
% \end{align}
% Thus in our coordinates the reset map is \cref{eq_reset_slip}

\section{Ancillary Phase Lemma}
The following appendix provides a general method for calculating the ancillary phase offsets s.t. $R(\bm x^*) = \bm x^*$

\subsection{Computation of trajectories}
We start by defining notation for the steady state averaged trajectories in $\z$, $\y$, and $\x$.
Let $f_{\z}, f_{\y}, f_{\x}$ be the stance dynamics in $\z$, $\y$, $\x$, respectively s.t.
\begin{align*}
    f_{\y} &= D \g  f_{\z} \circ \g^{-1} \\
    f_{\z} &= D \h D \g f_{\z} \circ \g^{-1} \circ \h^{-1}
\end{align*}
 Let $ \hat{\x}^* \in \mathcal{X}$ s.t. $\hat f_{\x}( \hat{\x}^*) = 0$ where $\hat f_{\x}$ denotes the averaged dynamics. Let the $\sigma$ varying trajectory of $\hat{\x}$ starting at $\hat{\x}, \sigma_0$ evaluated at $\sigma$ be
 
 \begin{equation*}
     \hat f_{\x}^{\sigma}( \hat{x}, \sigma_0) 
 \end{equation*}
 
 If $\hat{\x} = \hat{\x}^*$, then
 \begin{align*}
     \hat f_{\x}^{\sigma}(\bm{\x}^*, \sigma_0) = \hat{\x}^*
 \end{align*}
 
 We can then transform the trajectory from $\mathcal{X}$ to our other coordinates
 
 Let 
 \begin{equation*}
     T(\bm i_0, \sigma) \coloneqq \inf _{t>0} \Pi_\sigma \hat f_i^t(\bm i_0) = \sigma
 \end{equation*}
 be a the amount of time it takes a trajectory from $\bm i_0$ to reach $\sigma$ for $i \in \{\y,\z\}$. We can now write the trajectory in $\mathcal{Y}$ as
 \begin{align*}
     \hat f_{\y}^{T(\hat{\y}_0, \sigma)}(\hat{\y}_0)
 \end{align*}
 
 if $\hat{\y}_0 = \hat{\y}^*_0 \coloneqq \h^{-1}(\hat{\x}^*, \sigma_0)$, then
 \begin{align*}
     \hat f_{\y}^{T(\hat{\y}^*_0, \sigma)}(\hat{\y}^*_0) = \h^{-1}(\hat{\x}^*, \sigma)\,.
 \end{align*}
 
Since $\hat x$ is $\epsilon$ close to $x$, and $\h$ is continuous for $\omega(a) \not = 0$, then $\hat y$ will still be close to $y$ for some $\epsilon$.

 Similarly the trajectory in $\mathcal{Z}$ is 
  \begin{align*}
     \hat f_{\z}^{T(\hat{\z}_0, \sigma)}(\hat{\z}_0)
 \end{align*}
 and if $\hat{\z}_0 = \hat{\z}^*_0 \coloneqq \g^{-1} \circ \h^{-1}(\hat{\x}^*, \sigma_0)$ then
 \begin{align}
     \hat f_{\z}^{T(\hat{\z}^*_0, \sigma)}(\hat{\z}^*_0) = \g^{-1} \circ \h^{-1}(\hat{\x}^*, \sigma) \label{z_flow}
 \end{align}
 
 which is offset in the ancillary phases from the trajectory in $\mathcal{Y}$. Additionally since $\g$ is continuous, then $\hat{\z}$ and $\z$ are $\epsilon$ close for some $\epsilon$.

 \section{Sufficient Conditions on Reset Map and Ancillary Phase Offset}
  \begin{lemma}[Reset Map Sufficient Condition] \label{lemma_reset_map_general}
    Let $[\bm \delta^*, \bm a^*] \coloneqq \hat{\x}^*$ s.t. $\hat f_{\x}(\hat{\x}^*) = 0$ be the fixed point in the averaged stance dynamics. If 
    \begin{equation}
        R_z([\sigma_T,\bm \psi^*,\bm a^*]) = \begin{bmatrix}
        \sigma_T\\ \bm \psi^*\\ \bm a^*
        \end{bmatrix}
       - \begin{bmatrix}
       \sigma_T-\sigma_0\\ \Delta(\omega^*)^{-1} \bm 1 (\sigma_T-\sigma_0)\\0
       \end{bmatrix}\label{eq_reset_requirment}
    \end{equation}
     for $\hat{ \z}^* = [\sigma_T,\bm\psi^*,\bm a^*] \coloneqq \g^{-1} \circ \h^{-1}(\hat{\x}^*,\sigma_T)$, then $R(\hat{\x}^*) = \hat{\x}^*$
 \end{lemma}
 \begin{proof}
    To prove this lemma, we take our reset map $R_z([\sigma_T,\bm\psi^*,\bm a^*]) = [ \sigma_T, \bm \psi^*, \bm a^*] - [\sigma_T-\sigma_0, \Delta(\omega^*)^{-1} \bm 1 (\sigma_T-\sigma_0),0]$ and show that in $\bm x$ coordinates it is $R(\hat{\x}^*) = \hat{\x}^*$ where $ R = h \circ \g \circ R_z \circ \g^{-1} \circ \h^{-1}$. 
    
    First at the point of interest $ \g \circ R_z \circ \g^{-1}\circ h^{-1}(\hat{\x}^*, \sigma_T) = R_z\circ h^{-1}(\x, \sigma_T)$ since $\Pi_\psi R_z$ is a translation at the point of interest, $\hat{ \z}^*,$ and hence commutes with $\g$ at that point. Next looking at the simplified equation for at the point of interest $R(\hat{\x}^*) = \h \circ R_z \circ \h^{-1}(\hat{\x}^*, \sigma_T)$
\begin{align*}
    \begin{bmatrix}
    \bm \delta \\
    \bm a
    \end{bmatrix}_{td} &= \h \circ R_z \circ \h^{-1}([\bm \delta^*, \bm a^*], \sigma_T) \\
    &= \h \circ R_z \left(\begin{bmatrix}
    \sigma_T \\
    \Delta(\omega^*)^{-1}(\bm 1 \sigma_T - \bm \delta^*) \\
    \bm a^*
    \end{bmatrix}\right) \\\begin{split}
    &= \h \Bigg(\begin{bmatrix}
    \sigma_0 \\
    \Delta(\omega^*)^{-1}(\bm 1 \sigma_T - \bm\delta^*) \\
    \bm a^*
    \end{bmatrix} \\
    &\phantom{aaaaaaa}- \begin{bmatrix}
    0\\
    \Delta(\omega^*)^{-1} \bm 1 (\sigma_T-\sigma_0) \\
    0
    \end{bmatrix} \Bigg)
    \end{split} \\
    &= \h \left(\begin{bmatrix}
    \sigma_0 \\
    \Delta(\omega^*)^{-1}(\bm 1 \sigma_0 - \bm \delta^*) \\
    \bm a^*
    \end{bmatrix}\right)\\
    &= \begin{bmatrix}
    \sigma_0 \bm 1 - \Delta(\omega^*)\Delta(\omega^*)^{-1}(\bm 1 \sigma_0 - \bm \delta^*) \\
    \bm a^*
    \end{bmatrix} \\
    \begin{bmatrix}
    \bm \delta \\
    \bm a
    \end{bmatrix}_{td} &= \begin{bmatrix}
    \bm \delta^* \\
    \bm a^*
    \end{bmatrix} 
\end{align*}\qed
 \end{proof}
\newpage
 \begin{lemma}[Reset map phase offset] \label{lemma_reset_phase_offset}
 Let $\hat{\x}^* \coloneqq [0, \bm a^*] \in \mathcal{X}$ s.t. $\hat{f}_{\x}(\hat{\x}^*) = 0$ be the fixed point in the averaged stance dynamics. If $\bm \Xi = \bm \psi^* - \Delta(\omega^*)^{-1} \bm 1 \sigma_T$ then the averaged $z$ state at liftoff, $\hat f_{\z}^{T(\hat{\z}^*_0, \sigma_T)}(\hat{\z}^*_0) = [\sigma_T, \psi^*, a^*]$ for $\hat{\z}^*_0 \coloneqq \g^{-1} \circ \h^{-1}(\hat{\x}^*, \sigma_0)$.
 \end{lemma}
 \begin{proof}

Leveraging \cref{z_flow}
\begin{equation*}
    \hat f_{\z}^{T(\hat{\z}^*_0, \sigma_T)}(\hat{\z}^*_0) = \g^{-1} \circ \h^{-1}(\hat{\x}^*, \sigma_T)
\end{equation*}
thus

\begin{align*}
    \begin{bmatrix}
    \sigma\\
    \bm \psi\\
    \bm a
    \end{bmatrix}_T &= \g^{-1} \circ \h^{-1}(\hat{\x}^*, \sigma_T) \\
    &= \g^{-1} \left(\begin{bmatrix}
    \sigma_T\\
    \Delta(\omega^*)^{-1}(\sigma_T \bm 1)\\
    \bm a^*
    \end{bmatrix}\right) \\
    &= \begin{bmatrix}
    \sigma_T \\
    \Delta(\omega^*)^{-1}(\sigma_T \bm 1)\\
    \bm a^*
    \end{bmatrix} + \begin{bmatrix}
        0 \\
        \bm \Xi \\
        0
        \end{bmatrix}\\
    &= \begin{bmatrix}
    \sigma_T\\
    \Delta(\omega^*)^{-1}(\sigma_T \bm 1)\\
    \bm a^*
    \end{bmatrix} + \begin{bmatrix}
        0 \\
        \psi^* - \Delta(\omega^*)^{-1} \bm 1 \sigma_T \\
        0
        \end{bmatrix}\\
    \begin{bmatrix}
    \sigma\\
    \bm \psi\\
    \bm a\\
    \end{bmatrix}_T &= \begin{bmatrix}
    \sigma_T \\
    \bm \psi^* \\
    \bm a^*
    \end{bmatrix}   
\end{align*}
\qed 
 \end{proof}
 \begin{remark}
 \cref{lemma_reset_phase_offset} demonstrates our ability to move the liftoff coordinate fixed point of the averaged ancillary phases using the change of coordinate $\g$ so that we can more easily apply \cref{lemma_reset_map_general}. Since $\omega$ is a function of $\bm a$, and $\bm \Xi$ is a function of $\omega$, we need $\bm a^*$ to not depend on $\Xi$ otherwise we will likely end up with a transcendental set of equations for $\bm a^*$ and $\bm \Xi$
 \end{remark}
 \begin{remark}
 \cref{lemma_reset_phase_offset} shows how we can move the liftoff state of the ancillary phases in order to apply \cref{lemma_reset_map_general}.
 \end{remark}

\section{Liftoff to Apex Conversion}
Our fixed points are calculated in liftoff coordinates \cref{eq_3dof_slip_fixed}, but the natural outputs of the system are the apex height and fore-aft speed. For that reason when comparing the accuracy of our model to the experimental results we use apex coordinates and convert our analytical fixed points from liftoff coordinates to apex coordinates using:
\begin{subequations}\label{eq_apex_coordinate_change}
    \begin{align}
    p_{z_\text{lo}}&= r_0 \sin \theta_\text{lo}\\
    \dot p_{z_\text{lo}} &= a_e \cos(\theta_\text{lo} + \psi_e) \\
    \dot p_x{_\text{apex}}&= a_e \sin(\theta_\text{lo} + \psi_e) \\
    p_{z_\text{apex}}&= \frac{\dot p_{z_\text{lo}}^2}{2 g} + p_{z_\text{lo}}  \label{eq_zlo}
\end{align}
\end{subequations}

\section{Computation of Average Leg Angle}
\label{sec_xi_theta_g}
In \cref{ass_xi_theta_replacement} we do a small angle approximation of $\theta$ about $\Xi_{\theta,g}$ which is the average leg angle when we assume gravity acts radially. In this section we are computing the value of $\Xi_{\theta,g}$. First note that assuming gravity acts radially is equivalent to replacing $\Xi_{\theta,g}$ with $0$. Then using the values from \cref{eq_3dof_slip_fixed}, the fixed point for $a_e$ and $\bar\tau$ when we assume gravity acts radially are
\begin{align*}
    a_{e,g}^* &= \frac{2 \bar\tau_g^* \sec(\gamma \psi_e^d)\tan(\gamma \psi_e^d)}{b} \\
    \bar\tau_g^* &= \frac{m g}{r_0} \left(\chi d_x\right)\,.
\end{align*}
Combining these fixed points with the equation for $\Xi_\theta$ \cref{eq_xi_theta} yields the equation for $\Xi_{\theta,g}$ \cref{eq_xi_theta_g}.

\end{document}